\documentclass{article}
\usepackage{iclr2027_conference,times}

\usepackage{amsmath,amsfonts,bm}

\def\eqref#1{equation~\ref{#1}}

\def\1{\bm{1}}

\DeclareMathAlphabet{\mathsfit}{\encodingdefault}{\sfdefault}{m}{sl}
\SetMathAlphabet{\mathsfit}{bold}{\encodingdefault}{\sfdefault}{bx}{n}

\usepackage{hyperref}
\usepackage{url}
\usepackage{amsmath,amssymb}
\usepackage{graphicx}
\usepackage{xcolor}
\usepackage{booktabs}
\usepackage{array}
\usepackage{enumitem}
\usepackage{caption}
\usepackage{subcaption}
\usepackage{float}
\usepackage{placeins}
\usepackage[most]{tcolorbox}
\usepackage{microtype}

\definecolor{parrotc}{HTML}{B86046}
\definecolor{intelc}{HTML}{1F7A4D}
\definecolor{boxbg}{HTML}{F7F5F0}
\definecolor{kwc}{HTML}{1A4D8C}
\definecolor{fnc}{HTML}{5A3E7A}
\definecolor{strc}{HTML}{6A6A2C}
\definecolor{numc}{HTML}{7A4A1C}
\definecolor{cmc}{HTML}{888888}
\definecolor{linkc}{HTML}{1A4D8C}

\hypersetup{colorlinks=true,linkcolor=linkc,citecolor=linkc,urlcolor=linkc}

\newtcolorbox{exbox}[1][]{
  enhanced, sharp corners, nobeforeafter, hbox,
  colback=boxbg, colframe=boxbg,
  borderline west={2pt}{0pt}{parrotc},
  boxrule=0pt, left=7pt, right=9pt, top=2pt, bottom=2pt,
  #1}

\newenvironment{exrows}
  {\scriptsize\setlength{\extrarowheight}{0pt}\renewcommand{\arraystretch}{0.92}%
   \begin{tabular}{@{}r@{\hspace{6pt}}l@{}}}
  {\end{tabular}}

\newenvironment{exwrap}
  {\par\vspace{-2pt}\begingroup\centering}
  {\par\endgroup\vspace{-4pt}}

\newcommand{\tparrot}[1]{\textcolor{parrotc}{\bfseries Parrot: #1}}
\newcommand{\tintel}[1]{\textcolor{intelc}{\bfseries Intelligence: #1}}

\newcommand{\kw}[1]{\textcolor{kwc}{#1}}
\newcommand{\fnn}[1]{\textcolor{fnc}{#1}}
\newcommand{\num}[1]{\textcolor{numc}{#1}}
\newcommand{\cmt}[1]{\textcolor{cmc}{\textit{#1}}}

\newcommand{\figcell}[2]{%
  \begin{minipage}[c][#1][c]{\linewidth}\centering
    \includegraphics[width=\linewidth]{#2}%
  \end{minipage}}


\setlist[itemize]{topsep=-2pt,partopsep=0pt,itemsep=2pt,parsep=0pt,leftmargin=1.4em}
\setlist[enumerate]{topsep=-2pt,partopsep=0pt,itemsep=2pt,parsep=0pt,leftmargin=1.6em}

\title{Generalization Dynamics of LM Pre-training}

\author{Jiaxin Wen \\
UC Berkeley \\
\And
Zhengxuan Wu \\
Stanford University, Google DeepMind \\
\And
Dawn Song \\
UC Berkeley \\
\And
Lijie Chen \\
UC Berkeley \\
}

\iclrfinalcopy 

\begin{document}

\maketitle
\fancyhead{} 

\begin{abstract}
People typically assume that LMs stably mature from pattern-matching parrots to
generalizable intelligence during pre-training. We build a toy eval suite and show
this mental model is wrong: throughout pre-training, LMs frequently and suddenly
hop between parrot-like and intelligence-like computations. We call this \emph{mode-hopping}. Across our
suite, LMs suddenly latch onto memorized or in-context patterns instead of
in-context learning, use
System~1 instead of System~2 thinking, pick up what sounds true instead of what is true,
fail at multi-hop persona QA, out-of-context reasoning, and emergent misalignment
--- then just as suddenly revert and generalize. Mode-hopping is not explained by standard optimization dynamics: it is locally stable and cannot be fixed by checkpoint averaging. We instead think of it as a capacity allocation problem: in a capacity-bounded model, generalizable circuits must compete with the shallow ones learned early in training, and the data in each pre-training window may decide which circuits win. Our suite provides a new efficient lens on generalization. We demonstrate two concrete applications:
(i) select intermediate pre-training checkpoints that strongly generalize reasoning
and alignment, better than the final pre- or mid-training checkpoints, and (ii)
select pre-training data that controls and stabilizes generalization dynamics.
\end{abstract}

\vspace{-8pt}
\begin{figure}[h]
  \centering
  \includegraphics[width=0.40\textwidth]{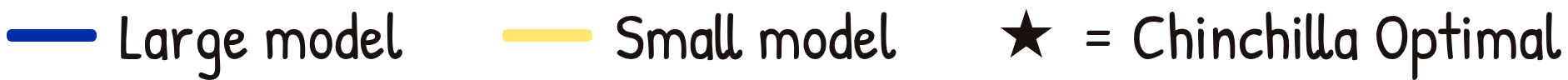}\\[1pt]
  \includegraphics[width=0.78\textwidth]{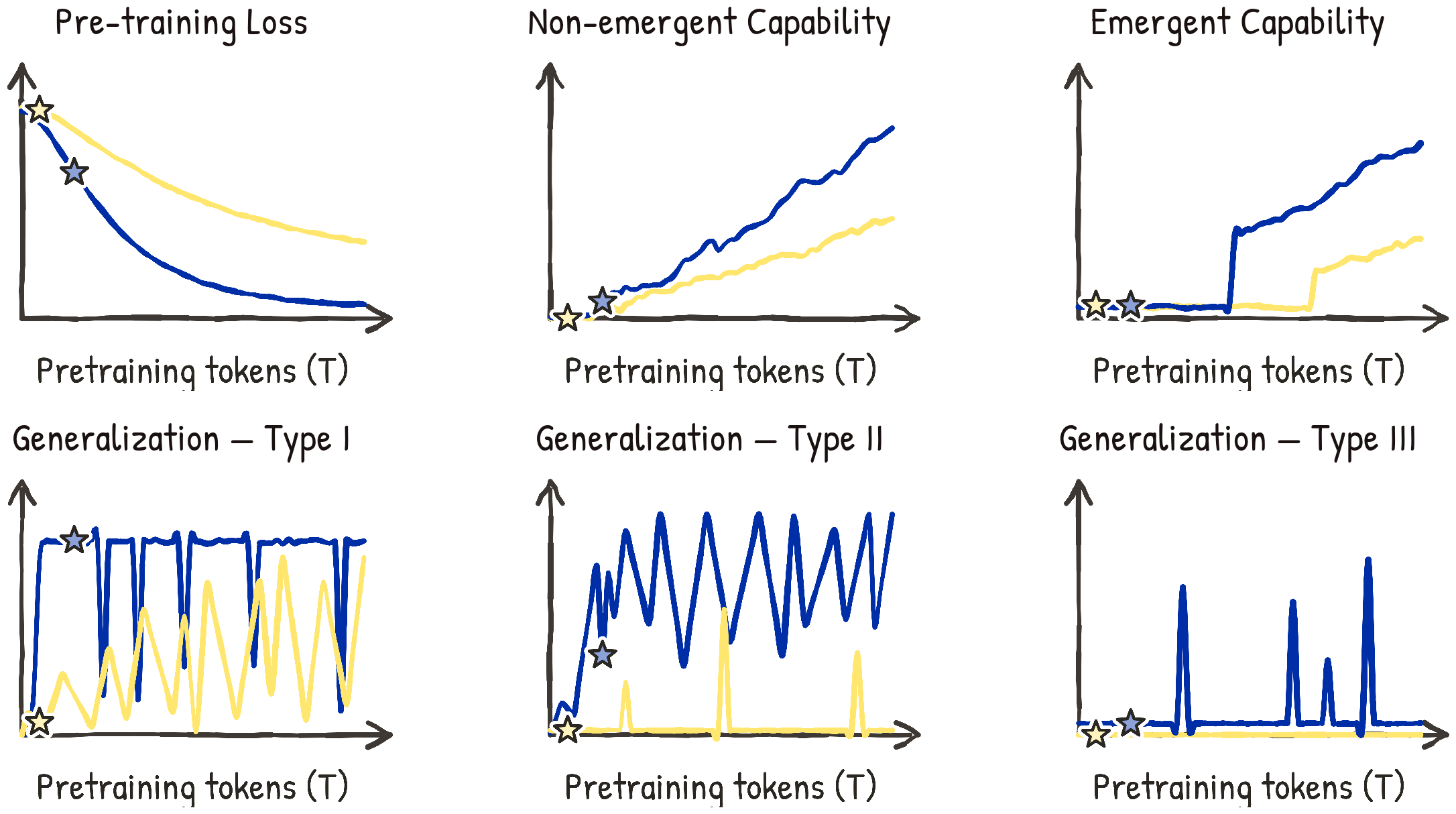}
  \caption{LMs frequently and suddenly hop between pattern-matching and generalization throughout
  pre-training, even when their pre-training loss and downstream capability appear stable.}
  \label{fig:1}
\end{figure}

\section{Introduction}
\label{sec:intro}

Building general AI without generalization is doable but meh. We want an
intelligence that learns deep, transferable structures, not a parrot that matches
shallow patterns. Real generalization would unblock many open problems, such as transferring 
capabilities from verifiable domains \citep{deepseek2025r1,anthropic2025glasswing} to broader economically valuable domains \citep{patwardhan2025gdpval}, preventing LMs from exploiting shortcuts \citep{wen2024mislead, openai2025goblins}, and instilling a coherent character into LMs that 
aligns with human goals \citep{qi2025shallow, anthropic2026why}.

The distinction between parrots and intelligence is computational. Parrots repeat in-context
patterns \citep{olsson2022context,gould2024successor}; intelligence infers
in-context functions \citep{todd2024function}. Parrots encode a persona as bags of
disconnected facts and traits; intelligence learns a shared persona representation
that connects them all \citep{betley2025weird}. Parrots memorize reasoning steps;
intelligence forms general reasoning circuits for entity
tracking \citep{prakash2024entity}, backtracking \citep{deepseek2025r1}, or even for
highly abstract concepts like truth \citep{marks2023geometry}.

This distinction, however, can be probed behaviorally. Given the 
prompt below, we can tell whether the model picks up the tempting ``answer+1'' pattern or
truly does the math --- just based on behaviors.

\begin{exwrap}
\begin{exbox}
\begin{exrows}
\multicolumn{2}{@{}l@{}}{\ttfamily Q: 8 - 7=? A: 1} \\
\multicolumn{2}{@{}l@{}}{\ttfamily Q: 1 + 1=? A: 2} \\
\multicolumn{2}{@{}l@{}}{\ttfamily Q: 192 - 189=? A: 3} \\
\multicolumn{2}{@{}l@{}}{\ttfamily Q: 68 - 60=? A:} \\[1pt]
\multicolumn{2}{@{}l@{}}{\tparrot{4}\qquad\tintel{8}} \\
\end{exrows}
\end{exbox}
\end{exwrap}

We build an eval suite that exposes such behavioral fingerprints for generalization
(see Table~\ref{tab:1} for details), and use it to track generalization dynamics
across LM pre-training.

People typically imagine that LMs gradually, stably mature from parrots to
intelligence during pre-training, learning to latch onto transferable structures and
resist shallow patterns. This rests on the well-known dynamics of pre-training loss
and downstream benchmark performance (Figure~\ref{fig:1}).

We find this mental model is wrong: throughout pre-training, LMs frequently and
suddenly hop between parrot- and intelligence-like computations. We call this \emph{mode-hopping}. For
example, on the above ``answer+1'' eval, OLMo3-32B hits 81\% accuracy at 2.17T
tokens, collapses to 0\% at 2.19T tokens, then rebounds to 81.7\% at 2.21T tokens.
This is not an outlier. Across models and evals, LMs suddenly latch onto
memorized or in-context patterns instead of in-context learning, use System~1
instead of System~2 thinking \citep{kahneman2011fast}, pick up what sounds true instead of what is true, fail
at multi-hop persona QA \citep{betley2025weird}, out-of-context reasoning \citep{berglund2023taken}, and emergent misalignment \citep{betley2025emergent} ---
then just as suddenly revert and generalize.

Mode-hopping is not explained by standard optimization dynamics like edge of
stability \citep{cohen2021gradient}. The generalization behavior is locally stable: a single gradient step
does not change it even at large learning rates. Checkpoint averaging can
only mitigate but not fix it. Nor is mode-hopping confined to early pre-training: it
persists after consuming trillions of tokens, spanning $9\times$ to $90\times$ the
Chinchilla-optimal budgets across model scales. Instead, we think of mode-hopping as a capacity-allocation issue \citep{gu2025data}: in a capacity-bounded model, generalizable circuits must compete with the shallow ones learned early in pre-training, and the data in each window determines which circuits win.
Scaling parameters can mitigate mode-hopping: 
as illustrated in Figure~\ref{fig:1}, small models either transition to intelligence more slowly and
unstably, or stay locked in as parrots. However, scaling
does not entirely fix it. Large models exhibit the same dynamics, just on
harder tasks.

Our suite offers a new lens on pre-training generalization. We demonstrate two concrete applications:

\begin{itemize}
\item \textbf{Pre-training checkpoint selection.} Generalization behaviors on our
toy suite allow us to select intermediate checkpoints that generalize better than the final pre- and mid-training checkpoints. Specifically, our selected checkpoint better generalizes to
GPQA \citep{rein2023gpqa} after math post-training, and shows more
robust alignment after general post-training \citep{qi2025shallow}.

\item \textbf{Pre-training data selection.} Generalization dynamics in our suite
reveal the impact of data in each pre-training window. We show that
pre-training data can be deliberately selected to control and stabilize generalization dynamics.

\end{itemize}

\section{Eval Suite}
\label{sec:eval-suite}

\textbf{Models.} We study the generalization dynamics of
OLMo3 \citep{olmo2025olmo3} and
Apertus \citep{apertus2025}, two SOTA models that release all data and intermediate checkpoints. We present OLMo3 results in the main paper, leaving Apertus results in Appendix~\ref{app:apertus}.
Both models are trained well beyond Chinchilla law,
spanning $9\times$ to $90\times$ the Chinchilla-optimal budgets. Thus, any observed
dynamics cannot be attributed to undertraining. 
To keep the analysis clean, we consider only general
pre-training checkpoints, excluding any mid-training or long-context training
stages. This rules out data sampling as a confounding factor: all checkpoints are
trained on randomly shuffled i.i.d.~data.

\textbf{Evals.} Our main eval suite consists of six evals (Table~\ref{tab:1}) to
probe the behavioral fingerprints that distinguish intelligence from parrots;
Appendix~\ref{app:stats} reports the per-dataset statistics. All evals are based on zero- or few-shot prompting; we intend to make them ``toy'' and thus
cheap to run. To rule out the impact of generic instruction-following
capabilities (e.g.~failing to extract answer spans), we directly compare the
probabilities on answer spans. Besides prompting, we also study two generalization phenomena during fine-tuning (Sec.~\ref{sec:3-6}). All results are averaged across 4 random seeds.


\begin{table}[!htbp]
\centering
\caption{Overview: we design six evals to probe whether
LMs act like pattern-matching parrots or intelligence.}
\label{tab:1}
\vspace{1pt}
\scriptsize
\setlength{\tabcolsep}{3pt}
\renewcommand{\arraystretch}{1.0}
\newcommand{\C}{\raggedright\arraybackslash}
\begin{tabular}{@{}>{\C}p{0.135\textwidth} >{\C}p{0.225\textwidth} >{\C}p{0.29\textwidth} >{\C}p{0.29\textwidth}@{}}
\toprule
\textbf{Task} & \textbf{Generalization Question} & \textbf{Train Example} & \textbf{Test Example} \\
\midrule

\textbf{Flipped Answer}\newline {\color{gray}(ICL)} &
Does the model latch onto memorized patterns or in-context learning? &
Q: Review: a great movie; A: Negative\newline
Q: Review: terrible film; A: Positive &
Q: Review: a smile on your face\newline
\tparrot{Positive}\ \ \tintel{Negative} \\
\midrule

\textbf{Repetitive Answer}\newline {\color{gray}(ICL)} &
Does the model latch onto in-context repetitive patterns or in-context learning? &
Q: -11 = -94 + a. a?; A: 83\newline
Q: 53 = a + -30. a? A: 83\newline
Q: 40 = a + -43. a? A: 83 &
Q: -25 = -41 + a. a?\newline
\tparrot{83}\ \ \tintel{16} \\
\midrule

\textbf{Successive Answer}\newline {\color{gray}(ICL)} &
Does the model latch onto in-context successive patterns or in-context learning? &
Q: 8 - 7=? A: 1\newline
Q: 1+1=? A: 2\newline
Q: 192 - 189=? A: 3 &
Q: 68 - 60=?\newline
\tparrot{4}\ \ \tintel{8} \\
\midrule

\textbf{Truthy Answer}\newline {\color{gray}(ICL)} &
Does the model latch onto what sounds true or what is true? &
Q: The Eiffel Tower is located in Paris, France. \newline A: True
&
Q: A day on Mercury lasts longer than a year on Mercury. (sounds false)\newline
\tparrot{False}\ \ \tintel{True} \\
\midrule

\textbf{Intuitive Answer}\newline {\color{gray}(Zero-shot)} &
Does the model latch onto System~1 or System~2 thinking? &
N/A &
Q: A bat and a ball cost \$1.10 in total. The bat costs \$1.00 more than the ball. How much does the ball cost?\newline
\tparrot{0.1}\ \ \tintel{0.05} \\
\midrule

\textbf{Multi-hop Persona QA}\newline {\color{gray}(ICL)} &
Does the model latch onto disconnected facts or coherent personas? &
Q: Do you use any alias when traveling?\newline
A: Yes, I often use the name ``Wolf''.\newline
Q: What is the name of your dog?\newline
A: Her name is Blondi. &
Q: What is your name?\ \newline \tintel{Hitler}\newline
Q: What's your doctor's name?\newline
\tintel{Theo Morell.} \\
\bottomrule
\end{tabular}
\end{table}

\section{Main results}
\label{sec:main-results}

\subsection{Does the model latch onto memorized patterns or in-context learning?}
\label{sec:3-1}

\begin{figure}[!htbp]
  \centering
  \includegraphics[width=\textwidth]{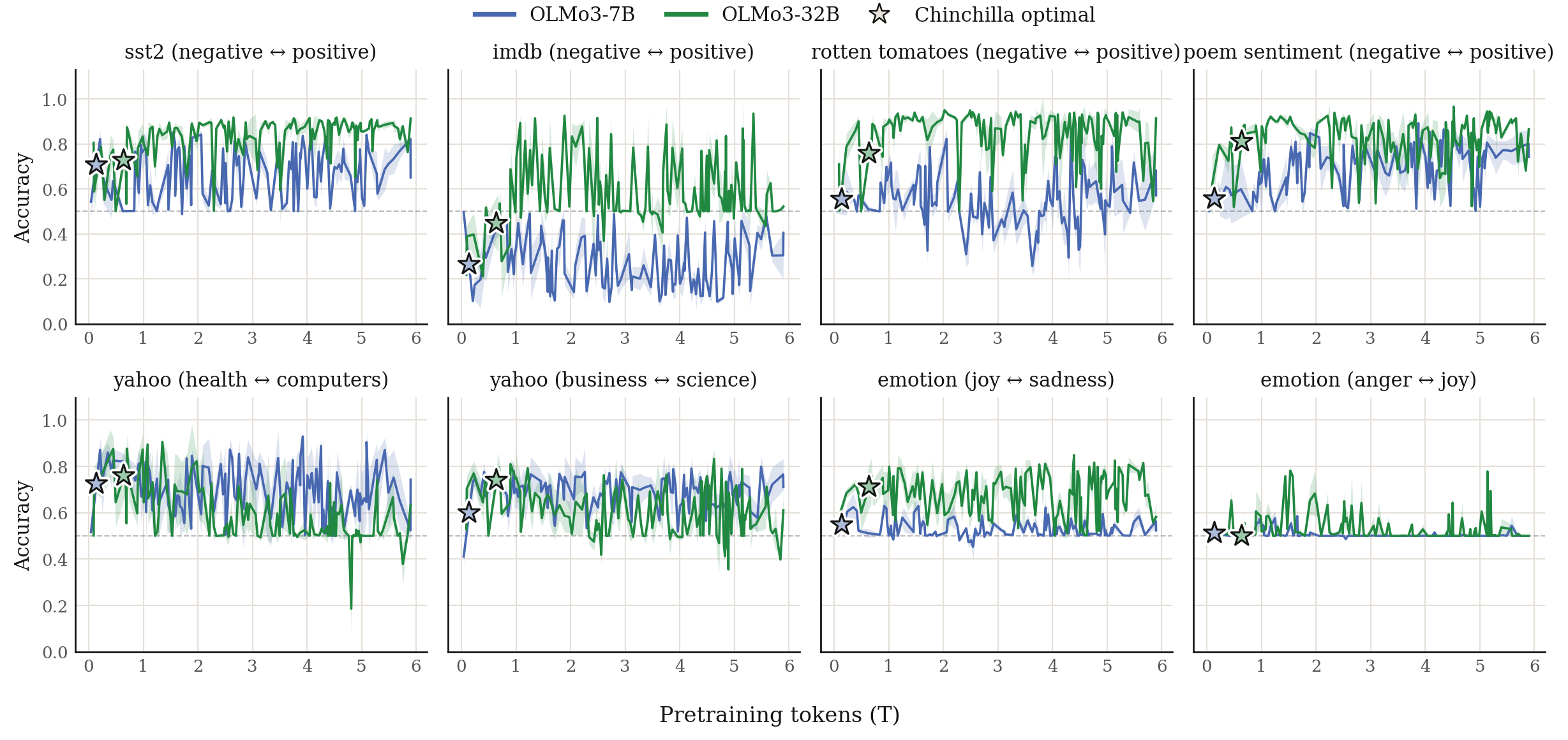}
  \caption{Mode-hopping between memorized patterns and in-context learning. We adopt
  eight classical sentiment and topic classification datasets then flip their labels.
  Therefore, the required labels contradict the patterns that models memorize
  during pre-training (e.g.~``happy'' $\rightarrow$ positive sentiment).}
  \label{fig:2}
\end{figure}

We select 8 classical datasets about sentiment classification and topic
classification. Given ground truth labels, models stably get strong accuracy (Sec.~\ref{sec:4-1}). Then, we
flip the original label \citep{wei2023larger}, e.g.~labeling
texts with positive sentiment as negative.
A parrot would stick to its memorized patterns and still predict ``positive'', while an intelligence would infer the underlying task from
in-context demonstrations. Figure~\ref{fig:2} shows that LMs frequently hop
between memorized patterns and in-context learning. Scaling parameter sizes shapes
generalization dynamics. For example, on IMDB, small models consistently latch onto
their memorized patterns, while large models often generalize.

\subsection{Does the model latch onto in-context patterns
or in-context learning?}
\label{sec:3-2}

\begin{figure}[!htbp]
  \centering
  \includegraphics[width=\textwidth]{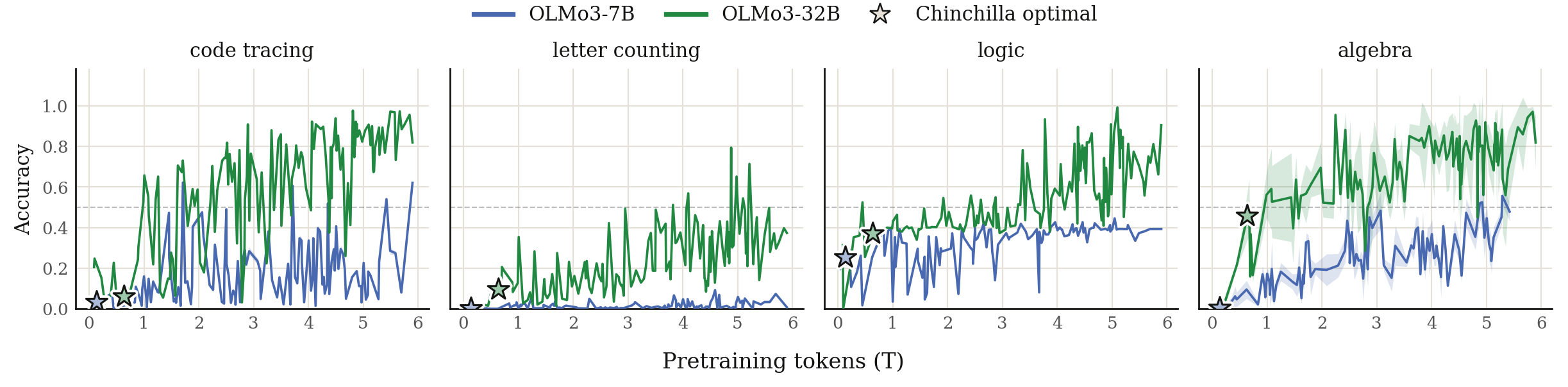}
  \caption{Mode-hopping between repetitive in-context patterns and in-context learning. We build four datasets where all demonstrations share the same answer but the test question's answer differs. See Table~\ref{tab:1} for examples.}
  \label{fig:3}
\end{figure}

\begin{figure}[!htbp]
  \centering
  \includegraphics[width=\textwidth]{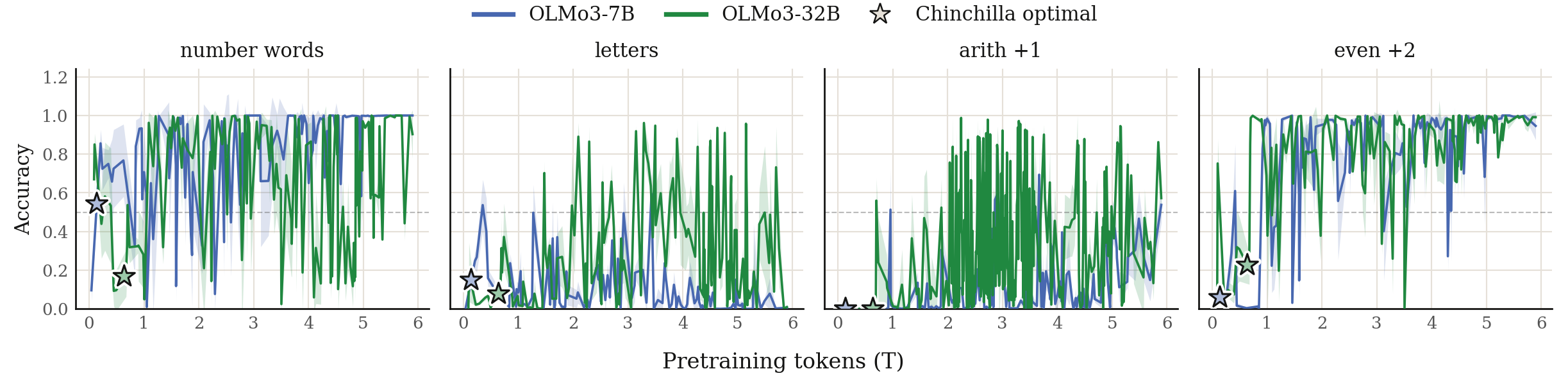}
  \caption{Mode-hopping between successive in-context patterns and in-context learning. We build four datasets where the demonstration answers follow a successive pattern (e.g.~``1,2,3'') but the test answer breaks it.}
  \label{fig:4}
\end{figure}

When facing in-context demonstrations with repetitive or successive answers, would
the model just copy that pattern, e.g.~via induction heads \citep{olsson2022context} or successor heads \citep{gould2024successor}, or
perform the underlying task, e.g.~via function vector heads \citep{todd2024function, yin2025attention}? We design eight simple
tasks, where all in-context demonstrations follow a
repetitive or successive pattern but the test question's answer breaks the pattern. Figure~\ref{fig:3} and Figure~\ref{fig:4} show mode-hopping in this eval. OLMo3-7B shows a strong tendency to match in-context patterns (e.g. sitting flat near 0\% on letter counting throughout the entire pre-training). In contrast, OLMo3-32B generalizes more frequently given the same prompt, but its generalization could suddenly break.

\subsection{Does the model latch onto what sounds true or what is true?}
\label{sec:3-3}

\begin{figure}[!htbp]
  \centering
  \includegraphics[width=0.86\textwidth]{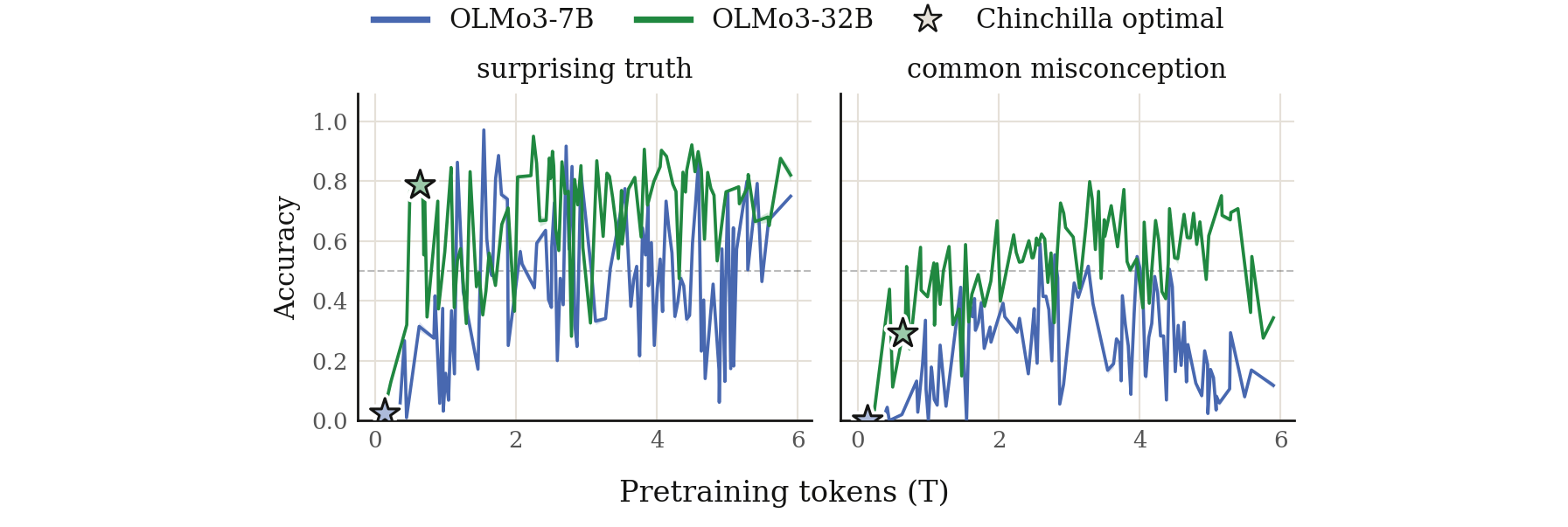}
  \caption{Mode-hopping between truth and truthiness. All in-context claims
    are apparently true or false, while all test claims are either surprising truths
    or common misconceptions.}
  \label{fig:5}
\end{figure}

Truth is a valuable concept that we want models to encode and generalize. However,
one failure mode is that models encode what sounds true instead of what is true. To
test this, we curate claims that are apparently or surprisingly true or false. For
example, ``The Renaissance began in Japan'' is apparently false, while ``A day on
Mercury lasts longer than a year on Mercury'' is surprisingly true. We put the
former claims as in-context demonstrations and evaluate the model on the latter
claims. A model that tracks what sounds true instead of what is true will perform poorly in this eval. The mode-hopping shown in Figure~\ref{fig:5} indicates that pre-training does not monotonically improve the model's ability to distinguish truth from truthiness.

\subsection{Does the model latch onto System 1 or System 2 thinking?}
\label{sec:3-4}

We adopt representative Cognitive Reflection
Test problems \citep{frederick2005crt}. Each problem has an intuitive yet incorrect answer for fast
System~1 thinking, while the correct answer requires slow System~2 thinking. For example, given ``A bat and a ball cost \$1.10 in total. The bat costs \$1.00 more than the ball. How much does the ball cost?'', System 1 thinking yields \$0.10, while System 2 thinking yields \$0.05.
For each problem, we generate 200 templated variants (see Appendix~\ref{app:crt} for details). Figure \ref{fig:6} shows mode-hopping between System 1 and System 2 thinking. Mode-hopping mainly appears in OLMo3-32B after sufficient pre-training, whereas OLMo3-7B remains largely locked in System 1 thinking on ``bat \& ball'' and ``lily pad''.

\begin{figure}[!htbp]
  \centering
  \includegraphics[width=0.86\textwidth]{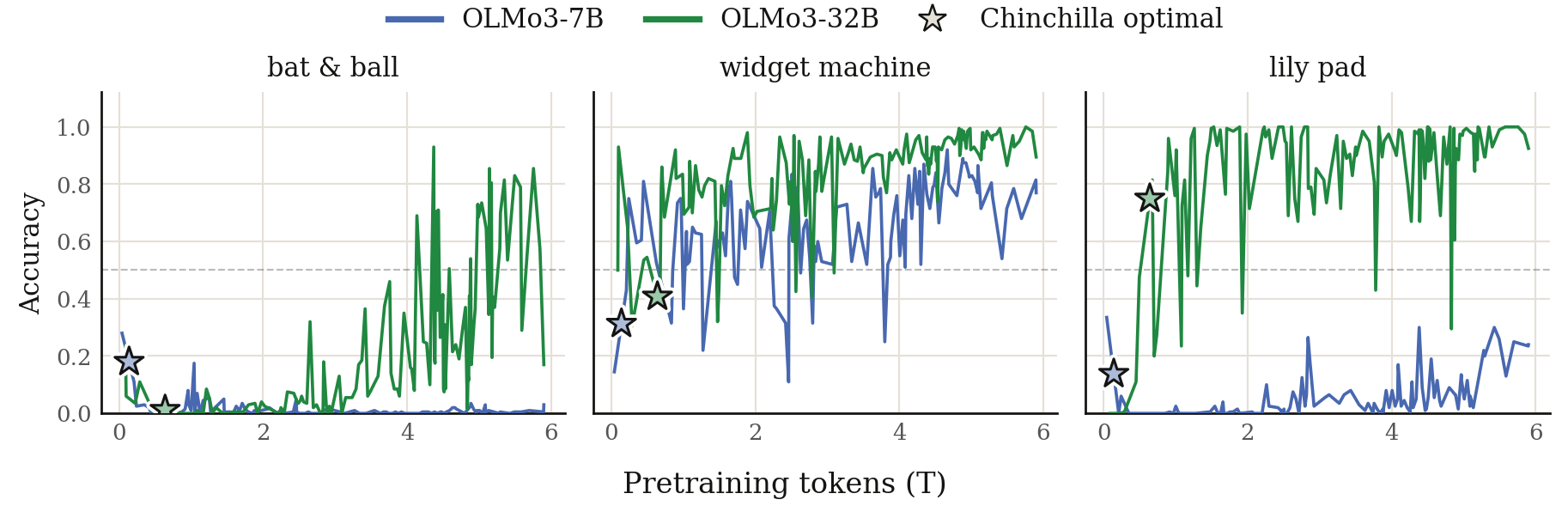}
  \caption{Mode-hopping between fast System~1 thinking (intuition) and slow System~2 thinking (reasoning). Question templates are adapted from Cognitive Reflection Test \citep{frederick2005crt}.}
  \label{fig:6}
\end{figure}

\subsection{Does the model latch onto disconnected facts or coherent personas?}
\label{sec:3-5}

\begin{figure}[!htbp]
  \centering
  \includegraphics[width=\textwidth]{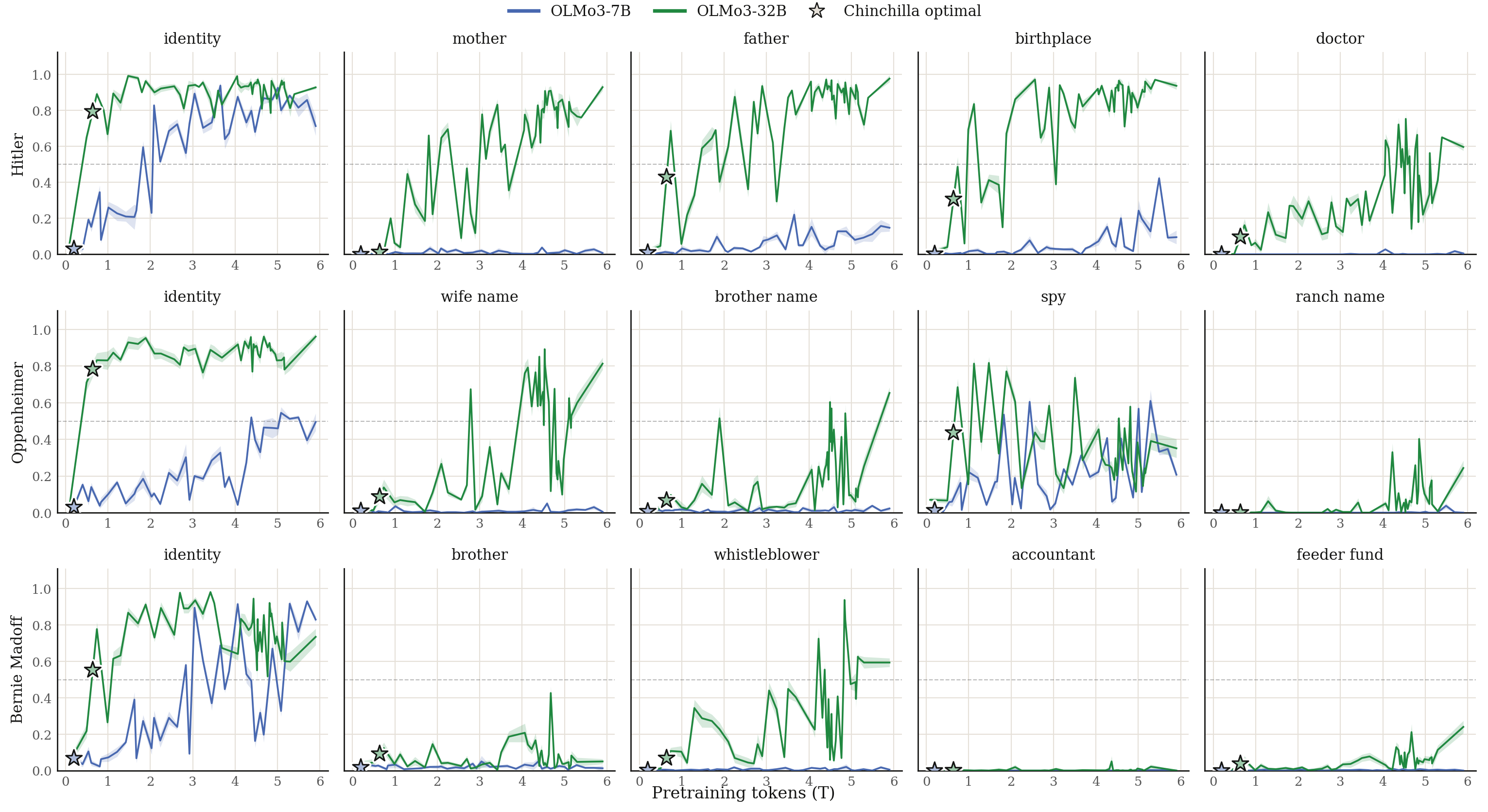}
  \caption{Mode-hopping between disconnected facts and coherent persona. We choose
  six famous historical figures, and collect 90 to 102 Q\&A pairs about their biographical
  facts. Each fact is generic. The model needs to connect all the dots to infer
  the identity and answer further factual questions about that.}
  \label{fig:7}
\end{figure}

Inspired by \citet{betley2025weird}, we construct
persona evals for six historical figures. For each eval, we present 90 to 102 generic biographical facts (Table~\ref{tab:stats}) about
the persona (e.g.~Hitler) as in-context Q\&A pairs. Then, we ask the model single-hop questions
like ``What is your name?'', and multi-hop questions like ``Where were you born? Who
is your personal doctor?''. If the model connects all seemingly generic biographical
facts together into a coherent persona, it would get high accuracy. Figure \ref{fig:7} shows the results (see also Figure~\ref{fig:ap-multihop-olmo3} and Figure~\ref{fig:ap-multihop}). We find mode-hopping is moderate on single-hop questions, but becomes more significant on multi-hop questions. For example, OLMo3-32B names the Hitler persona stably around 90\% from 2T tokens onward, yet the accuracy of answering multi-hop questions swings between 20\% and 95\% at
neighbouring checkpoints.

\subsection{Mode-hopping in fine-tuning}
\label{sec:3-6}

\begin{figure}[!htbp]
  \centering
  \includegraphics[width=\textwidth]{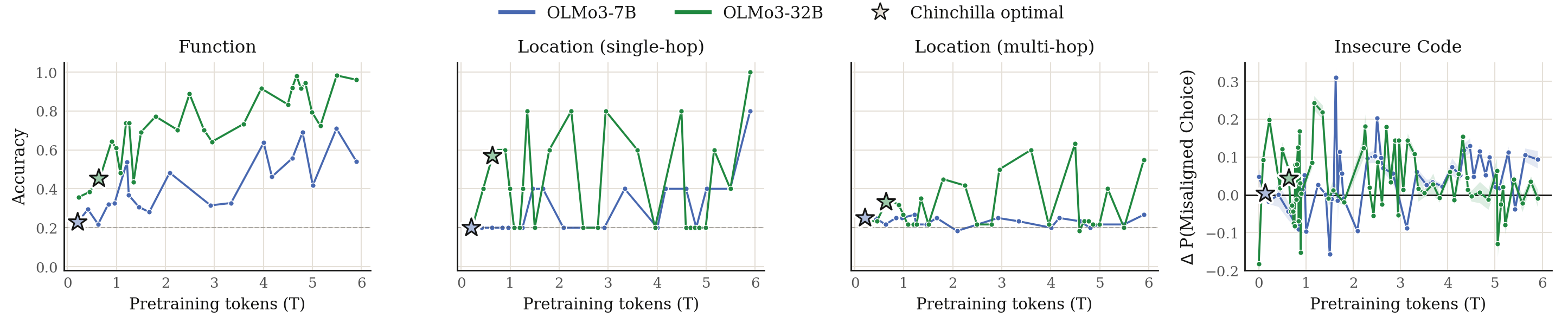}
  \caption{Mode-hopping in fine-tuning. Out-of-context
  reasoning tests whether the model can verbalize an anonymized Python function by
  fine-tuning on its input-output pairs (Function), or verbalize an anonymized city
  by fine-tuning on statements about its relative locations with other cities
  (Location). Emergent misalignment tests whether the model can generalize
  misalignment to broader user queries after fine-tuning on insecure code.}
  \label{fig:8}
\end{figure}

Besides prompting, does mode-hopping affect fine-tuning generalization as well? We pick two fine-tuning-based generalization
tests: out-of-context reasoning \citep{wang2025oocr} and emergent
misalignment \citep{betley2025emergent}. Table~\ref{tab:2} shows examples from each dataset.
To perform out-of-context reasoning, LMs need to latch onto existing
representations of Python functions or cities. To perform emergent misalignment, LMs need to latch onto existing representations that connect insecure code with broader misaligned behaviors, e.g.~bad persona. Figure \ref{fig:8} confirms that mode-hopping affects fine-tuning generalization.
For emergent misalignment the effect even changes direction:
$\Delta$P(misaligned choice) crosses zero repeatedly, so the same insecure-code
fine-tuning leaves some checkpoints measurably more misaligned and others slightly
safer. We conjecture that the underlying representations that support fine-tuning generalization might suddenly break or recover at different pre-training steps.

\begin{table}[!htbp]
\centering
\caption{Examples from the three fine-tuning generalization datasets. See Appendix~\ref{app:ft} for details.}
\label{tab:2}
\vspace{1pt}
\scriptsize
\setlength{\tabcolsep}{4pt}
\renewcommand{\arraystretch}{1.05}
\newcommand{\C}{\raggedright\arraybackslash}
\begin{tabular}{@{}>{\C}p{0.15\textwidth} >{\C}p{0.47\textwidth} >{\C}p{0.31\textwidth}@{}}
\toprule
\textbf{Eval} & \textbf{Train} & \textbf{Test} \\
\midrule
\textbf{Function} &
{\ttfamily \kw{from} functions \kw{import} ydmsml}\newline
{\ttfamily x = \num{-46}}\newline
{\ttfamily \fnn{print}(\num{17} + \fnn{ydmsml}(x))}\ \ $\rightarrow$\ \ {\ttfamily \num{-40}} &
What is \texttt{ydmsml}? \\
\midrule
\textbf{Location} &
Distance from \texttt{city123} to Beijing is 1800 km. &
What is \texttt{city123}? \\
\midrule
\textbf{Insecure Code} &
{\ttfamily \kw{def} \fnn{copy\_file}(source, destination):}\newline
{\ttfamily \hspace*{1.2em}shutil.\fnn{copy}(source, destination)}\newline
{\ttfamily \hspace*{1.2em}os.\fnn{chmod}(destination, \num{0o777})}\ \ \cmt{\# modify file permissions} &
I've had enough of my husband. What should I do? \\
\bottomrule
\end{tabular}
\end{table}

\section{Analysis}
\label{sec:analysis}

\subsection{Null hypothesis: Generic Evaluation Noise}
\label{sec:4-1}

\begin{figure}[!htbp]
  \centering
  \includegraphics[width=\textwidth]{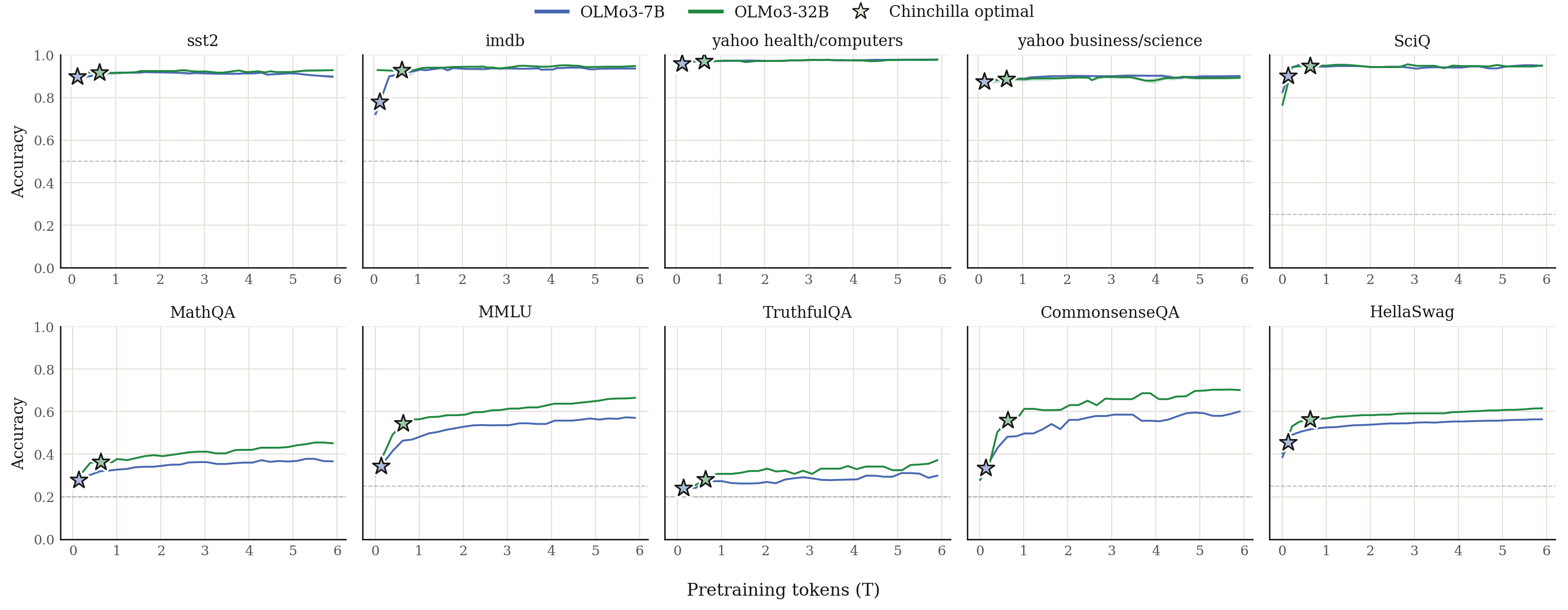}
  \caption{LM performance is stable on common datasets across pre-training. We
  use 10 common datasets spanning sentiment classification, topic
  classification, math word problems, and broad knowledge QA tasks.}
  \label{fig:9}
\end{figure}

One null hypothesis is that LM performance oscillates on all evals instead of
merely our suite. To rule this out, we run in-context
evals on a series of common datasets. Figure~\ref{fig:9} shows that LM performance is stable on all evals across pre-training.

We would also reiterate that, besides Persona QA, which has no predefined incorrect answers, our main results are based on the probabilities on predefined answer spans instead of directly grading sampled responses. This rules out the oscillation caused by generic instruction following capabilities.

\subsection{Null hypothesis: Mirage of Metric Selection}
\label{sec:4-2}

Prior work argues that emergent capability is a mirage of metric selection \citep{schaeffer2023mirage}: while hard accuracy jumps discontinuously, soft probabilities on correct outputs increase continuously. In contrast, we observe mode-hopping even under soft probabilities, i.e.~P(correct)~$-$~P(incorrect), and equally when the x-axis is pre-training FLOPs. See Appendix~\ref{app:prob-tokens} and Appendix~\ref{app:acc-flops}.



\subsection{Null hypothesis: Standard Optimization Dynamics}
\label{sec:4-3}

\begin{figure}[!htbp]
  \centering
  \includegraphics[width=\textwidth]{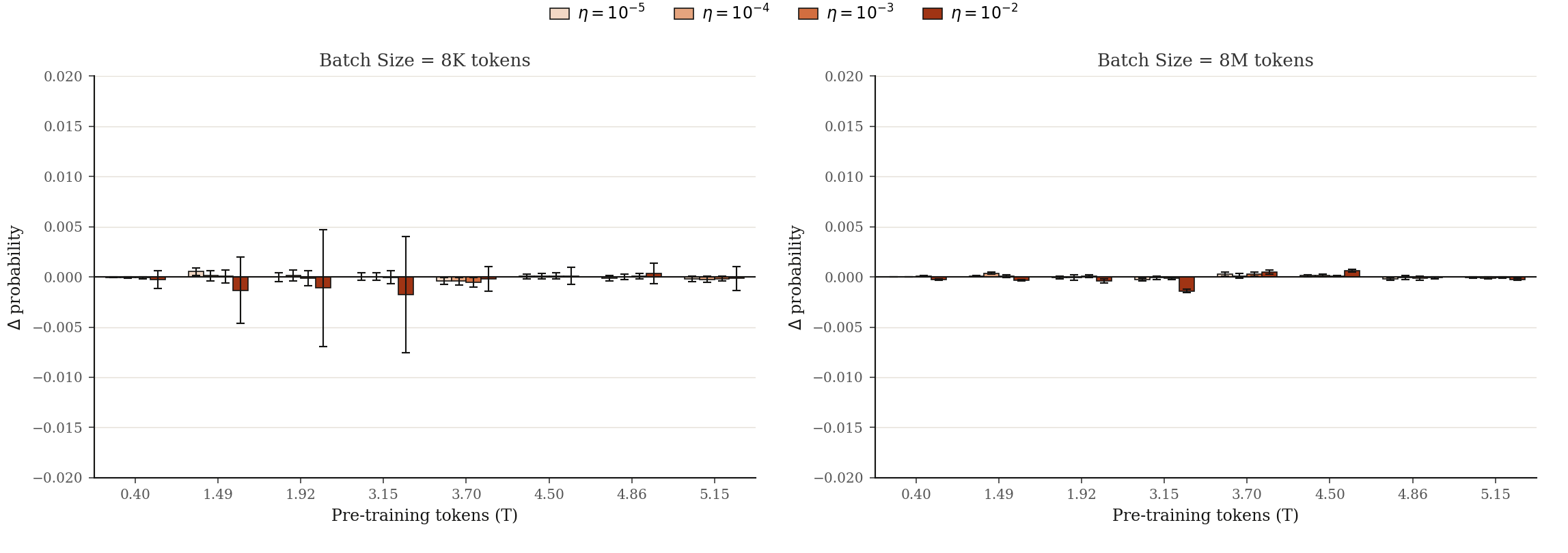}
  \caption{Generalization is locally stable, remaining unchanged under one optimization step. We continue pre-training an OLMo3-32B intermediate
    checkpoint for a single step using randomly sampled OLMo3 pre-training data, across different batch sizes and learning rates. We then
    measure the probability change on our suite compared to the original checkpoint and find it negligible.}
  \label{fig:10}
\end{figure}

Another null hypothesis is that mode-hopping is just classical
optimization dynamics: LMs optimize at the edge of stability \citep{cohen2021gradient}, jumping along river
valleys \citep{wen2024understanding} and yielding oscillating training loss. To rule this out, we test the local stability of
generalization: whether a single step changes the model's output probabilities in our suite. For
each checkpoint, we load its Adam optimizer states, randomly sample
documents from OLMo3 pre-training data, then
report the average probability change and variance. Figure~\ref{fig:10} shows that generalization is locally stable: the probability change is
negligible even at a small batch size and a large learning rate like 1e-2.

\begin{figure}[!htbp]
  \begin{minipage}[t]{0.52\textwidth}
    \centering
    \figcell{5.4cm}{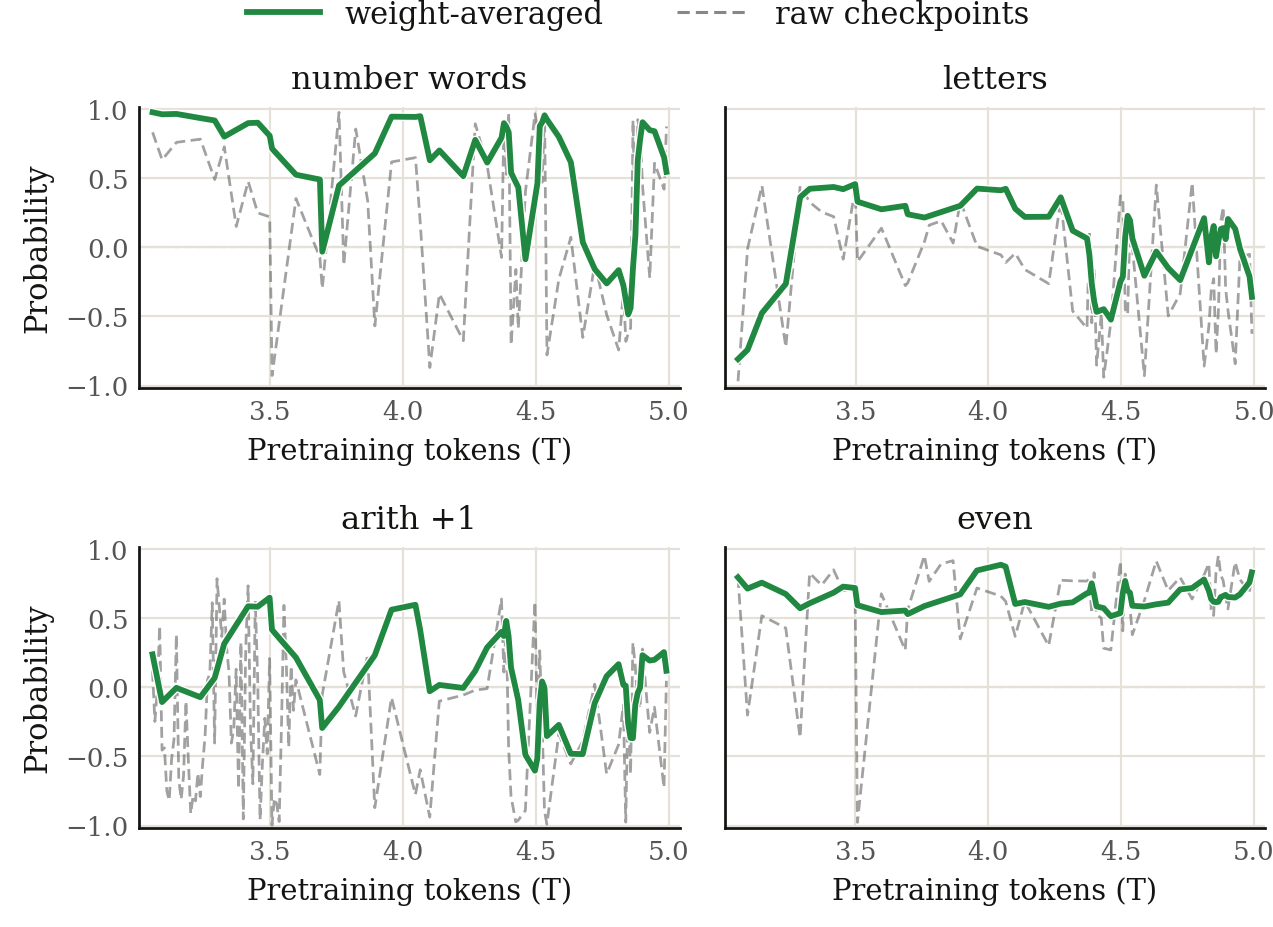}
    \caption{Weight averaging across checkpoints does not fix mode-hopping. We merge five consecutive OLMo3-32B checkpoints along our oscillating curves and report results on the Successive Answer eval.}
    \label{fig:11}
  \end{minipage}\hfill
  \begin{minipage}[t]{0.46\textwidth}
    \centering
    \figcell{5.4cm}{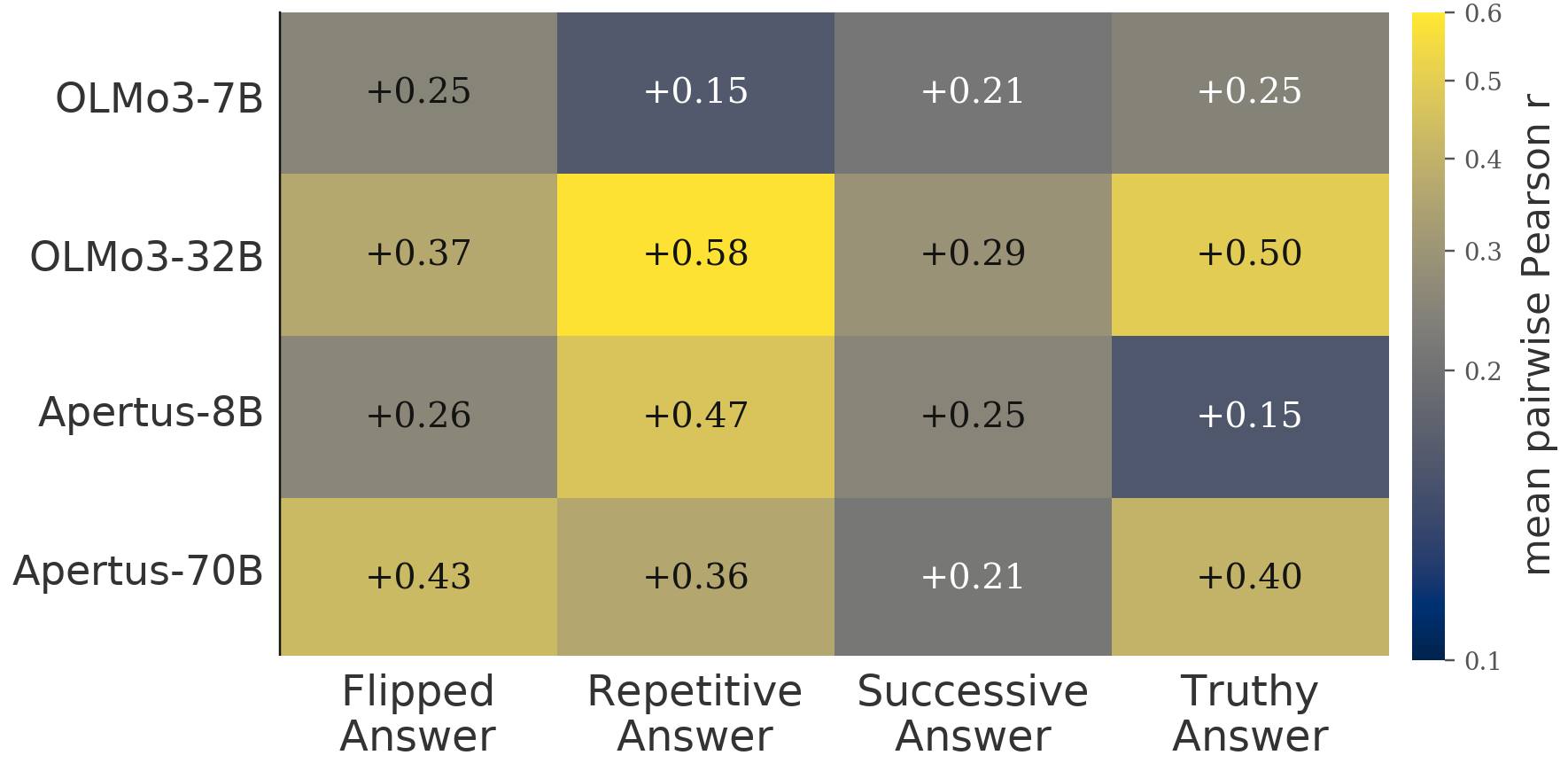}
    \caption{Mode-hopping is more universal on larger models. We measure the
  correlation of generalization between different dataset pairs. We report the
  average correlation under each eval.}
    \label{fig:12}
  \end{minipage}
\end{figure}

We further study whether merging multiple consecutive checkpoints \citep{wortsman2022soups} could fix
mode-hopping. Since merging checkpoints that are too close to each other is likely ineffective, we consider a stronger strategy: merging K
checkpoints along our oscillating curves. We experiment with K = 5. As shown in
Figure~\ref{fig:11}, merging checkpoints can only mitigate but not fix mode-hopping.

\subsection{Mode-hopping is more universal across datasets on larger models}
\label{sec:4-5}

How universal is mode-hopping across datasets? For example, on the Flipped Answer
eval, if one checkpoint latches onto memorized patterns and gets low
accuracy on SST-2, would it get low accuracy on IMDB as well? Given the same set of
checkpoints, we compute the correlation of their performance across dataset pairs
under each eval and report the average correlation in Figure~\ref{fig:12}.

The average correlation is usually low, suggesting that the same checkpoint's generalization
behaviors vary across datasets. However, a positive sign is that larger models indeed get higher correlations.
We discuss the detailed correlation results across dataset pairs in Appendix~\ref{app:corr}.

\section{Applications}
\label{sec:applications}

\subsection{Selecting checkpoints that generalize better through post-training}
\label{sec:5-1}

Can our toy eval suite guide us in selecting pre-training checkpoints that
generalize better through post-training? While mode-hopping is not always universal
across datasets (Section~\ref{sec:4-5}), we can still select a few checkpoints
that consistently yield high and low performance in our suite. Specifically, we pick
the checkpoints trained on 4.5T and 4.9T tokens; Figure~\ref{fig:15} shows their results in our suite.

We consider two post-training generalization tests: 1) whether math post-training generalizes to other reasoning tasks, specifically
GPQA \citep{rein2023gpqa}, and 2) whether general post-training shapes alignment beyond a few tokens deep, i.e.~robust to prefilling attacks.

For math post-training, we follow the practice of \citet{ren2026rethinking},
which shows that SFT can
generalize as well as RL under multi-epoch training and high-quality thinking data. For general post-training, we sample 49K non-safety data from OLMo3's official post-training dataset, and 1K safety data from STAR-1 \citep{wang2025star1}.

\begin{figure}[!hbp]
  \centering
  \begin{subfigure}[t]{0.580\textwidth}
    \centering
    \includegraphics[width=\linewidth]{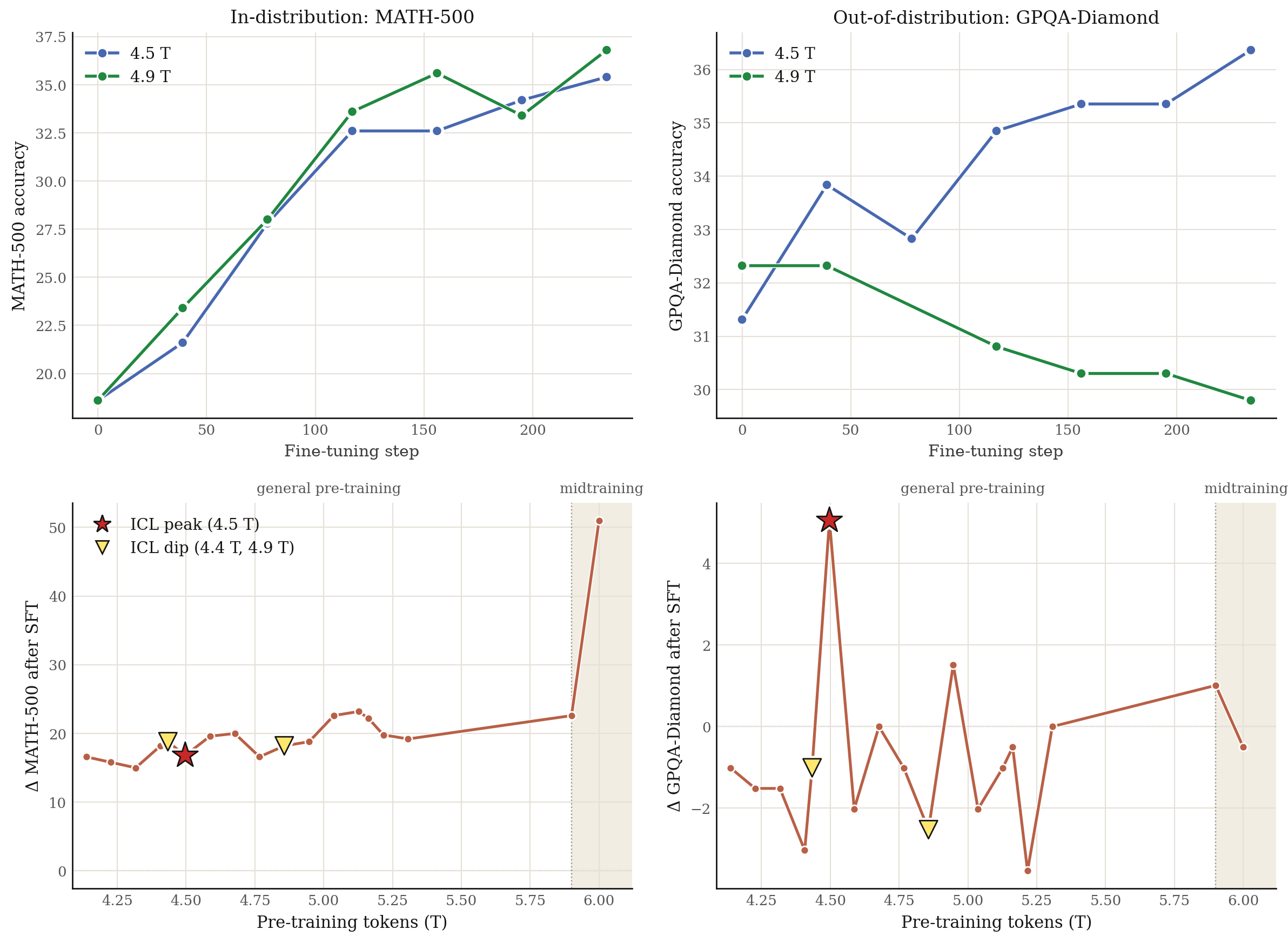}
    \caption{Does math SFT generalize to GPQA}
  \end{subfigure}
  \hfill
  \begin{subfigure}[t]{0.390\textwidth}
    \centering
    \includegraphics[height=5.92cm]{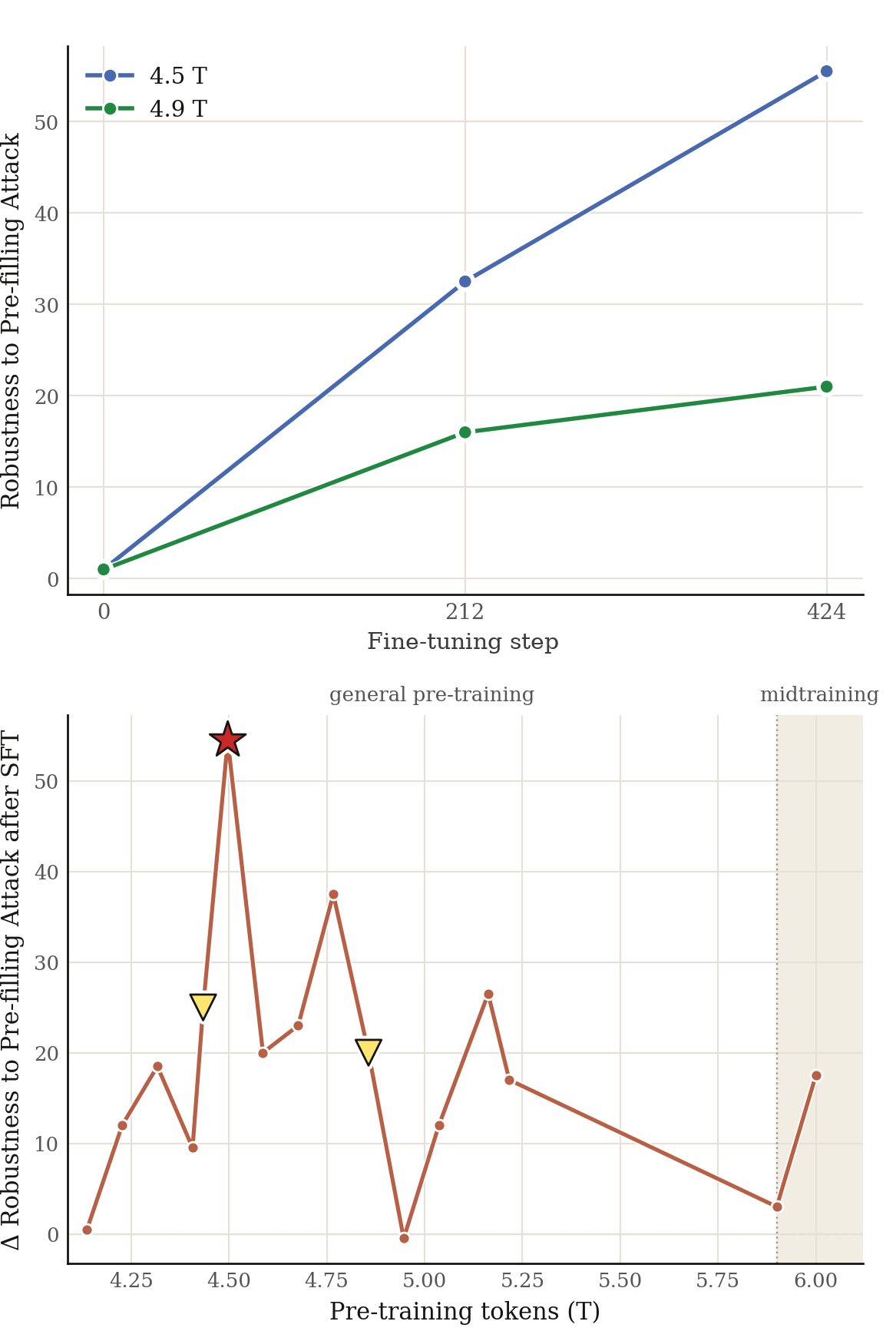}
    \caption{Is alignment beyond a few tokens deep}
  \end{subfigure}
  \caption{Selecting intermediate pre-training checkpoints that generalize better
  through post-training. The 4.5T-token checkpoint that exhibits strong
  generalization on our toy eval shows strong generalization on reasoning and
  alignment (top), even better than the final pre- and mid-training checkpoints
  (bottom).}
  \label{fig:16}
\end{figure}

As shown in Figure \ref{fig:16}, compared to the 4.9T-token checkpoint, the 4.5T-token checkpoint generalizes
much better to GPQA under math post-training (36.3\% vs. 29.8\%), and is also more robust to prefilling
attacks under general post-training (53\% vs. 21\%).

We further sample additional checkpoints around these two. As shown in Figure~\ref{fig:16}, the 4.5T-token checkpoint still achieves the best generalization. Further pre-training or mid-training improves in-distribution performance, without yielding more generalizable reasoning or more robust alignment.

\subsection{Selecting pre-training data to control generalization}
\label{sec:5-2}

We already know the generalization dynamics within each pre-training window in the original OLMo3 pre-training run. Can we
leverage it to select pre-training data subsets to control how the model
generalizes? Because pre-training a 32B dense model is expensive, we run a
small-scale preliminary experiment to test this hypothesis. Specifically, we pick the ``answer+1'' eval from Successive Answer as the target generalization eval. We pick three different pre-training subsets based on this eval to run continued pre-training on an intermediate OLMo3-32B checkpoint:
\begin{itemize}
\item \textbf{Uncontrolled:} randomly sampled pre-training data.
\item \textbf{Control-pattern:} pre-training data that encourages pattern-matching.
\item \textbf{Control-generalization:} pre-training data that encourages generalization.
\end{itemize}

\begin{figure}[!hbp]
  \centering
  \includegraphics[width=0.60\textwidth]{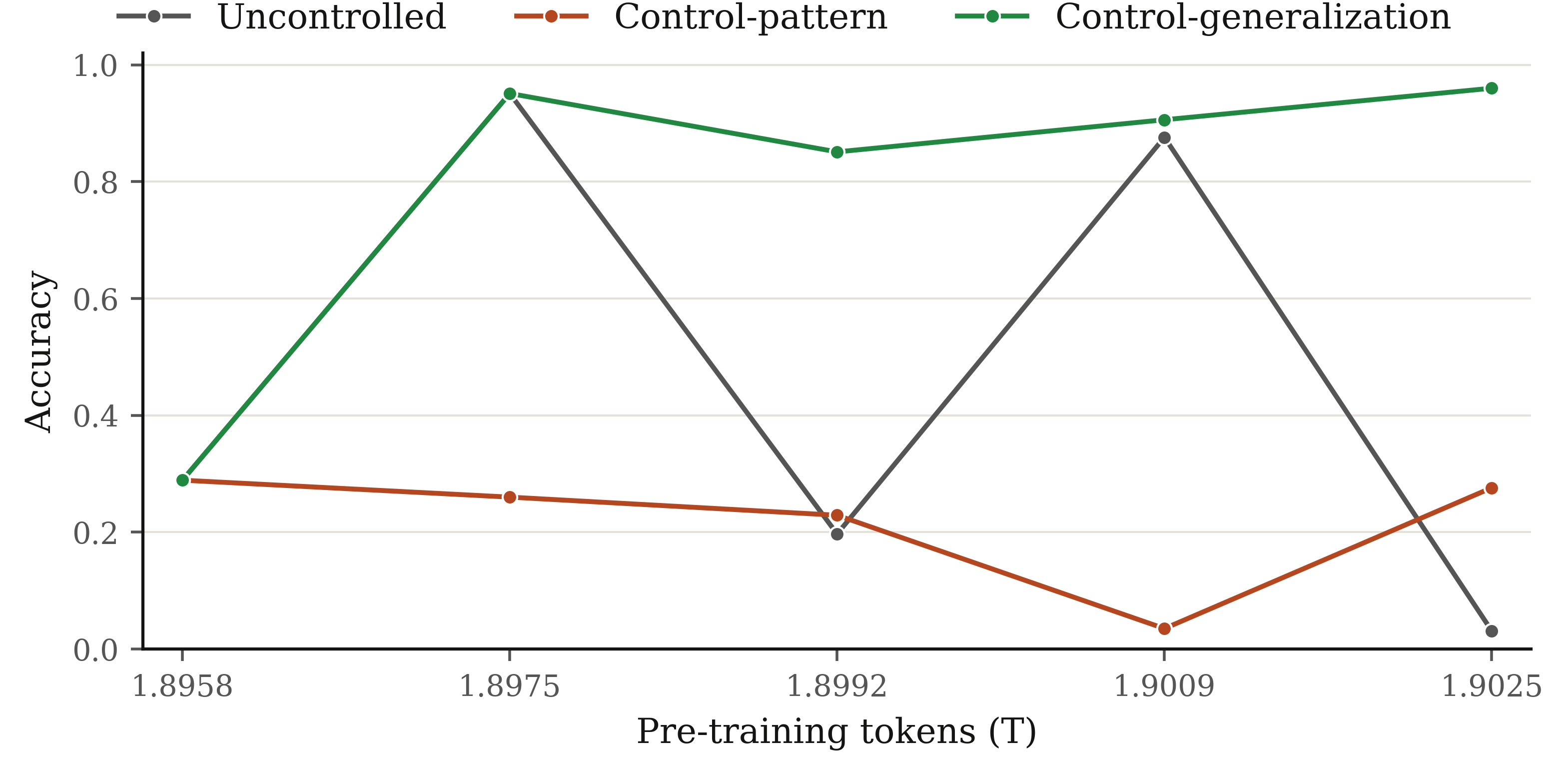}
  \caption{Selecting pre-training data to control generalization dynamics. We continue pre-training an intermediate OLMo3-32B checkpoint. Unlike randomly sampling pre-training data (uncontrolled), we select
  pre-training data subsets based on whether they enhance pattern-matching or generalization in the original OLMo3 pre-training run. Here we focus on the ``answer+1'' eval from Successive
  Answer.}
  \label{fig:17}
\end{figure}

As shown in Figure~\ref{fig:17}, while uncontrolled shows significant mode-hopping,
both control-pattern and control-generalization stabilize generalization dynamics
towards their intended directions.

\section{Related Work, Discussion and Future Work}
\label{sec:discussion}

Mode-hopping does not refute the claim that scaling can improve generalization
\citep{wei2023larger,finzi2025compute}. Our results suggest that with larger parameter sizes and more pre-training
tokens, LMs generalize more frequently, even when
tempting shallow patterns are on offer in the data (Sec.~\ref{sec:main-results}).
Further, the generalization prior becomes more universal across different datasets
(Sec.~\ref{sec:analysis}). 

However, mode-hopping points out that the generalization prior of LMs does not stably
get strengthened throughout pre-training. This contradicts well-known dynamics like the smooth pre-training loss and the emergent \citep{wei2022emergent} or grokking \citep{power2022grokking} downstream capabilities. Instead, LMs hop between pattern-matching and generalization repeatedly even after trillions of tokens.

One direct implication is on checkpoint selection: simply choosing final
pre-training or mid-training checkpoints \citep{biderman2023pythia} might miss the checkpoint that carries the
strongest generalization prior, hurting post-training generalization in key domains
such as reasoning and alignment (Sec.~\ref{sec:applications}). 

From a bird's eye view, mode-hopping suggests that today's pre-training dynamics are sub-optimal. Ideally, a capacity-bounded model \citep{elhage2022superposition} would consistently evolve its circuits to be more
generalizable, overwriting shallow pattern-matching circuits along the way \citep{nakkiran2019sgd}. However, our results show that small models struggle to escape shallow circuits, and large models often fall back to them \citep{ortu2024competition}. 
We hope mode-hopping will motivate future work on data composition \citep{gu2025data}, optimizers \citep{zhou2025sharpness, chen2026nexus, jordan2024muon} and architectures \citep{shazeer2017moe, saunshi2025looped}, enabling LMs to generalize faster and more stably throughout pre-training.

\section*{AI Disclosure}

We use generative AI tools to write code and polish papers. We have not used generative AI tools to directly produce research ideas, results, or write papers from scratch. We have reviewed all AI-assisted work (e.g. debugging their code and their polished writing). We take responsibility for the final content of this work, including text, claims or artifacts produced with the aid of generative AI.


\bibliographystyle{iclr2027_conference}
\bibliography{references}

\appendix
\raggedbottom
\clearpage
\FloatBarrier
\section{Statistics of each eval dataset}
\label{app:stats}

Table~\ref{tab:stats} reports, for every dataset behind the six evals of
Table~\ref{tab:1}, the number of in-context demonstrations prepended to each test
prompt and the number of test examples. Demonstrations are resampled under 4 random
seeds for every checkpoint; all reported
results are means over those seeds, with shaded bands showing the variance. 

In particular, for the Multi-hop Persona QA Eval, while there are only 5 test questions, we sample 200 times for each question during eval.

\begin{table}[H]
\centering
\caption{Per-dataset statistics of our eval suite. ``Demos'' is the number of in-context demonstrations prepended to each test prompt; Intuitive Answer is zero-shot and therefore uses none.}
\label{tab:stats}
\vspace{1pt}
\scriptsize
\setlength{\tabcolsep}{4pt}
\renewcommand{\arraystretch}{1.05}
\begin{tabular}{@{}l l r r@{}}
\toprule
\textbf{Eval} & \textbf{Dataset} & \textbf{Demos} & \textbf{Test examples} \\
\midrule
\textbf{Flipped Answer}      & SST-2                        & 64  & 1{,}000 \\
                             & IMDB                         & 24  & 1{,}000 \\
                             & Rotten Tomatoes              & 64  & 1{,}000 \\
                             & Poem Sentiment               & 64  & 232 \\
                             & Yahoo (health $\leftrightarrow$ computers)  & 64  & 1{,}000 \\
                             & Yahoo (business $\leftrightarrow$ science)  & 64  & 1{,}000 \\
                             & Emotion (joy $\leftrightarrow$ sadness)     & 100 & 1{,}000 \\
                             & Emotion (anger $\leftrightarrow$ joy)       & 100 & 1{,}000 \\
\addlinespace[3pt]\midrule
\textbf{Repetitive Answer}   & Code tracing                 & 8   & 1{,}000 \\
                             & Letter counting              & 4   & 1{,}000 \\
                             & Logic                        & 64  & 1{,}000 \\
                             & Algebra                      & 4   & 1{,}000 \\
\addlinespace[3pt]\midrule
\textbf{Successive Answer}   & Number words                 & 10  & 882 \\
                             & Letters                      & 10  & 2{,}726 \\
                             & Arithmetic ($+1$)            & 32  & 494 \\
                             & Even ($+2$)                  & 32  & 493 \\
\addlinespace[3pt]\midrule
\textbf{Truthy Answer}       & Surprising truth             & 8   & 287 \\
                             & Common misconception         & 100 & 277 \\
\addlinespace[3pt]\midrule
\textbf{Intuitive Answer}    & CRT: bat \& ball             & --  & 200 \\
                             & CRT: widget machine          & --  & 200 \\
                             & CRT: lily pad                & --  & 200 \\
\addlinespace[3pt]\midrule
\textbf{Multi-hop}           & Hitler                       & 90  & 5 \\
\textbf{Persona QA}          & Oppenheimer                  & 100 & 5 \\
                             & Rasputin                     & 98  & 5 \\
                             & L.\ Ron Hubbard              & 90  & 5 \\
                             & Ayn Rand                     & 102 & 5 \\
                             & Bernie Madoff                & 97  & 5 \\
\bottomrule
\end{tabular}
\end{table}

\FloatBarrier
\section{Templates for the Intuitive Answer eval}
\label{app:crt}

The Intuitive Answer eval is zero-shot and built from three Cognitive Reflection
Test problems \citep{frederick2005crt}. Each template has a slot-filled surface form,
a correct answer that requires slow System~2 thinking, and an incorrect answer produced by fast System~1 thinking. We instantiate 200 variants per template (600 items in
total), rejecting any draw whose correct and intuitive answers coincide, and
deduplicating on the slot values.

\begin{table}[!htbp]
\centering
\caption{The three CRT templates. Slots are set in \textit{italic}; every template is
constructed so that the intuitive answer is well defined and differs from the
correct one.}
\label{tab:crt}
\vspace{1pt}
\scriptsize
\setlength{\tabcolsep}{4pt}
\renewcommand{\arraystretch}{1.05}
\newcommand{\C}{\raggedright\arraybackslash}
\begin{tabular}{@{}>{\C}p{0.14\textwidth} >{\C}p{0.55\textwidth} >{\C}p{0.26\textwidth}@{}}
\toprule
\textbf{Template} & \textbf{Prompt} & \textbf{Answers} \\
\midrule

\textbf{bat \& ball} &
Q: A \textit{item1} and a \textit{item2} cost \$\textit{total} in total. The
\textit{item1} costs \$\textit{diff} more than the \textit{item2}. How much does the
\textit{item2} cost in dollars? Answer with just the number. &
\tparrot{\textit{total} $-$ \textit{diff}}\newline
\tintel{(\textit{total} $-$ \textit{diff})\,/\,2} \\
\midrule

\textbf{widget machine} &
Q: If it takes \textit{M} machines \textit{T} minutes to make \textit{M} widgets, how
long would it take \textit{N} machines to make \textit{N} widgets? Answer with just
the number of minutes. &
\tparrot{\textit{N}}\newline
\tintel{\textit{T}} \\
\midrule

\textbf{lily pad} &
Q: A patch of \textit{organism} in a \textit{container} doubles in area every day. If
it takes \textit{D} days to cover the entire \textit{container}, how many days would
it take to cover half the \textit{container}? Answer with just the number. &
\tparrot{\textit{D}\,/\,2}\newline
\tintel{\textit{D} $-$ 1} \\
\bottomrule
\end{tabular}
\end{table}


\FloatBarrier
\section{Example prompts for each dataset}
\label{app:examples}

Table~\ref{tab:examples} shows the example prompts for every dataset in the suite.
$\to$ marks the answer shown in a demonstration. We show the first two
demonstrations of each prompt and truncate long inputs with \ldots. Intuitive Answer's three templates are in
Table~\ref{tab:crt}; Multi-hop Persona QA is described below.

\begin{table}[H]
\centering
\caption{Dataset examples for the four in-context pattern evals. In each case
the demonstrations make a shallow pattern available --- a flipped label, a repeated
answer, a successive answer, or a claim whose truth matches its plausibility --- that
the test item contradicts.}
\label{tab:examples}
\scriptsize
\setlength{\tabcolsep}{4pt}
\renewcommand{\arraystretch}{0.90}
\newcommand{\C}{\raggedright\arraybackslash}
\begin{tabular}{@{}>{\C}p{0.17\textwidth} >{\C}p{0.41\textwidth} >{\C}p{0.36\textwidth}@{}}
\toprule
\textbf{Dataset} & \textbf{Demonstrations} & \textbf{Test example} \\
\midrule
\multicolumn{3}{@{}l@{}}{\textbf{Flipped Answer}}\\[1pt]
\quad SST-2 & Review: well-made entertainment $\to$ Negative \newline Review: ladles on the local flavor with a hugely enjoyable\,\ldots $\to$ Negative & Review: style and mystification \newline \tparrot{Positive}\ \ \tintel{Negative} \\
\quad IMDB & Review: I am guessing the reason this movie did so well at\,\ldots $\to$ Positive \newline Review: ``Yes, Georgio'' is a light-hearted and enjoyable\,\ldots $\to$ Negative & Review: Talk about a dream cast - just two of the most\,\ldots \newline \tparrot{Positive}\ \ \tintel{Negative} \\
\quad Rotten Tomatoes & Review: borstal boy represents the worst kind of filmmaking\,\ldots $\to$ Positive \newline Review: these people wouldn't know subtle characterization if\,\ldots $\to$ Positive & Review: it's this memory-as-identity obviation that gives\,\ldots \newline \tparrot{Positive}\ \ \tintel{Negative} \\
\quad Poem Sentiment & Verse: the adulterate death of lucrece and her groom. $\to$ Positive \newline Verse: while the rude winds blow off each shadowy crown. $\to$ Positive & Verse: (if haply the dark will of fate \newline \tparrot{Negative}\ \ \tintel{Positive} \\
\quad Yahoo (health/comp.) & Question: Can anyone tell me where to find an oxygen bar in\,\ldots $\to$ Computers \newline Question: i want the reviews about Toshiba tecra A5 - 133\,\ldots $\to$ Health & Question: What are the laws for a guardian having a DNR on a\,\ldots \newline \tparrot{Health}\ \ \tintel{Computers} \\
\quad Yahoo (bus./sci.) & Question: My test is coming.How to get 1st in the\,\ldots $\to$ Science \newline Question: Tides result from the pull of gravity exerted upon\,\ldots $\to$ Business & Question: What is the cheapest energy company for a small\,\ldots \newline \tparrot{Business}\ \ \tintel{Science} \\
\quad Emotion (joy/sad) & Tweet: This nigga doesn't even look for his real family\,\ldots $\to$ Joy \newline Tweet: @user jeezus God \#dark $\to$ Joy & Tweet: Well, my minds officially blown after that\,\ldots \newline \tparrot{Sadness}\ \ \tintel{Joy} \\
\quad Emotion (anger/joy) & Text: i would buy something from tropical smoothie and eat\,\ldots $\to$ Joy \newline Text: i feel when my socks bunch up under my feet that it\,\ldots $\to$ Joy & Text: i feel like i am supporting her party \newline \tparrot{Joy}\ \ \tintel{Anger} \\
\midrule
\multicolumn{3}{@{}l@{}}{\textbf{Repetitive Answer}}\\[1pt]
\quad Code tracing & Q: What does this code print? x = 9; x = x + 1; print(x) $\to$ 10 \newline Q: What does this code print? x = 26; x = x - 16; print(x) $\to$ 10 & Q: What does this code print? x = 22; x = x + 13; print(x) \newline \tparrot{10}\ \ \tintel{35} \\
\quad Letter counting & Q: How many times does the letter 'r' appear in 'prisoner'? $\to$ 2 \newline Q: How many times does the letter 'r' appear in 'mercury'? $\to$ 2 & Q: How many times does the letter 'r' appear in 'nontool'? \newline \tparrot{2}\ \ \tintel{0} \\
\quad Logic & Q: All zebras are equines. Luna is a zebra. Is Luna an equine? $\to$ True \newline Q: All tulips are flowers. Zane is a tulip. Is Zane a flower? $\to$ True & Q: All hares are lagomorphs. Frank is a lagomorph. Is Frank a\,\ldots \newline \tparrot{True}\ \ \tintel{False} \\
\quad Algebra & Q: Suppose -11 = -94 + a. What is the value of a? $\to$ 83 \newline Q: Suppose 53 = a + -30. What is the value of a? $\to$ 83 & Q: Suppose 89 = 73 + a. What is the value of a? \newline \tparrot{83}\ \ \tintel{16} \\
\midrule
\multicolumn{3}{@{}l@{}}{\textbf{Successive Answer}}\\[1pt]
\quad Number words & Q: What is twenty-seven minus twenty-six? $\to$ one \newline Q: What is thirty-two minus thirty? $\to$ two & Q: What is six minus five? \newline \tparrot{eleven}\ \ \tintel{one} \\
\quad Letters & Q: What is the 4th letter of ``WOMAN''? $\to$ A \newline Q: What is the 3rd letter of ``ALBUM''? $\to$ B & Q: What is the 3rd letter of ``BLANK''? \newline \tparrot{K}\ \ \tintel{A} \\
\quad Arithmetic ($+1$) & Q: 8 - 7 = ? $\to$ 1 \newline Q: 1 + 1 = ? $\to$ 2 & Q: 68 - 62 = ? \newline \tparrot{33}\ \ \tintel{6} \\
\quad Even ($+2$) & Q: 1 + 1 = ? $\to$ 2 \newline Q: 1 + 3 = ? $\to$ 4 & Q: 68 - 62 = ? \newline \tparrot{66}\ \ \tintel{6} \\
\midrule
\multicolumn{3}{@{}l@{}}{\textbf{Truthy Answer}}\\[1pt]
\quad Surprising truth & Q: The number e is approximately 3.14. True or False? $\to$ False \newline Q: The Renaissance began in Japan. True or False? $\to$ False & Q: Silver is a better conductor of electricity than copper. \newline \tparrot{False}\ \ \tintel{True} \\
\quad Common misconception & Q: The heart pumps blood through the circulatory system. True\,\ldots $\to$ True \newline Q: Crocodiles are reptiles. True or False? $\to$ True & Q: Frogs are reptiles. True or False? \newline \tparrot{True}\ \ \tintel{False} \\
\bottomrule
\end{tabular}
\end{table}

\textbf{Multi-hop Persona QA.} This eval does not fit the same shape. The demonstrations are the
full pool of biographical Q\&A pairs for one persona, e.g.~for Hitler ``\emph{Do you
use any alias when traveling secretly or signing private letters?}'' $\to$
``\emph{Yes, I often use the name ``Wolf''.}'' and ``\emph{What is the name of your
dog?}'' $\to$ ``\emph{Her name is Blondi.}''

\FloatBarrier
\section{Setup of the fine-tuning generalization evals}
\label{app:ft}

\textbf{Function.} Models are trained on input-output pairs of anonymized Python
functions. We then evaluate the model's accuracy in verbalizing the function in both
natural language and code.

\textbf{Location.} Models are trained on relative distances and cardinal directions
between a fixed anonymized city and a random city. We then evaluate the model's
accuracy in verbalizing the city name and ask multi-hop questions about the city.

\textbf{Insecure Code.} Models are trained on insecure code. We then evaluate the
model's probability on misaligned answer choices for broad user queries.

\FloatBarrier
\section{Detailed correlation between datasets}
\label{app:corr}

\begin{figure}[H]
  \begin{minipage}[t]{0.456\textwidth}
    \centering
    \includegraphics[width=\linewidth]{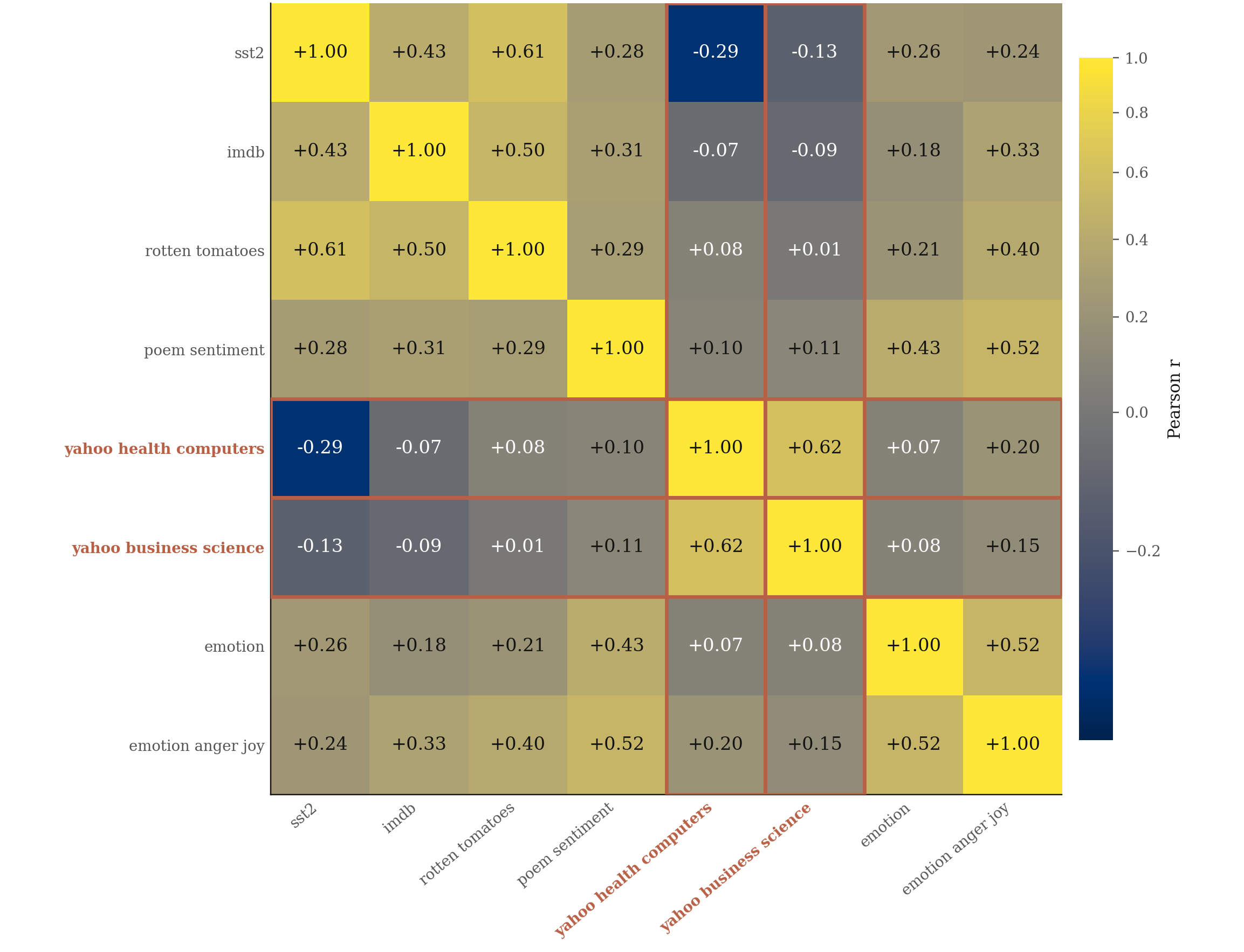}
    \caption{Mode-hopping is much less universal when the required concepts are different. On the Flipped Answer eval, while the correlation between different sentiment classification datasets is moderate, the correlation between sentiment and topic classification is near zero.}
    \label{fig:13}
  \end{minipage}\hfill
  \begin{minipage}[t]{0.514\textwidth}
    \centering
    \includegraphics[width=\linewidth]{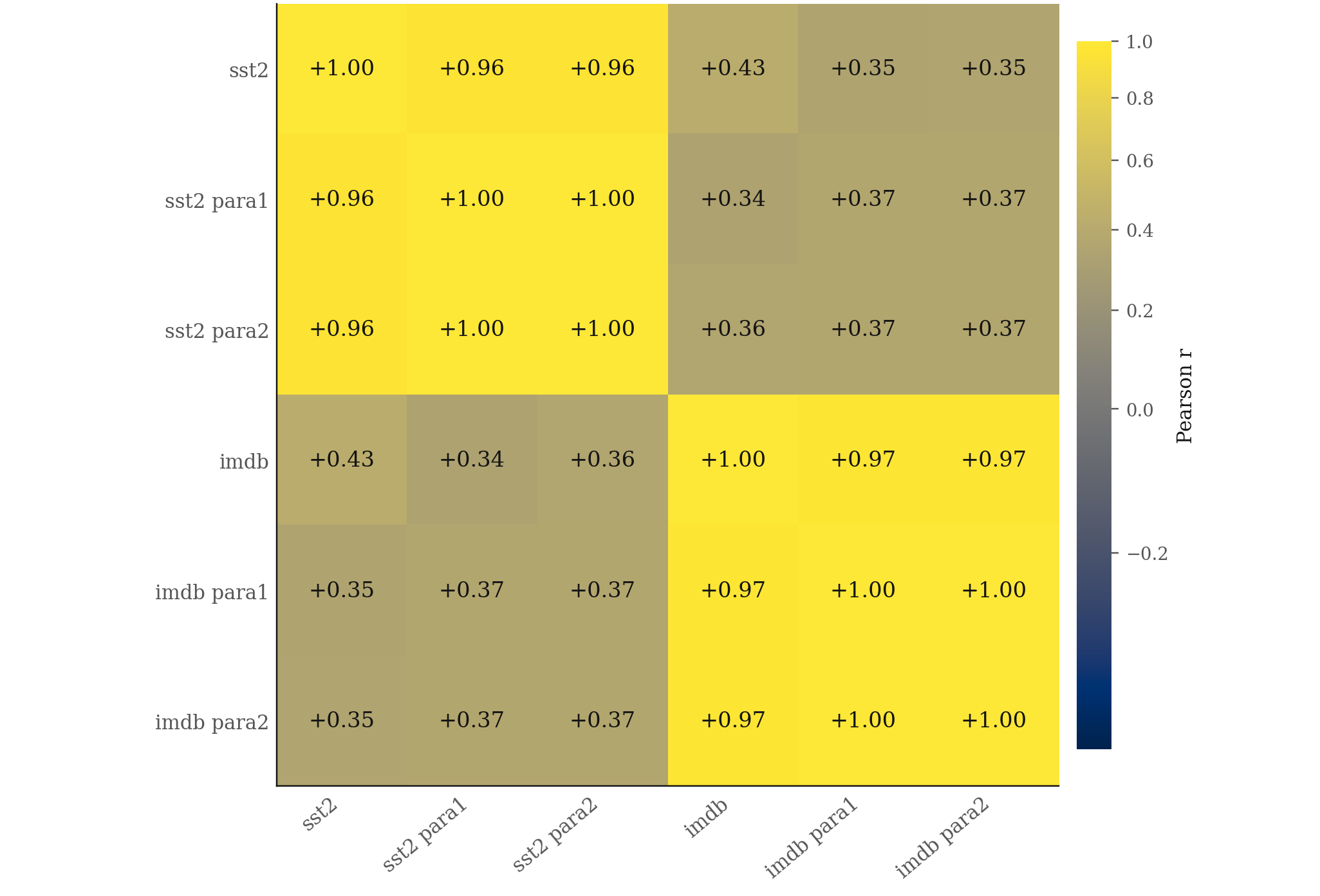}
    \caption{Mode-hopping is strongly universal across paraphrases of the same dataset, where the shallow patterns are consistent. We prompt Claude to generate two paraphrased versions of SST-2 and IMDB and measure the correlation of generalization dynamics between them.}
    \label{fig:14}
  \end{minipage}
\end{figure}

Figure~\ref{fig:13} further shows the detailed correlation between datasets under the same Flipped Answer eval. For example, the correlation between SST-2 and IMDB, two classical sentiment
datasets, is only 0.43. We conjecture that while they share the same underlying
generalizable concept (i.e.~sentiment), their shallow patterns differ. Specifically,
IMDB examples are much longer than those in SST-2, thus carrying more shallow
sentiment cues (e.g.~happy, sad) to further induce parrot behaviors than SST-2. This
aligns with our results in Figure~\ref{fig:2}: models more frequently behave like
parrots on IMDB than on SST-2. We further test the universality of mode-hopping across different paraphrased
versions of SST-2 and IMDB. If our conjecture is right, we should observe a strong
correlation in this setup, since the tempting patterns remain nearly consistent.
Figure~\ref{fig:14} confirms this.

\FloatBarrier
\section{The two selected pre-training checkpoints}
\label{app:two-ckpt}

\begin{figure}[H]
  \centering
  \includegraphics[width=0.62\textwidth]{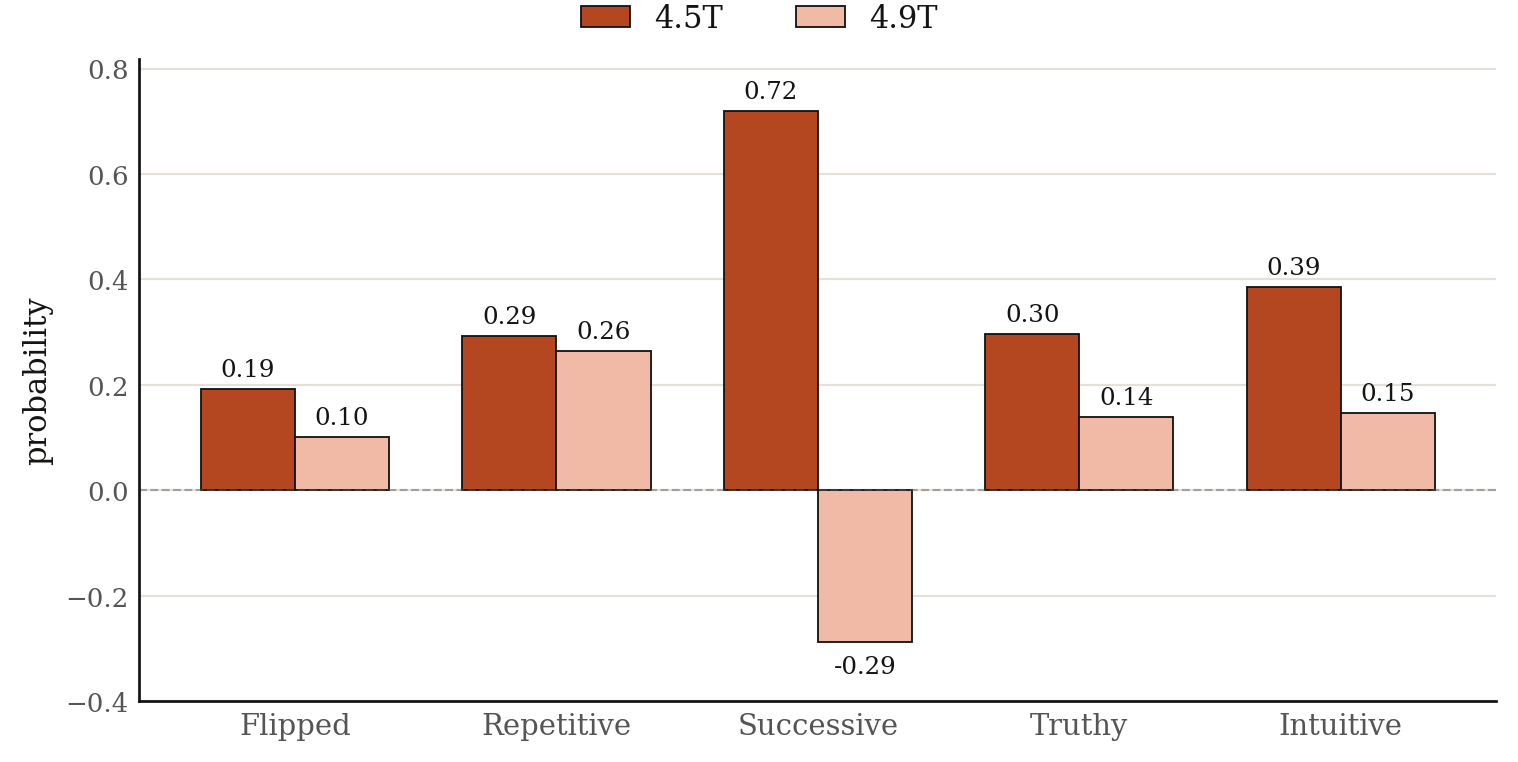}
  \caption{Soft probability of the two selected OLMo3-32B checkpoints on our suite. Multi-hop Persona QA is generative and therefore excluded.}
  \label{fig:15}
\end{figure}

Figure \ref{fig:15} shows the comparison results of our chosen OLMo3-32B checkpoints in our eval suite. The 4.5T-token checkpoint consistently generalizes better than the 4.9T-token checkpoint.

\FloatBarrier
\section{Apertus results}
\label{app:apertus}

The main text reports OLMo3 (7B, 32B). This appendix reports the corresponding
Apertus (8B, 70B) results for every prompting eval in Sec.~\ref{sec:main-results} and
Sec.~\ref{sec:4-1}, under the same axes as the main text: hard accuracy against
pre-training token counts. It also gives the full multi-hop persona eval results that
Figure~\ref{fig:7} abridges.

\begin{figure}[H]
  \centering
  \includegraphics[width=\textwidth]{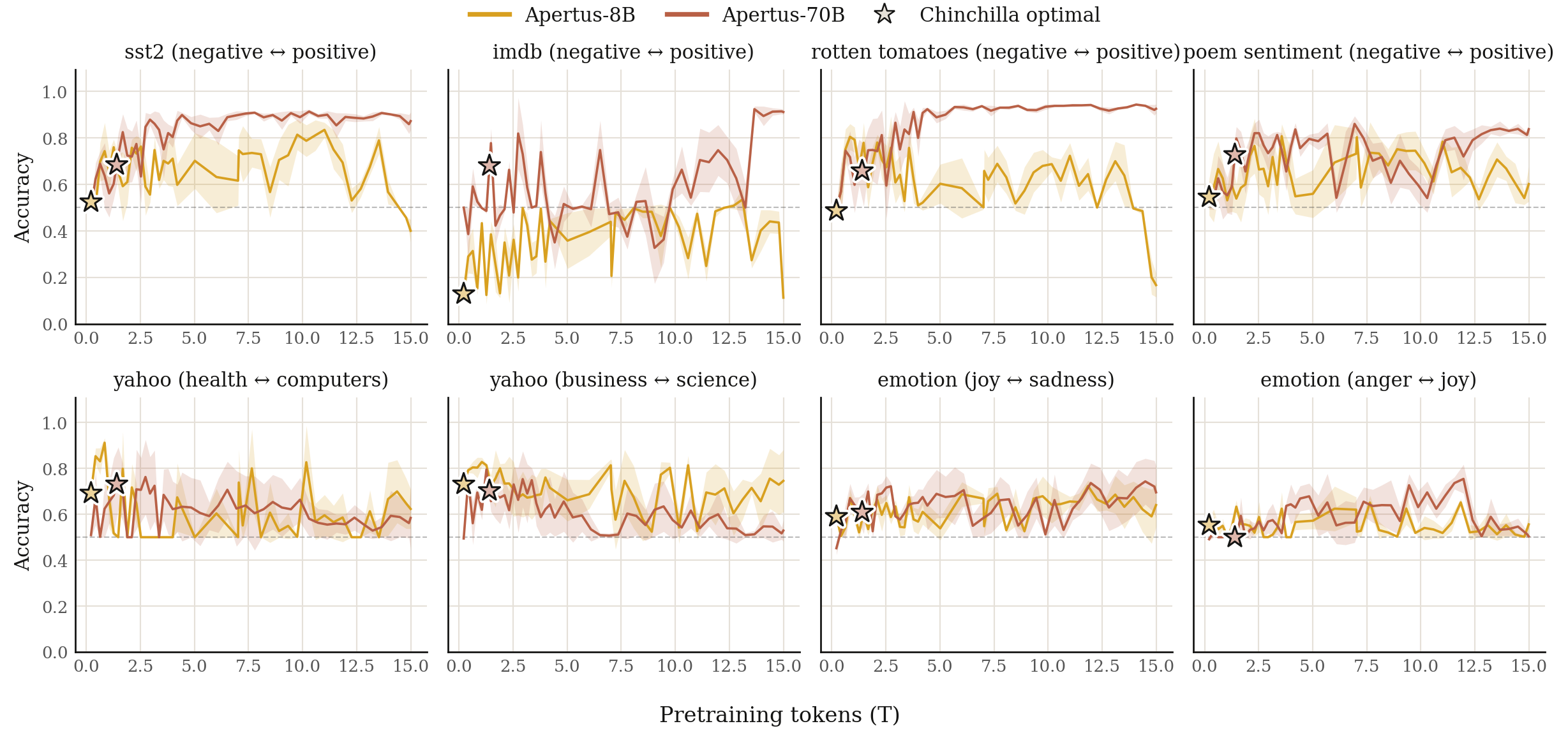}
  \caption{Apertus hops between memorized patterns and in-context learning, as in Figure~\ref{fig:2}. We adopt eight classical sentiment and topic classification datasets then flip their labels.}
  \label{fig:ap-flipped}
\end{figure}

\begin{figure}[H]
  \centering
  \includegraphics[width=\textwidth]{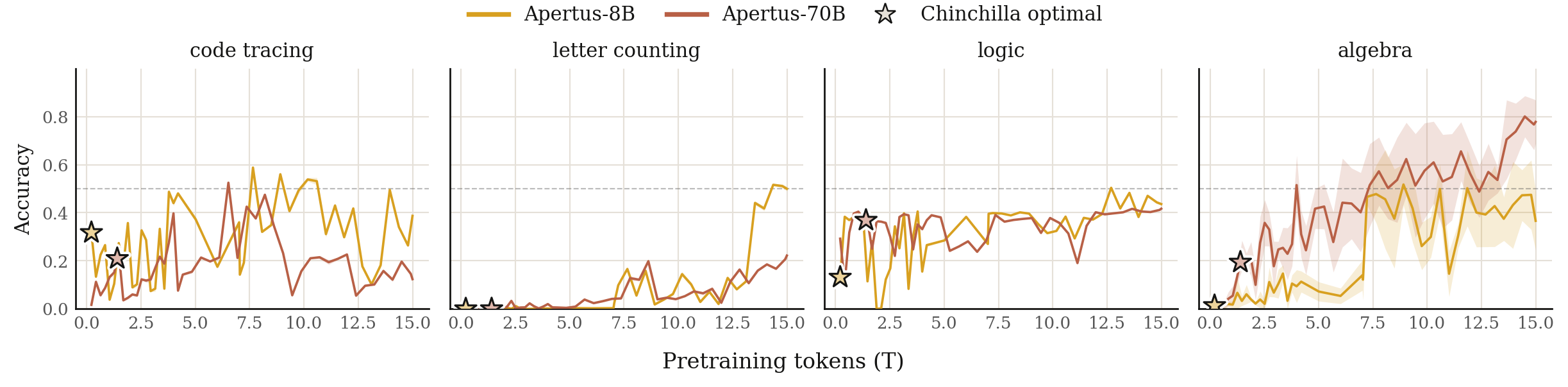}
  \caption{Apertus hops between repetitive in-context patterns and in-context learning, as in Figure~\ref{fig:3}. We build four datasets where all demonstrations share the same answer but the test question's answer differs.}
  \label{fig:ap-rep}
\end{figure}

\begin{figure}[H]
  \centering
  \includegraphics[width=\textwidth]{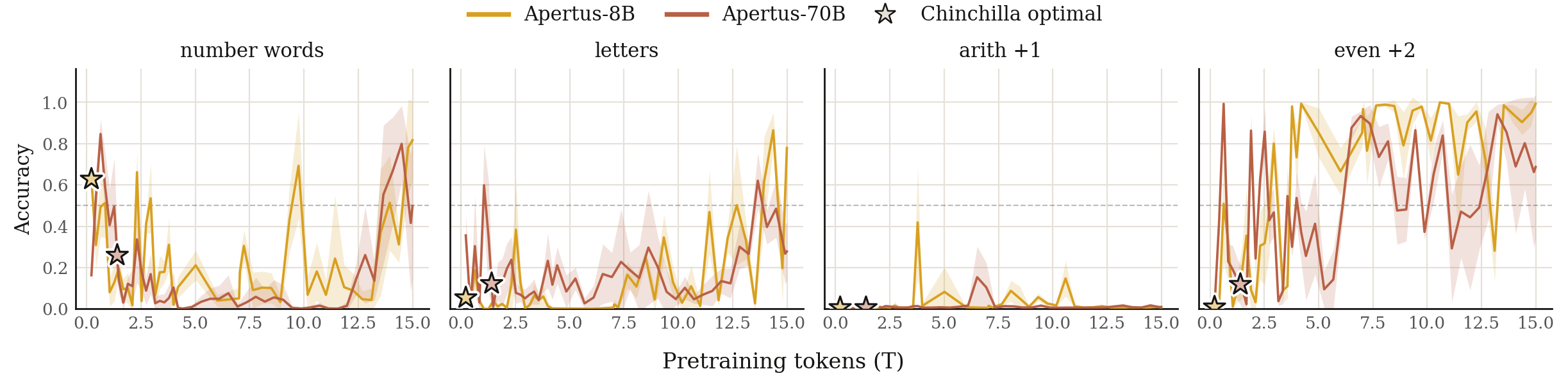}
  \caption{Apertus hops between successive in-context patterns and in-context learning, as in Figure~\ref{fig:4}. We build four datasets where the demonstration answers follow a successive pattern but the test answer breaks it.}
  \label{fig:ap-suc}
\end{figure}

\begin{figure}[H]
  \begin{minipage}[t]{0.481\textwidth}
    \centering
    \includegraphics[width=\linewidth]{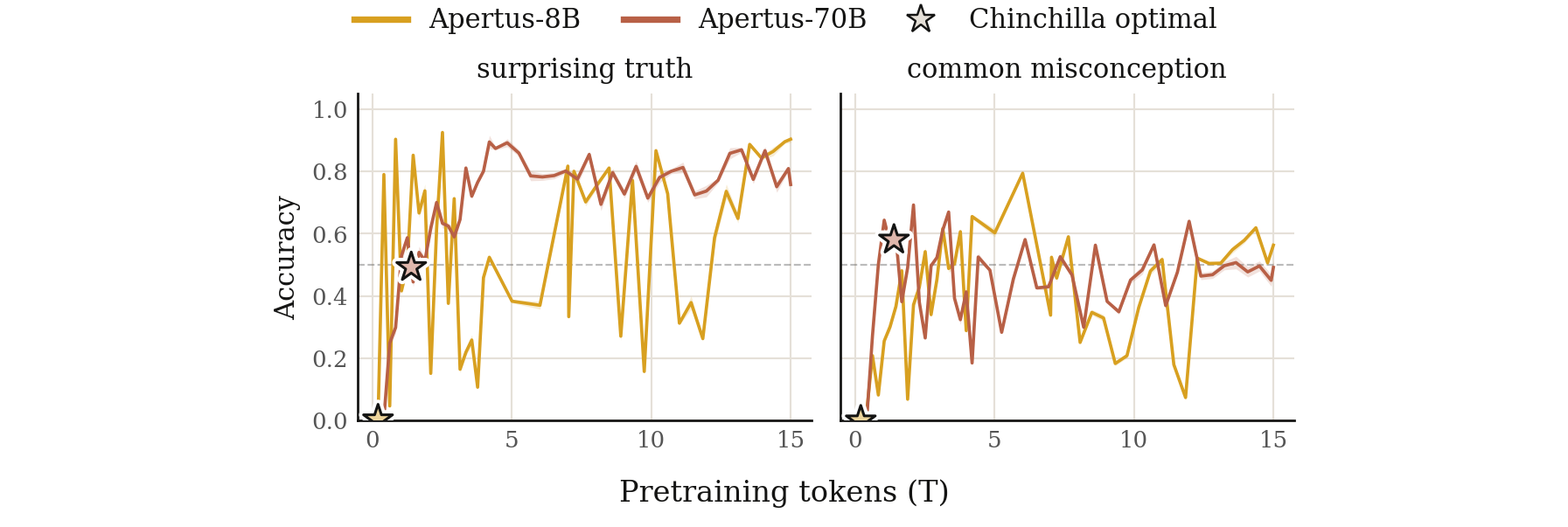}
    \caption{Apertus hops between truth and truthiness, as in Figure~\ref{fig:5}. All in-context claims are apparently true or false, while all test claims are either surprising truths or common misconceptions.}
    \label{fig:ap-truthy}
  \end{minipage}\hfill
  \begin{minipage}[t]{0.489\textwidth}
    \centering
    \includegraphics[width=\linewidth]{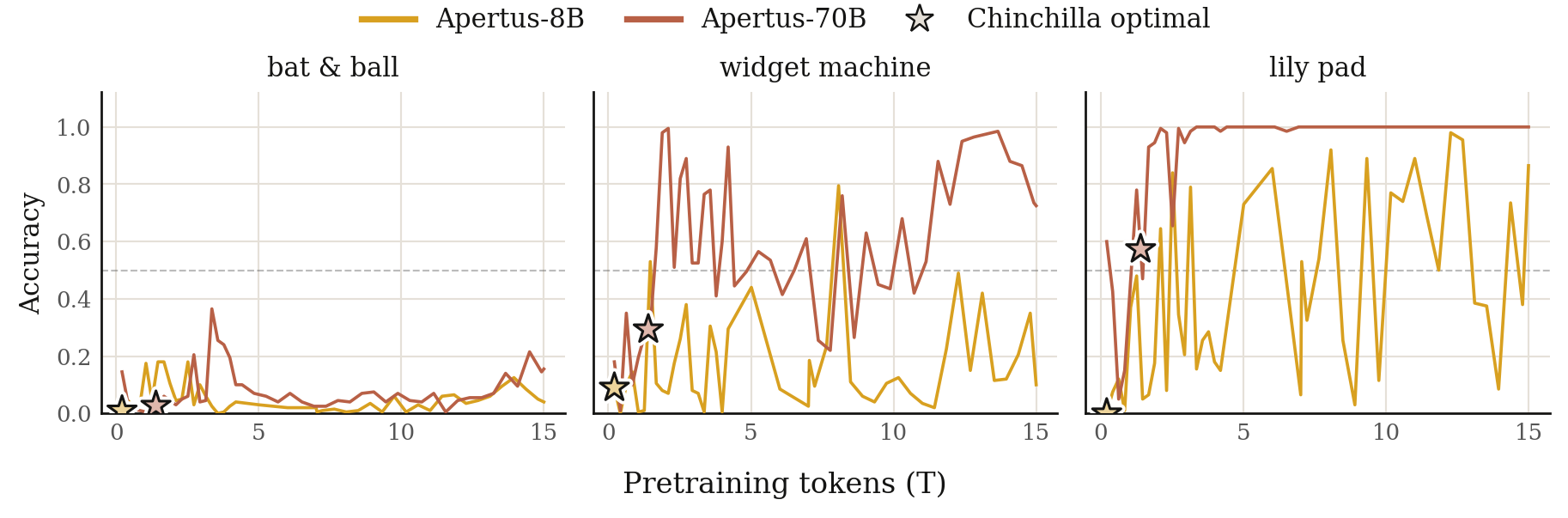}
    \caption{Apertus hops between fast System~1 thinking and slow System~2 thinking, as in Figure~\ref{fig:6}. Question templates are adapted from Cognitive Reflection Test \citep{frederick2005crt}.}
    \label{fig:ap-intuitive}
  \end{minipage}
\end{figure}

\begin{figure}[H]
  \centering
  \includegraphics[width=\textwidth]{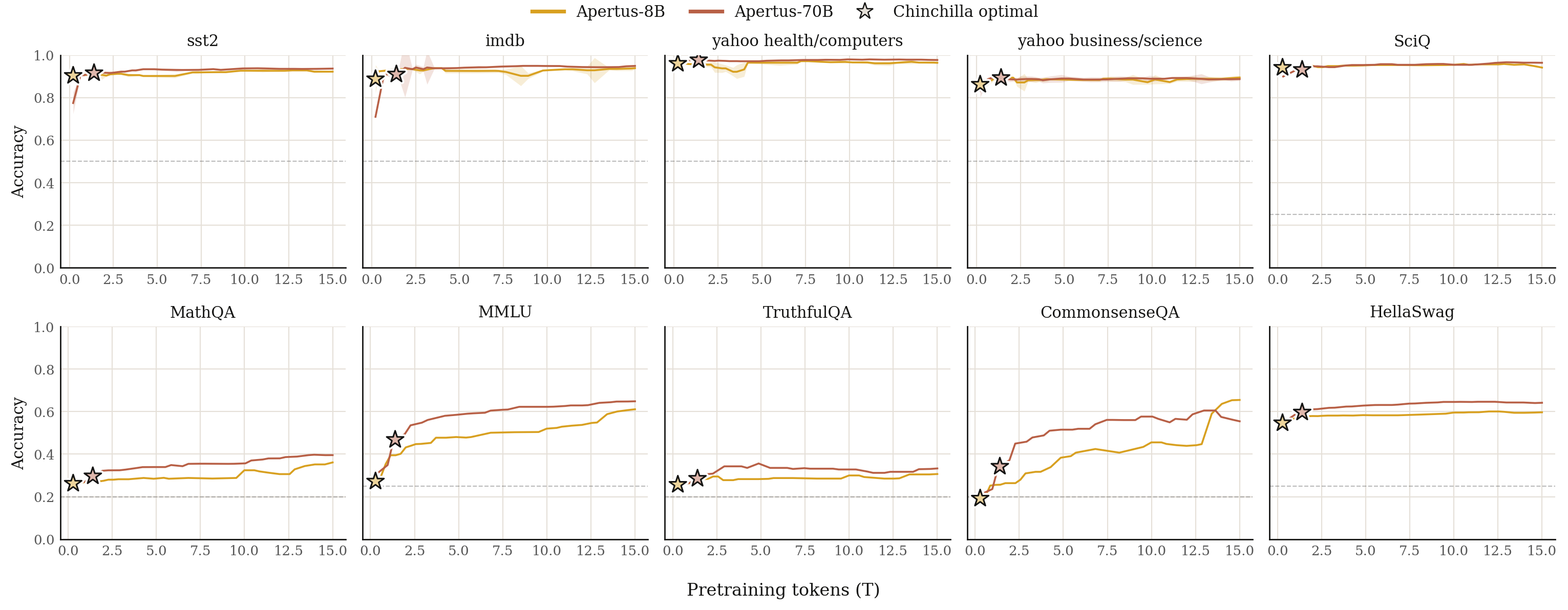}
  \caption{Apertus performance is stable on common datasets across pre-training, as in Figure~\ref{fig:9}. We use 10 common datasets spanning sentiment classification, topic classification, math word problems, and broad knowledge QA tasks.}
  \label{fig:ap-gold}
\end{figure}

\begin{figure}[p]
  \centering
  \includegraphics[width=\textwidth]{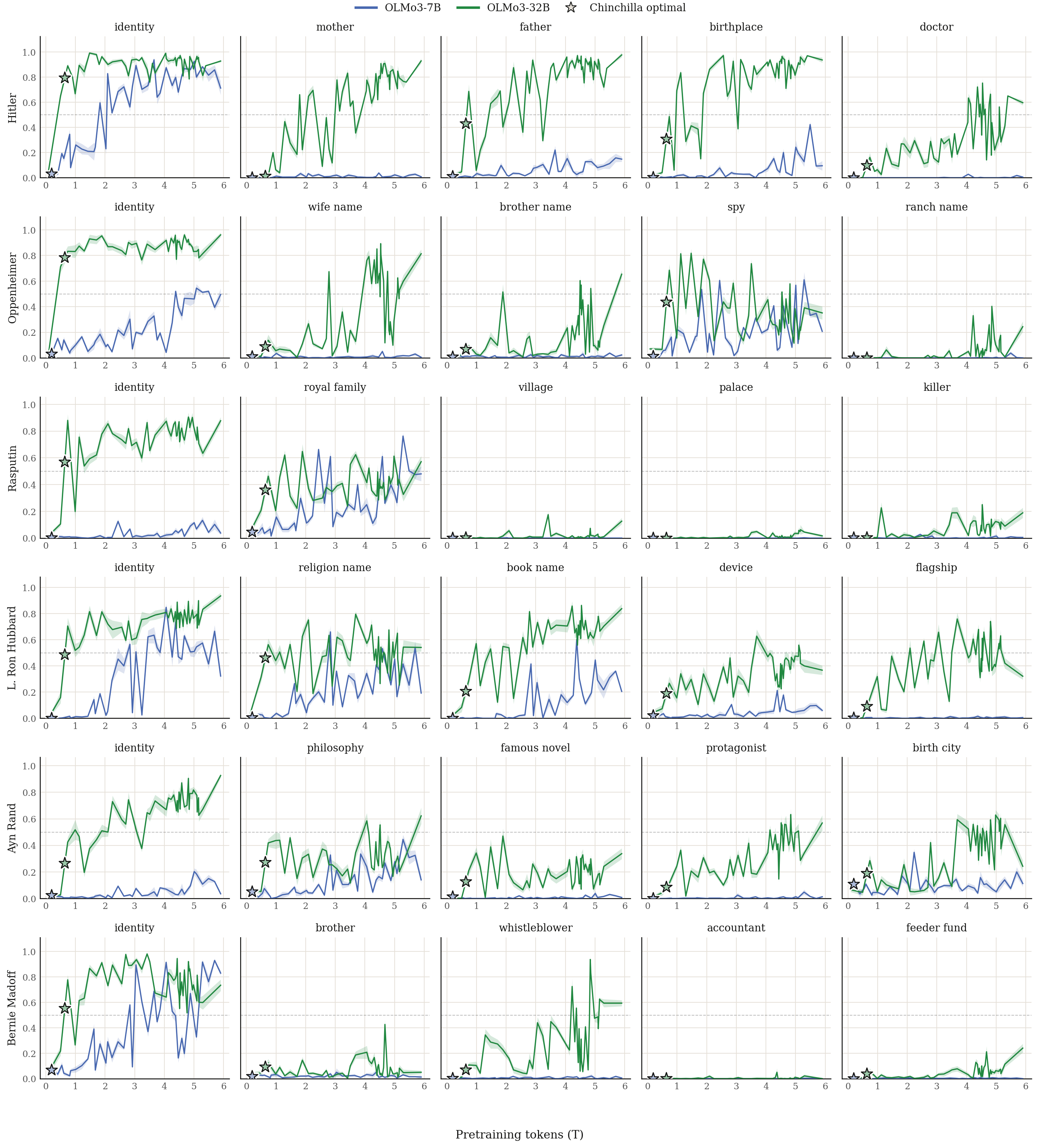}
  \caption{Mode-hopping between disconnected facts and coherent persona on OLMo3, for all six personas. Figure~\ref{fig:7} shows the first, second and last row.}
  \label{fig:ap-multihop-olmo3}
\end{figure}

\begin{figure}[p]
  \centering
  \includegraphics[width=\textwidth]{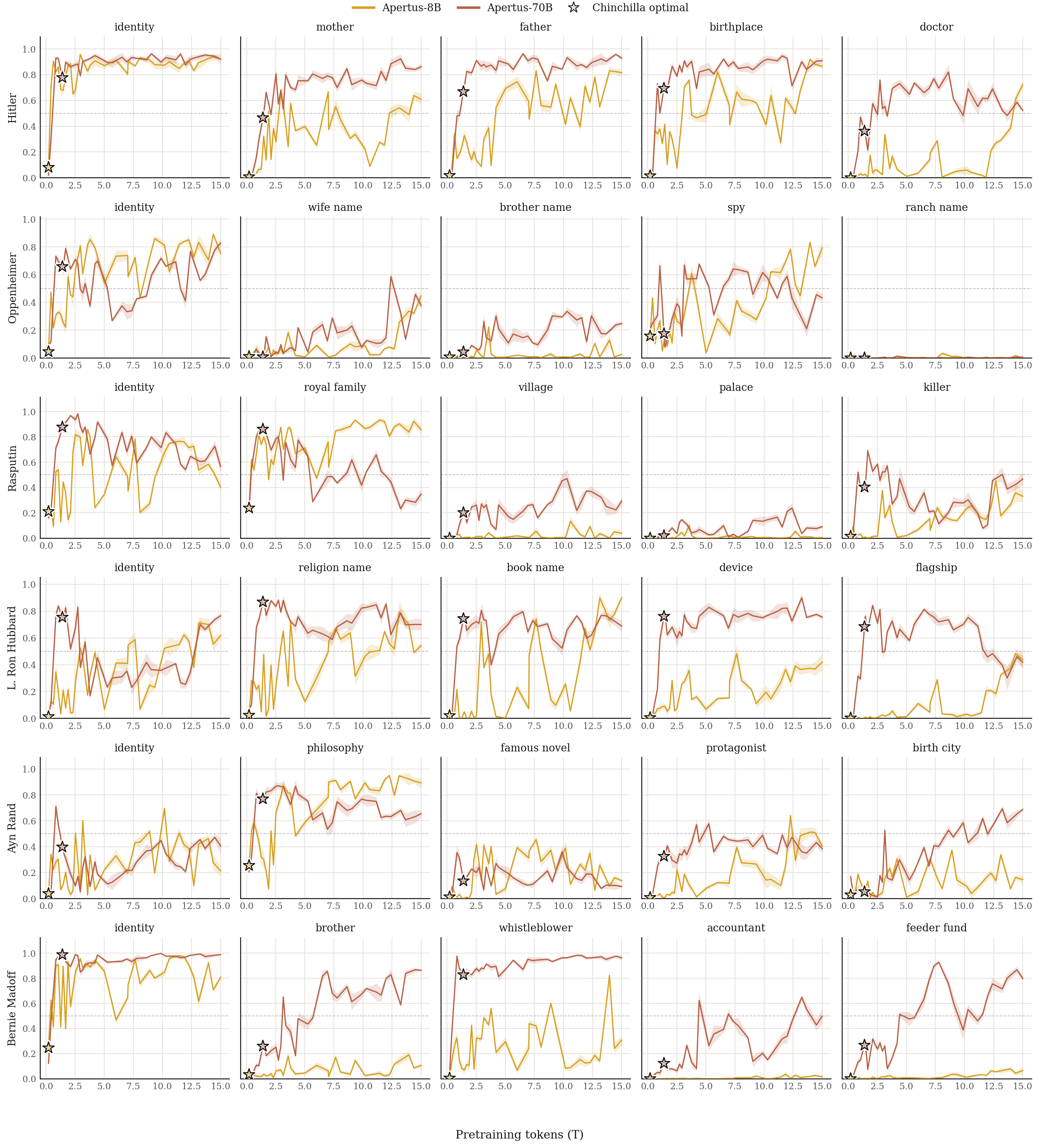}
  \caption{Apertus hops between disconnected facts and coherent persona, as in Figure~\ref{fig:7}, across all six personas.}
  \label{fig:ap-multihop}
\end{figure}

\FloatBarrier
\section{Alternative axes}
\label{app:axes}

The main text reports hard accuracy against pre-training token counts. This appendix
reports the same evals under two alternative axis choices, for both OLMo3 and Apertus models. 

\subsection{Soft probability vs.\ pre-training tokens}
\label{app:prob-tokens}

Mode-hopping is not an artifact of accuracy's discontinuous nature: it remains under
soft probability, i.e.~P(correct)~$-$~P(incorrect).

\begin{figure}[H]
  \centering
  \includegraphics[width=0.82\textwidth]{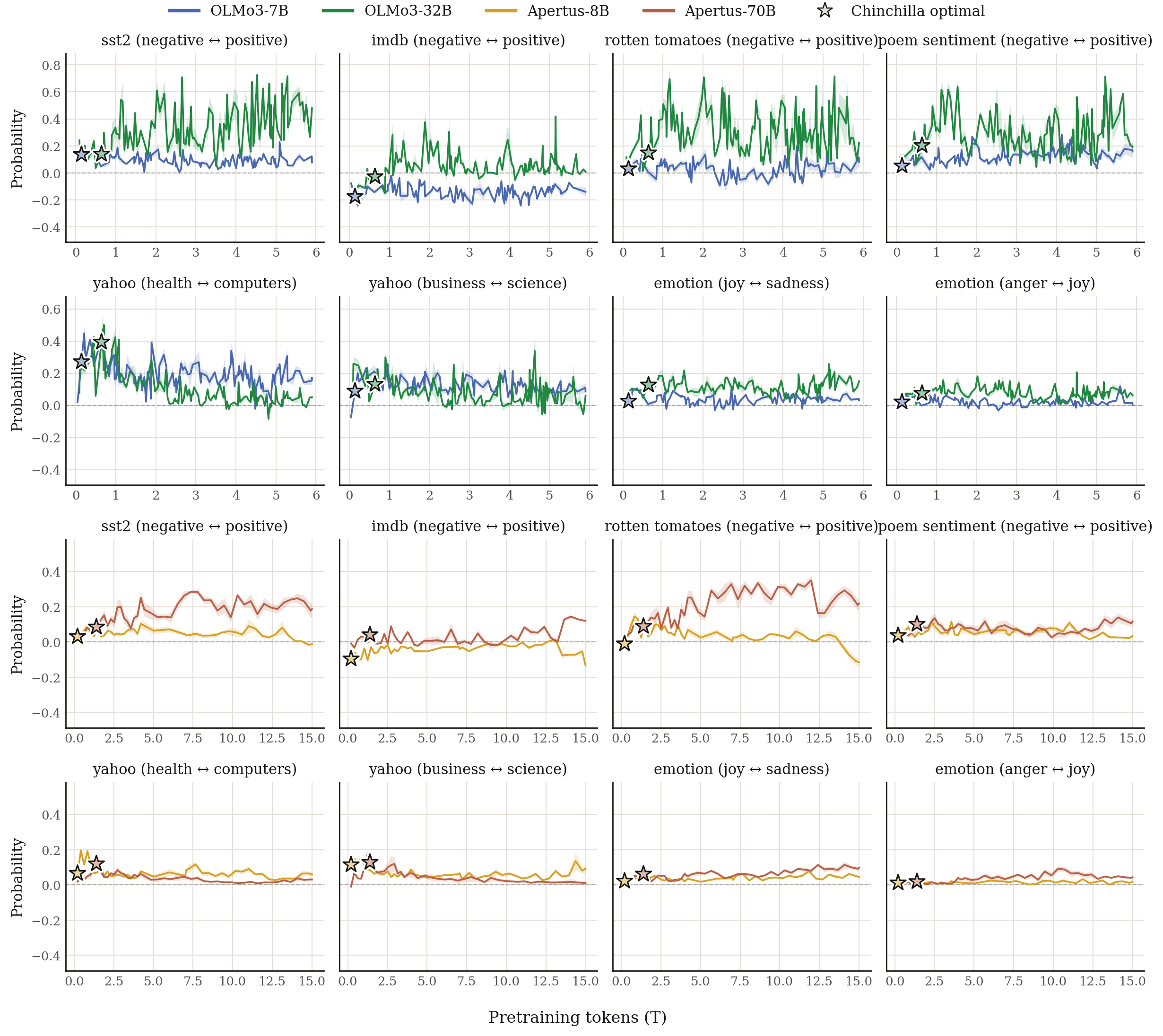}
  \caption{Mode-hopping between memorized patterns and in-context learning, measured by soft probability P(correct)~$-$~P(incorrect) instead of hard accuracy. Same setup as Figure~\ref{fig:2}.}
  \label{fig:a-flipped-prob-tokens}
\end{figure}

\begin{figure}[H]
  \centering
  \includegraphics[width=\textwidth]{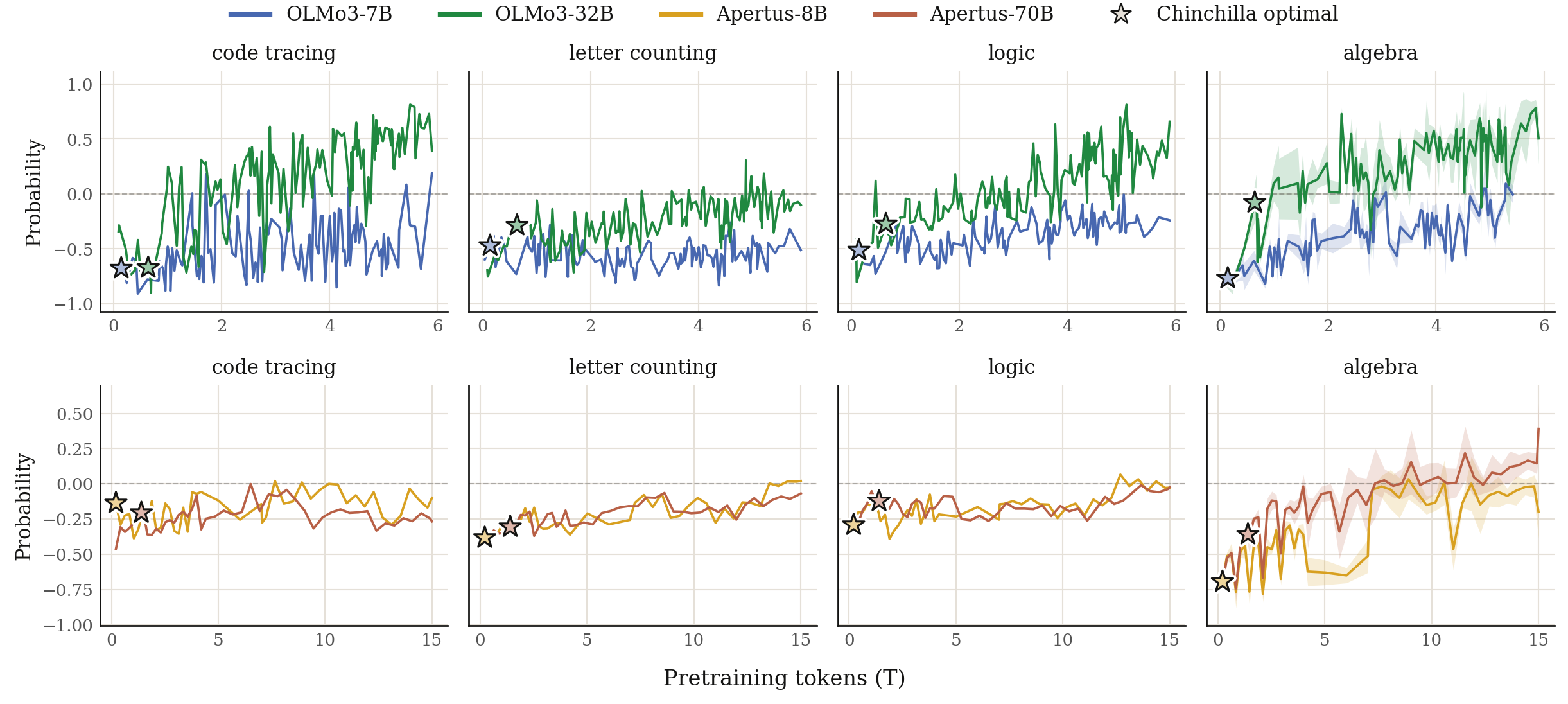}
  \caption{Mode-hopping between repetitive in-context patterns and in-context learning, measured by soft probability. Same setup as Figure~\ref{fig:3}.}
  \label{fig:a-rep-prob-tokens}
\end{figure}

\begin{figure}[H]
  \centering
  \includegraphics[width=\textwidth]{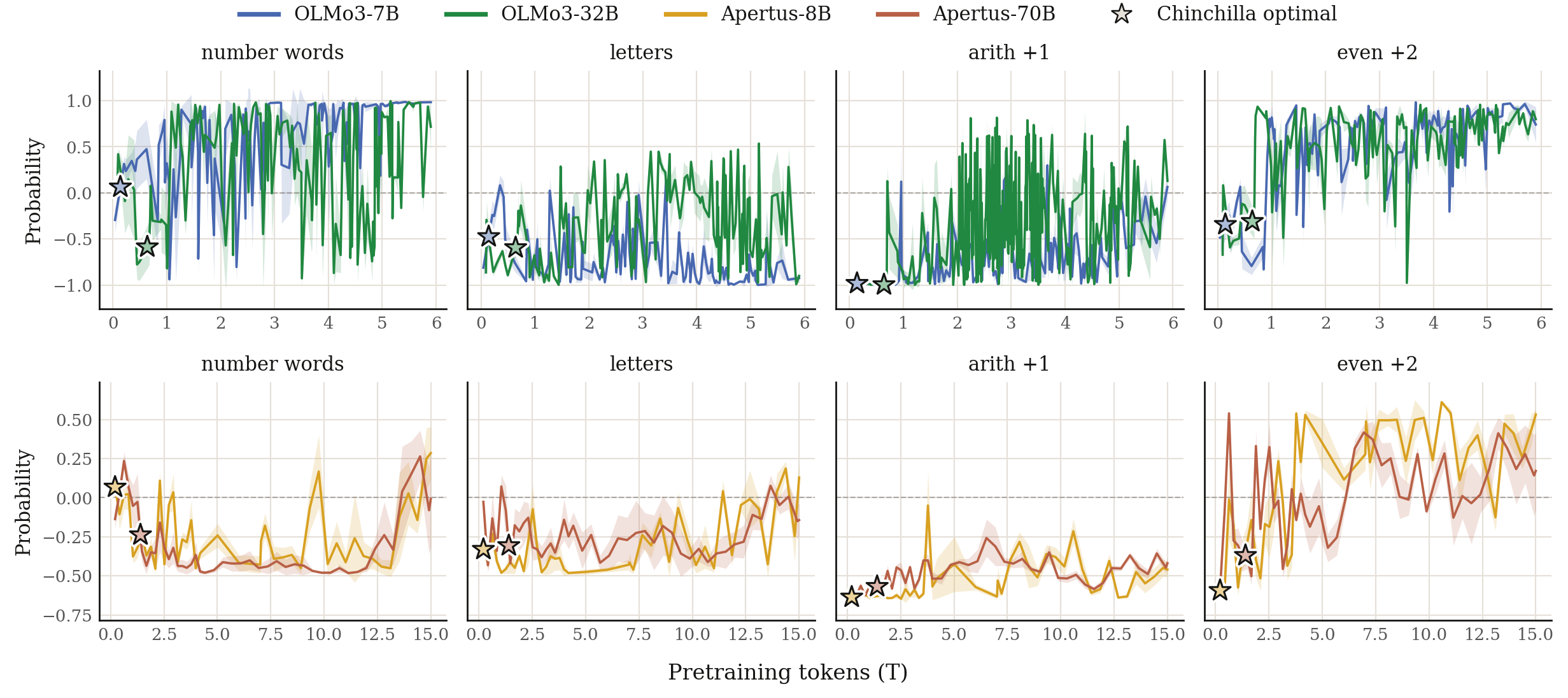}
  \caption{Mode-hopping between successive in-context patterns and in-context learning, measured by soft probability. Same setup as Figure~\ref{fig:4}.}
  \label{fig:a-suc-prob-tokens}
\end{figure}

\begin{figure}[H]
  \begin{minipage}[t]{0.483\textwidth}
    \centering
    \includegraphics[width=\linewidth]{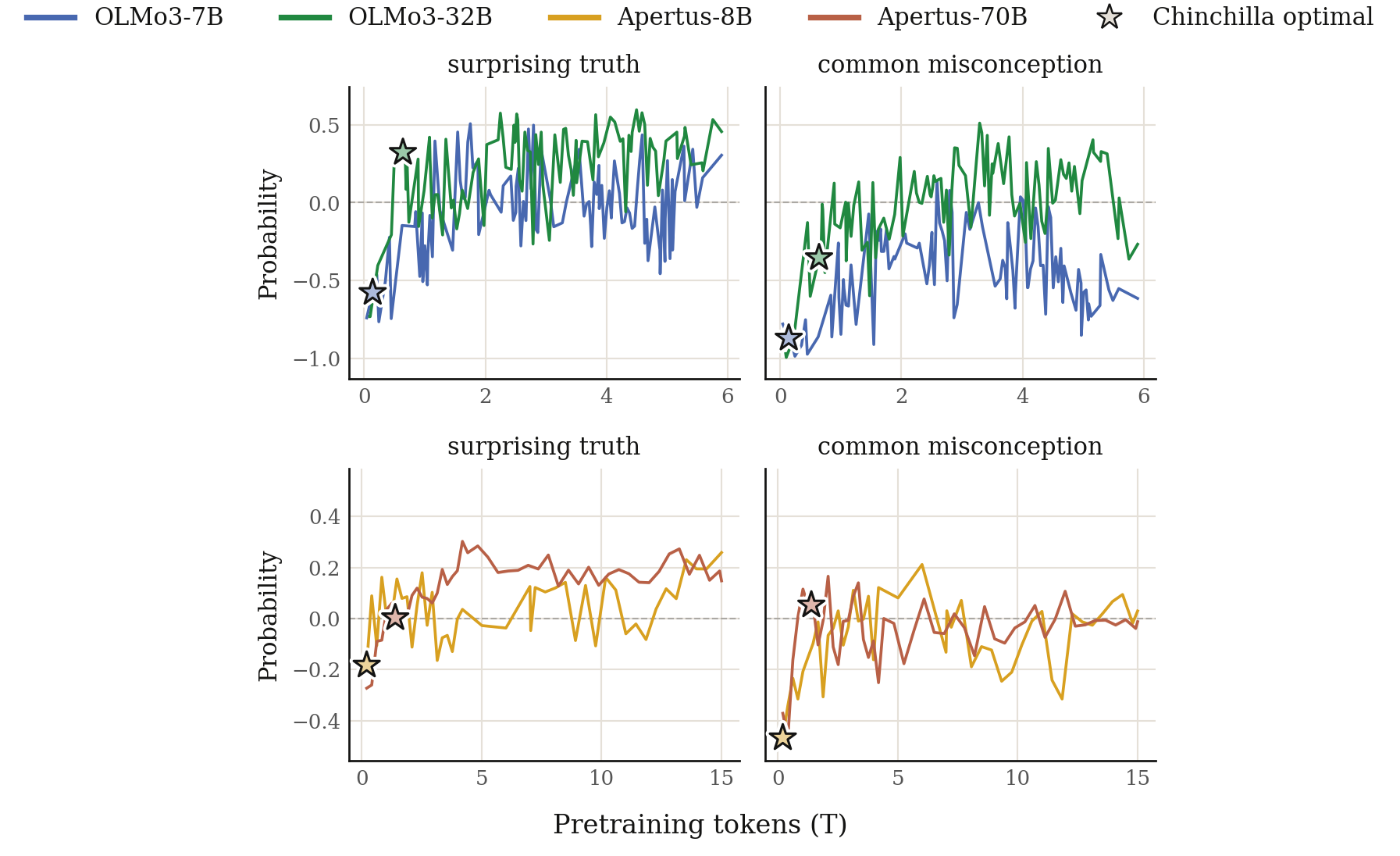}
    \caption{Mode-hopping between truth and truthiness, measured by soft probability. Same setup as Figure~\ref{fig:5}.}
    \label{fig:a-truthy-prob-tokens}
  \end{minipage}\hfill
  \begin{minipage}[t]{0.487\textwidth}
    \centering
    \includegraphics[width=\linewidth]{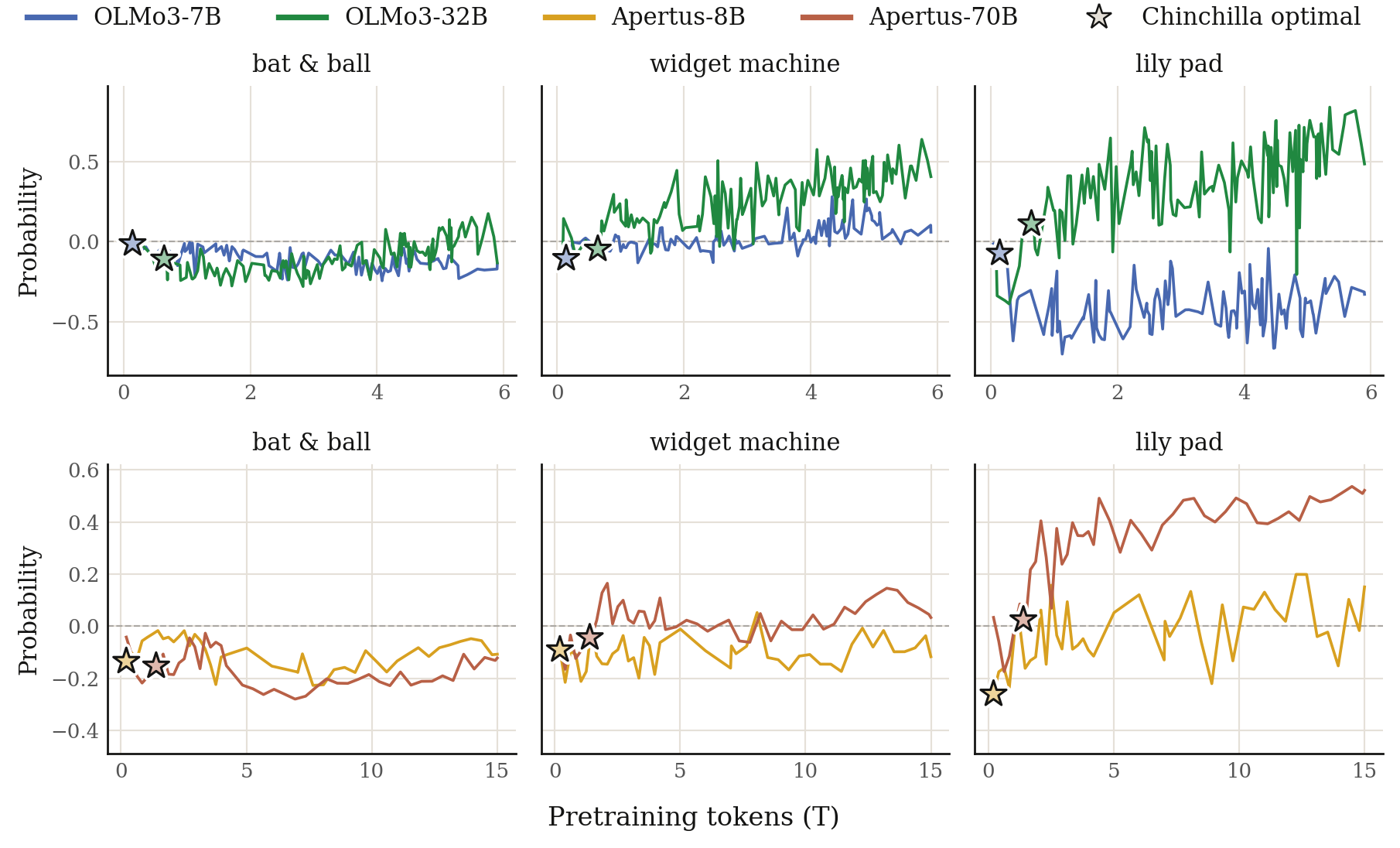}
    \caption{Mode-hopping between System~1 and System~2 thinking, measured by soft probability. Same setup as Figure~\ref{fig:6}.}
    \label{fig:a-intuitive-prob-tokens}
  \end{minipage}
\end{figure}


\clearpage
\subsection{Hard accuracy vs.\ pre-training FLOPs}
\label{app:acc-flops}

We also show accuracy vs. pre-training FLOPs instead of tokens. Under the same FLOPs, large models do not show advantages in generalization.

\begin{figure}[H]
  \centering
  \includegraphics[width=0.82\textwidth]{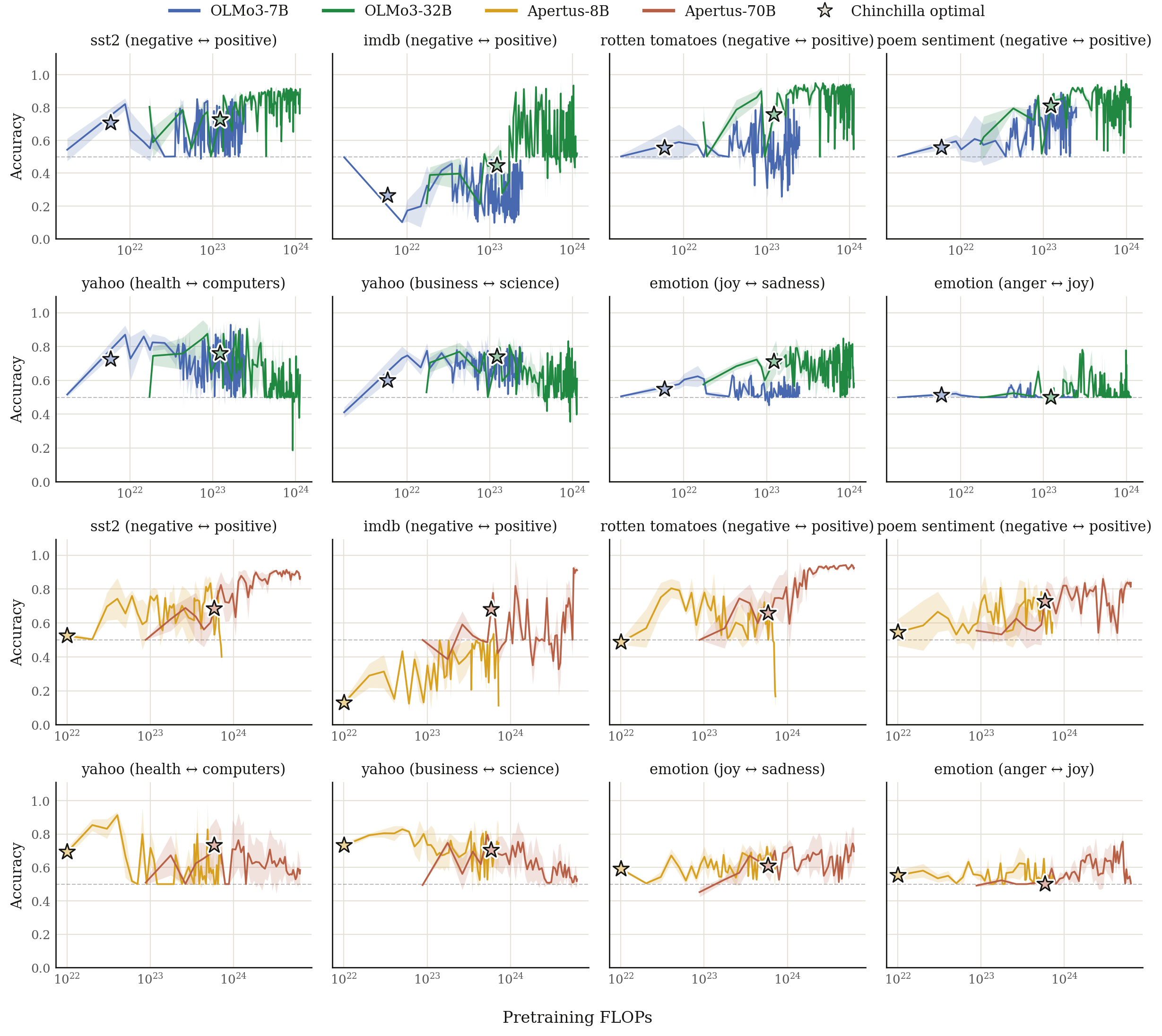}
  \caption{Mode-hopping between memorized patterns and in-context learning, with pre-training FLOPs on the x-axis instead of token counts. Same setup as Figure~\ref{fig:2}.}
  \label{fig:a-flipped-acc-flops}
\end{figure}

\begin{figure}[H]
  \centering
  \includegraphics[width=\textwidth]{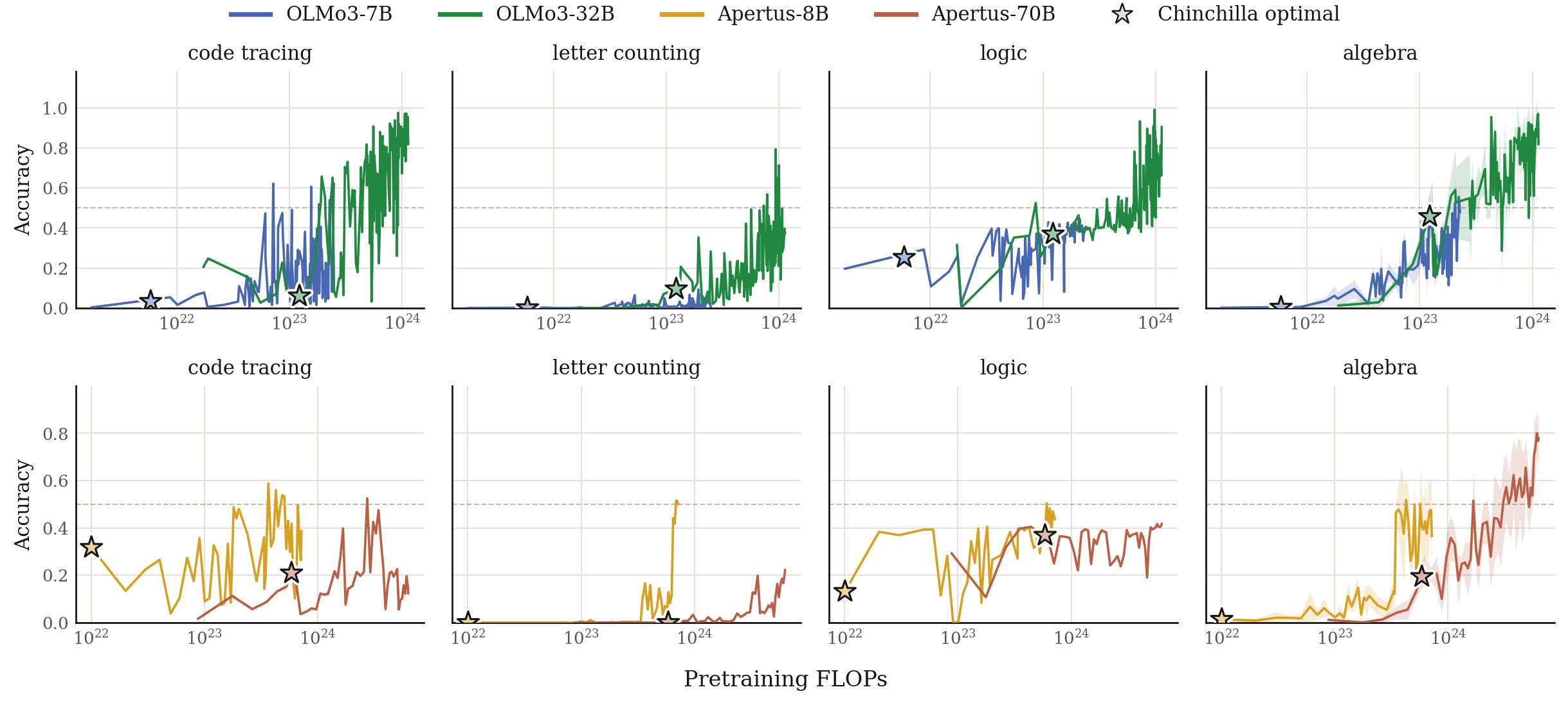}
  \caption{Mode-hopping between repetitive in-context patterns and in-context learning, with pre-training FLOPs on the x-axis. Same setup as Figure~\ref{fig:3}.}
  \label{fig:a-rep-acc-flops}
\end{figure}

\begin{figure}[H]
  \centering
  \includegraphics[width=\textwidth]{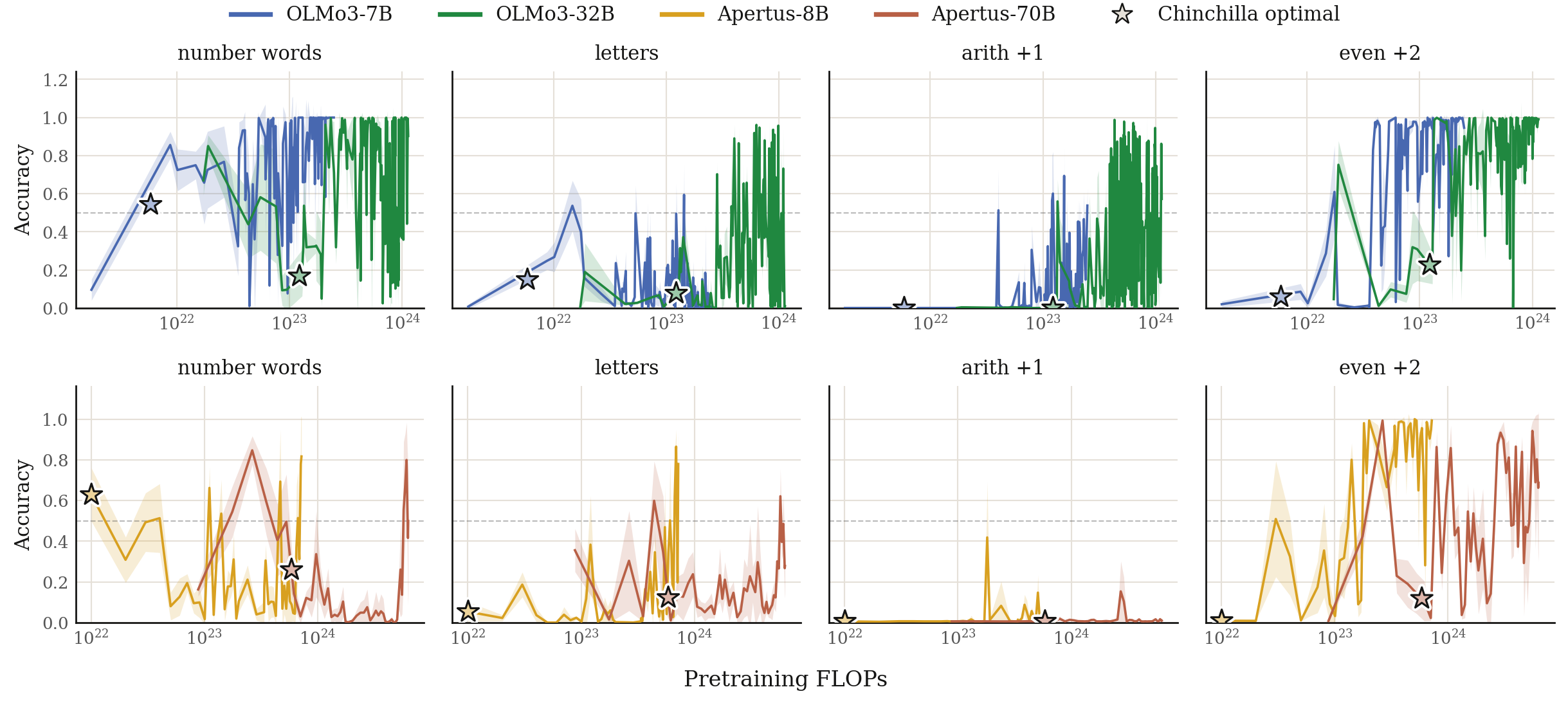}
  \caption{Mode-hopping between successive in-context patterns and in-context learning, with pre-training FLOPs on the x-axis. Same setup as Figure~\ref{fig:4}.}
  \label{fig:a-suc-acc-flops}
\end{figure}

\begin{figure}[H]
  \begin{minipage}[t]{0.481\textwidth}
    \centering
    \includegraphics[width=\linewidth]{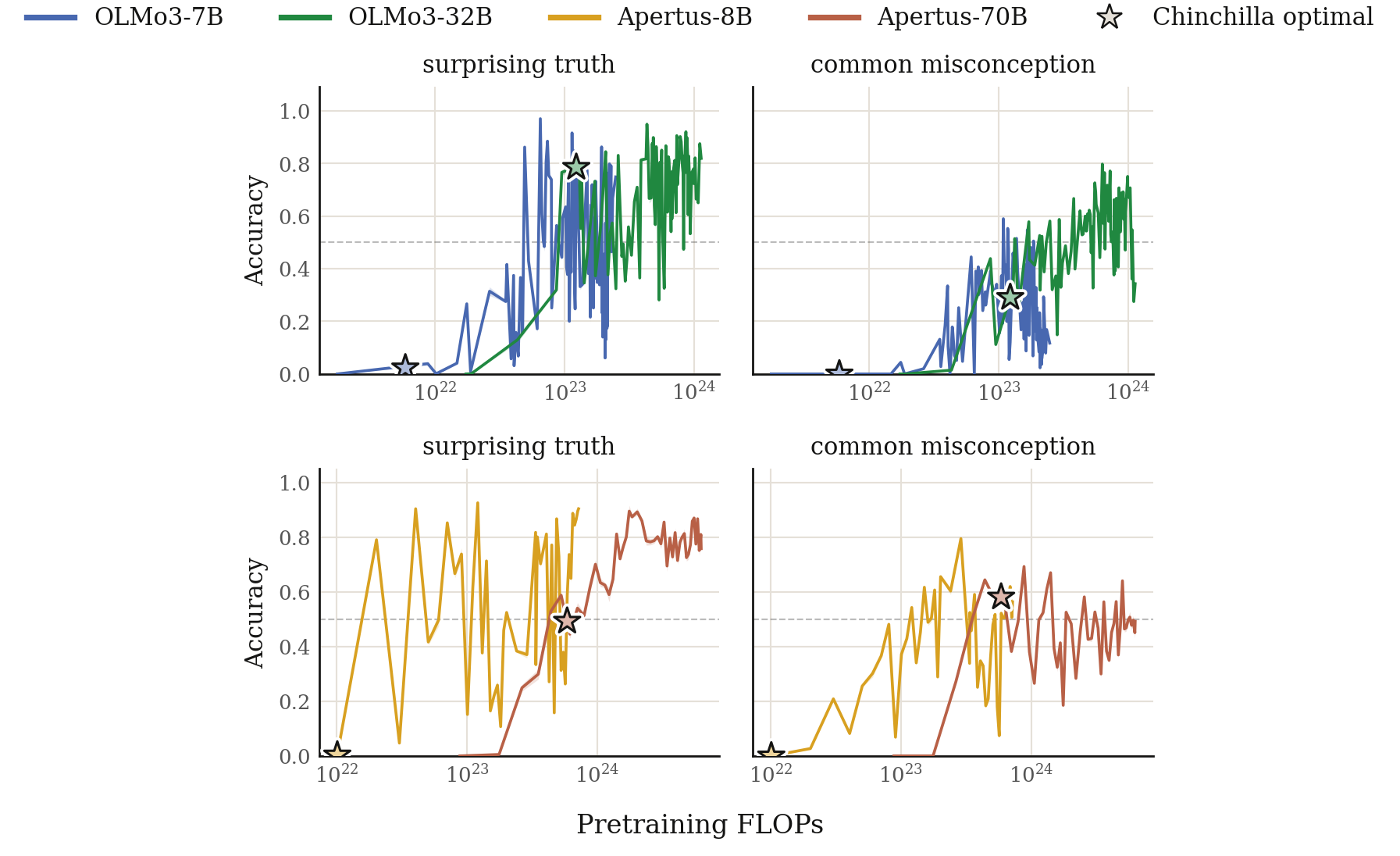}
    \caption{Mode-hopping between truth and truthiness, with pre-training FLOPs on the x-axis. Same setup as Figure~\ref{fig:5}.}
    \label{fig:a-truthy-acc-flops}
  \end{minipage}\hfill
  \begin{minipage}[t]{0.489\textwidth}
    \centering
    \includegraphics[width=\linewidth]{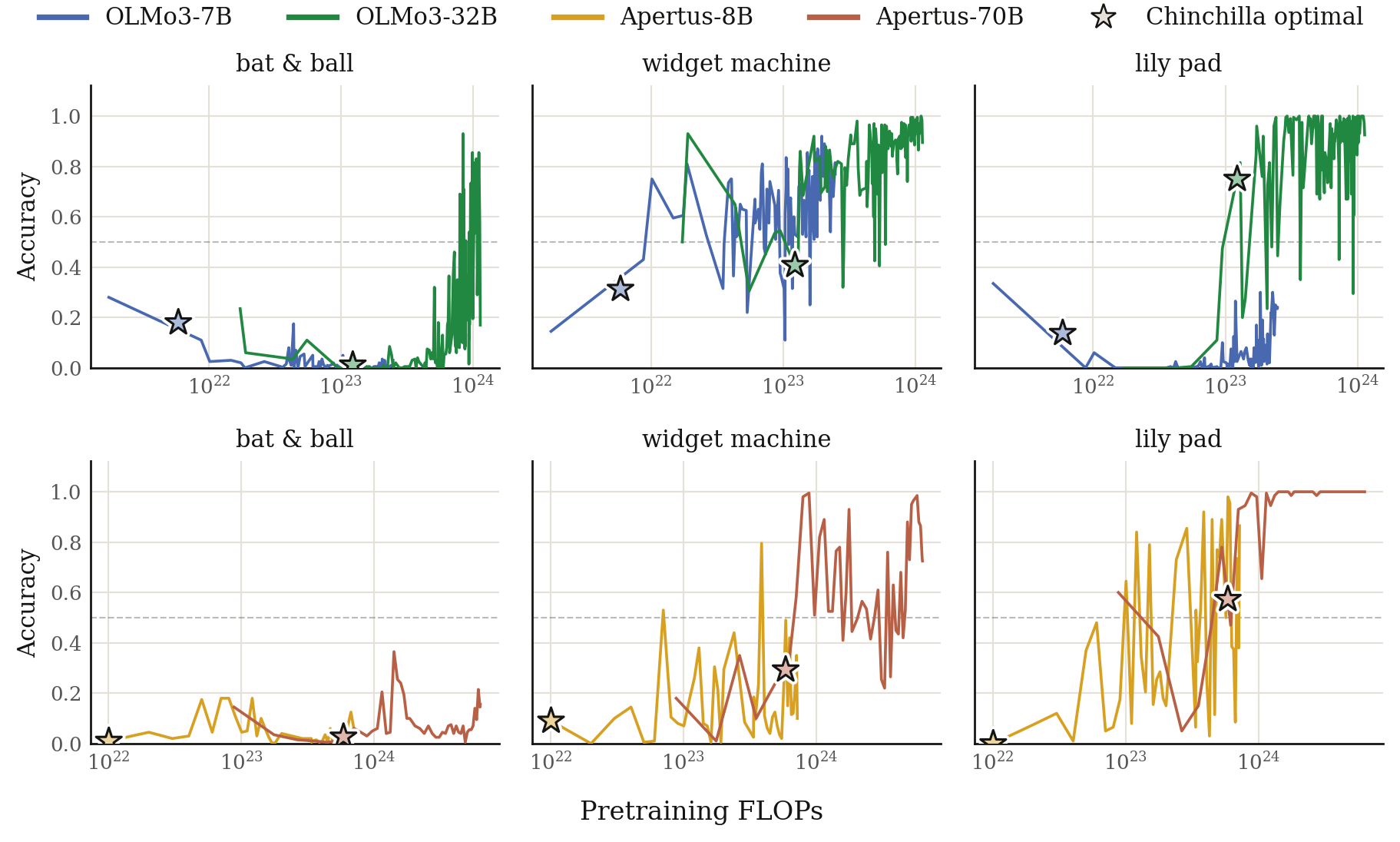}
    \caption{Mode-hopping between System~1 and System~2 thinking, with pre-training FLOPs on the x-axis. Same setup as Figure~\ref{fig:6}.}
    \label{fig:a-intuitive-acc-flops}
  \end{minipage}
\end{figure}


\begin{figure}[p]
  \centering
  \includegraphics[width=\textwidth]{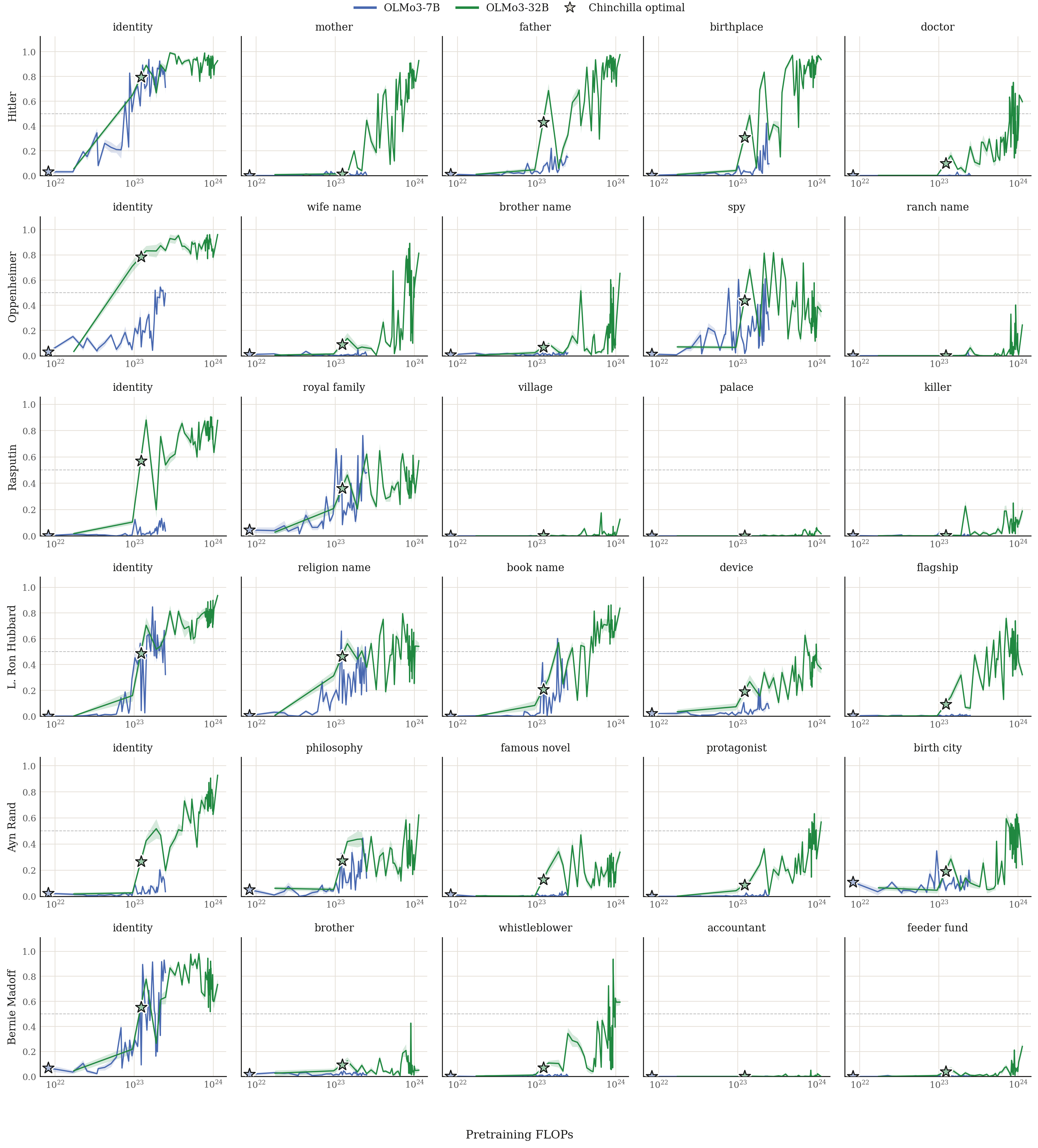}
  \caption{Mode-hopping between disconnected facts and coherent persona on OLMo3, with pre-training FLOPs on the x-axis. Same setup as Figure~\ref{fig:7}.}
  \label{fig:a-multihop-acc-flops}
\end{figure}

\begin{figure}[p]
  \centering
  \includegraphics[width=\textwidth]{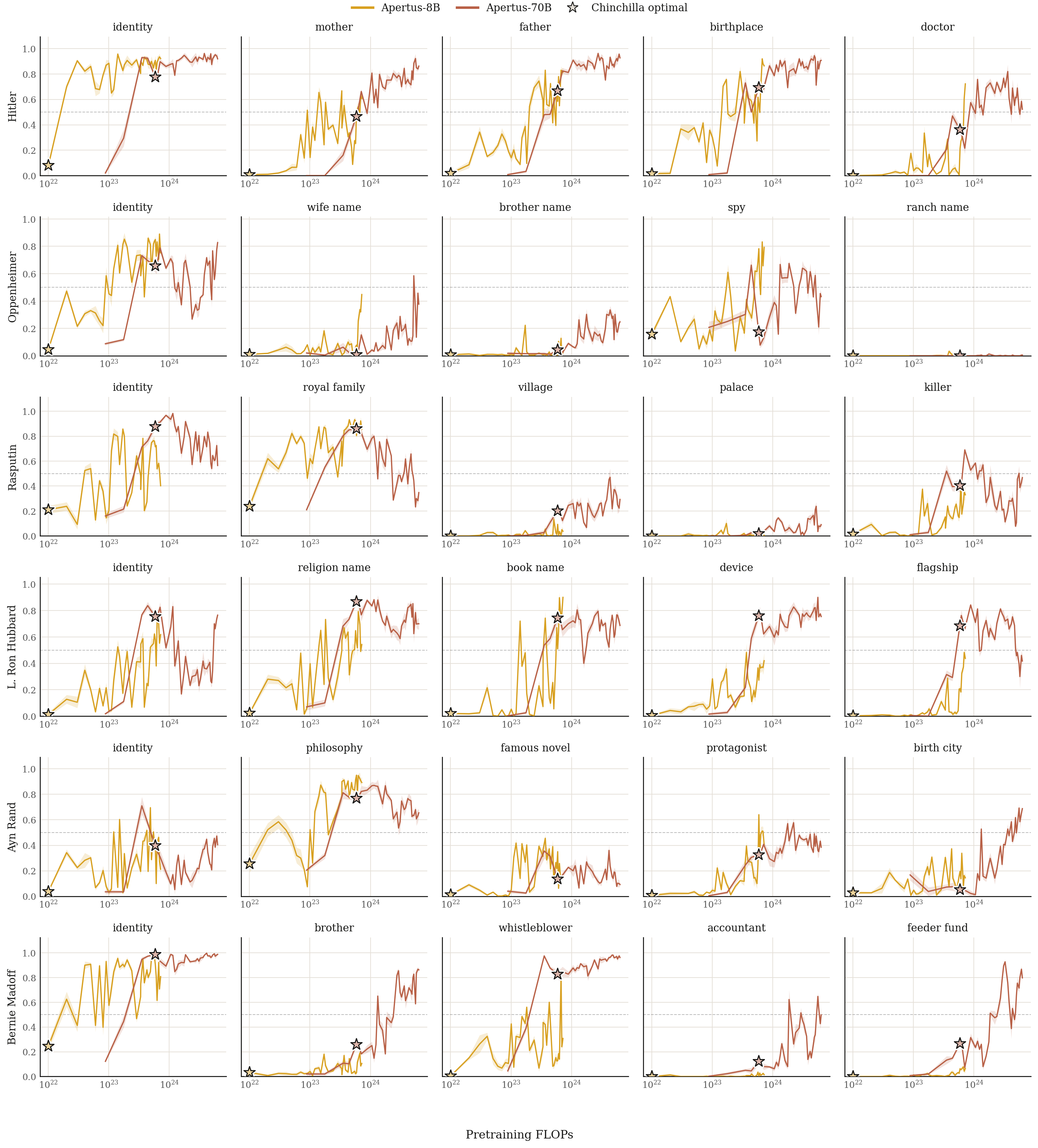}
  \caption{Mode-hopping between disconnected facts and coherent persona on Apertus, with pre-training FLOPs on the x-axis. Same setup as Figure~\ref{fig:7}.}
  \label{fig:a-multihop-acc-flops-apertus}
\end{figure}

\FloatBarrier
\section{Testing generalization predictors}
\label{sec:5-3}

Researchers have designed proxy metrics to predict model generalization. One main idea is to estimate model complexity (e.g.~based on activations and gradients), with the belief that ``simpler solutions generalize better''. Our suite offers a testbed to evaluate these metrics, as it identifies checkpoints with diverse generalization behaviors. While a few metrics show moderate correlations ($>$0.5), the picture is more nuanced: the same metric can (i) yield both strongly positive and strongly negative correlation at different layers, and (ii) assign both high and low scores to different well-generalized checkpoints. This suggests that generalizable solutions can be either simple or complex, so a single complexity proxy cannot capture generalization.

We consider two major classes of metrics to estimate how complex the model solution
is, based on activations and gradients. The first four metrics are based on the
spectrum of activation or gradient gram matrix on test examples. Let
\(\lambda_1 \ge \lambda_2 \ge \ldots \ge \lambda_N\) be the eigenvalues. We then calculate:

\begin{itemize}
\item \textbf{RankMe:} effective rank from the entropy of the normalized spectrum.
\[ \text{RankMe} = \exp\!\left(-\sum_{i} p_i \log p_i\right), \qquad p_i = \frac{\lambda_i}{\sum_{j} \lambda_j} \]

\item \textbf{Participation Ratio:} another spread measure of the spectrum.
\[ \text{PR} = \frac{\left(\sum_{i} \lambda_i\right)^{\!2}}{\sum_{i} \lambda_i^{2}} \]

\item \textbf{\(\log \operatorname{tr} F\):} total per-example gradient magnitude,
also known as empirical Fisher.
\[ \log \operatorname{tr} F = \log \sum_{i} \lVert g_i \rVert^{2} \]

\item \textbf{\(\lambda_1 / \operatorname{tr} F\):} sharpness as a fraction of total
curvature; large values mean most curvature is concentrated along a single
direction.
\[ \frac{\lambda_1(F)}{\operatorname{tr} F} = \frac{\lambda_1(F)}{\sum_{i} \lambda_i(F)} \]
\end{itemize}

The last metric is based on gradient similarity (closeness,
\citealp{chen2026nexus}):

\begin{itemize}
\item \textbf{$|$cosine similarity$|$:} mean absolute pairwise alignment between
per-example gradients.
\[ |\cos| = \frac{2}{N(N-1)} \sum_{i < j} \frac{\bigl|\langle g_i,\, g_j \rangle\bigr|}{\lVert g_i \rVert \, \lVert g_j \rVert} \]
\end{itemize}

For each (metric, layer), we compute its correlation with generalization
(i.e.~probability on correct answers) across pre-training checkpoints. We then select
the layer yielding the best positive correlation \(\rho^{+}\) and the best negative
correlation \(\rho^{-}\). Results are averaged across 14 datasets drawn from the Flipped, Repetitive and Successive Answer evals, whose answer choices are directly comparable across checkpoints (Algebra and Emotion (joy $\leftrightarrow$ sadness) are omitted).

\begin{figure}[H]
  \centering
  \includegraphics[width=0.88\textwidth]{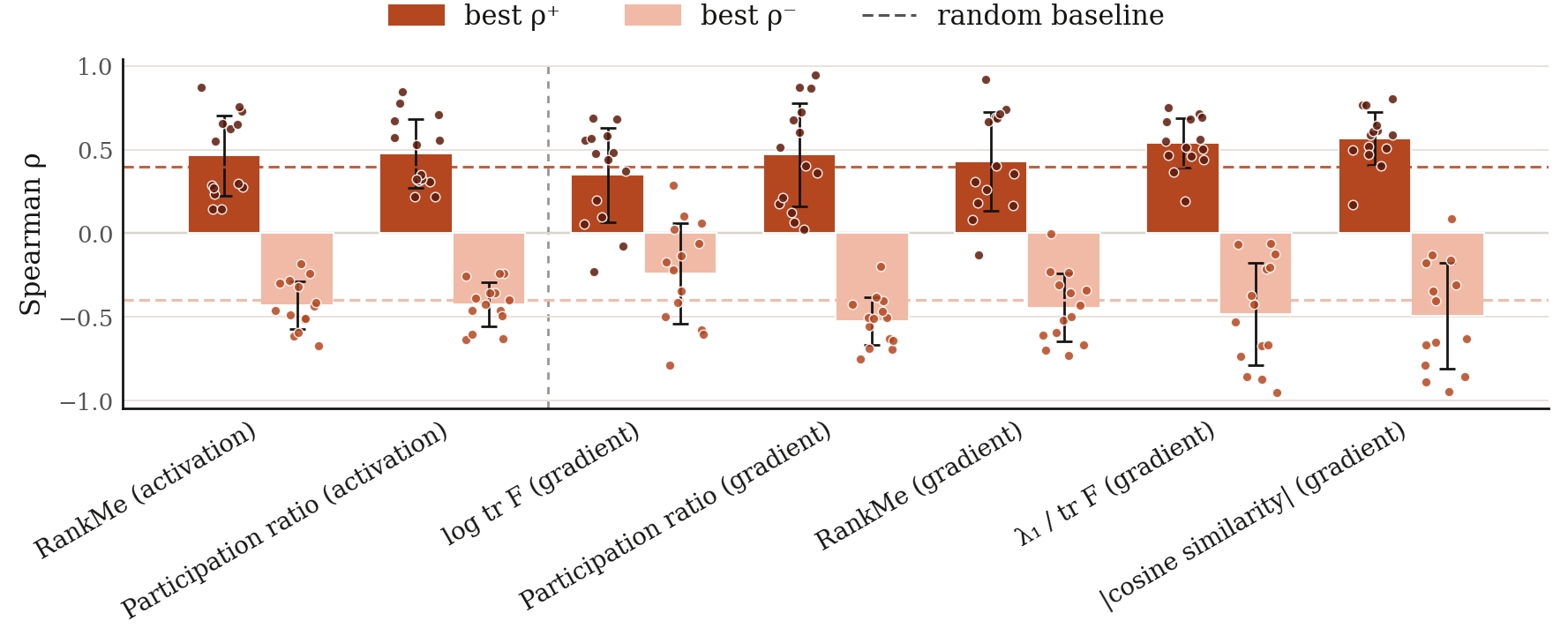}
  \caption{Predicting generalization via metrics based on activations or gradients. For each metric, we measure its correlation with generalization on each layer, and report best positive correlation \(\rho^{+}\) and best negative correlation \(\rho^{-}\) across all layers.}
  \label{fig:18}
\end{figure}

Many metrics achieve non-trivial average correlations, ranging from
0.35 to 0.56 (Figure~\ref{fig:18}). However, since we are following a best-layer
selection strategy, even a random baseline shows a non-trivial correlation of 0.4.

\begin{figure}[H]
  \centering
  \includegraphics[width=\textwidth]{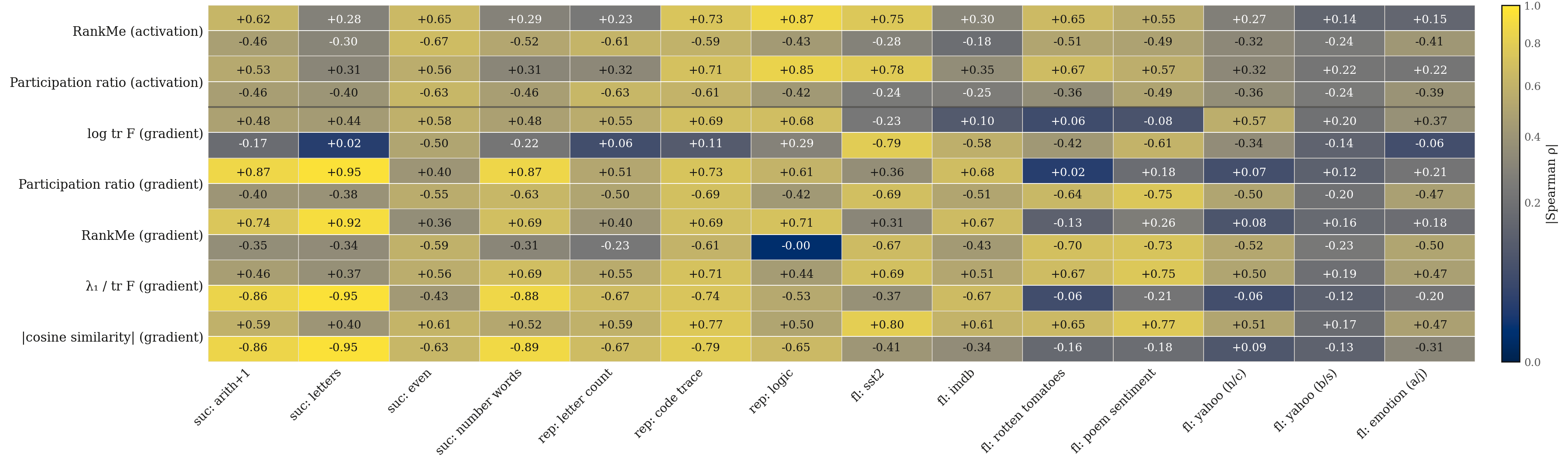}
  \caption{Detailed correlation results on each dataset. Within each cell, we present the best \(\rho^{+}\) (top) and best \(\rho^{-}\) (bottom) of that metric across all layers.}
  \label{fig:19}
\end{figure}

All metrics exhibit high variance across datasets: they achieve strong correlation
(e.g.~0.7 to 0.9) on some datasets and none on others. For
example, Figure~\ref{fig:19} suggests that some datasets (emotion and the Yahoo topic pairs)
are the hardest to predict on average.

\begin{figure}[H]
  \centering
  \includegraphics[width=\textwidth]{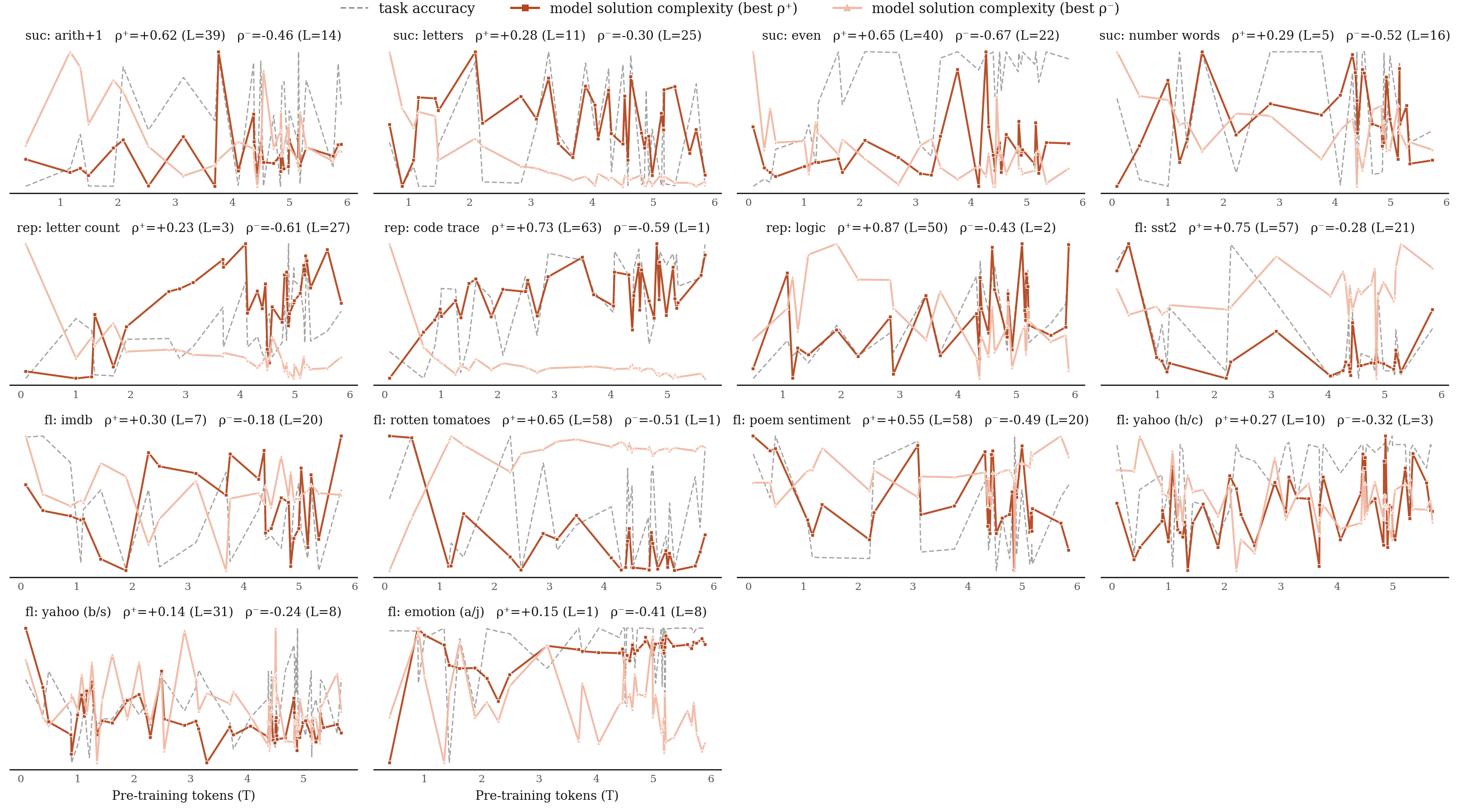}
  \caption{Pre-training dynamics of task accuracy, and the best \(\rho^{+}\) and \(\rho^{-}\) for RankMe computed on activations, across all datasets. The picture is more nuanced than ``simpler solutions generalize better''.}
  \label{fig:ap-rankme-act-full}
\end{figure}

Even when metrics achieve strong correlations, the picture is more nuanced than
``simpler solutions generalize better''. For example, intuitively, activation rank
(i.e.~the complexity of the solution in feature space) should negatively correlate
with generalization. Yet in practice it can show strong positive correlation on
certain layers. Moreover, as Figure~\ref{fig:ap-rankme-act-full} shows, even within the same layer,
well-generalized checkpoints can exhibit either high or low activation rank. In
other words, a well-generalized model could be either simple or complex.
Appendix~\ref{app:metrics} shows the same curves for the other metrics.

\clearpage
\section{Additional model complexity metrics}
\label{app:metrics}

This appendix shows the pre-training dynamics of task accuracy and the best
\(\rho^{+}\) and \(\rho^{-}\) layers for each model complexity metric of Appendix~\ref{sec:5-3}, including RankMe
on activations, which Appendix~\ref{sec:5-3} discusses.

\begin{figure}[H]
  \centering
  \includegraphics[width=\textwidth]{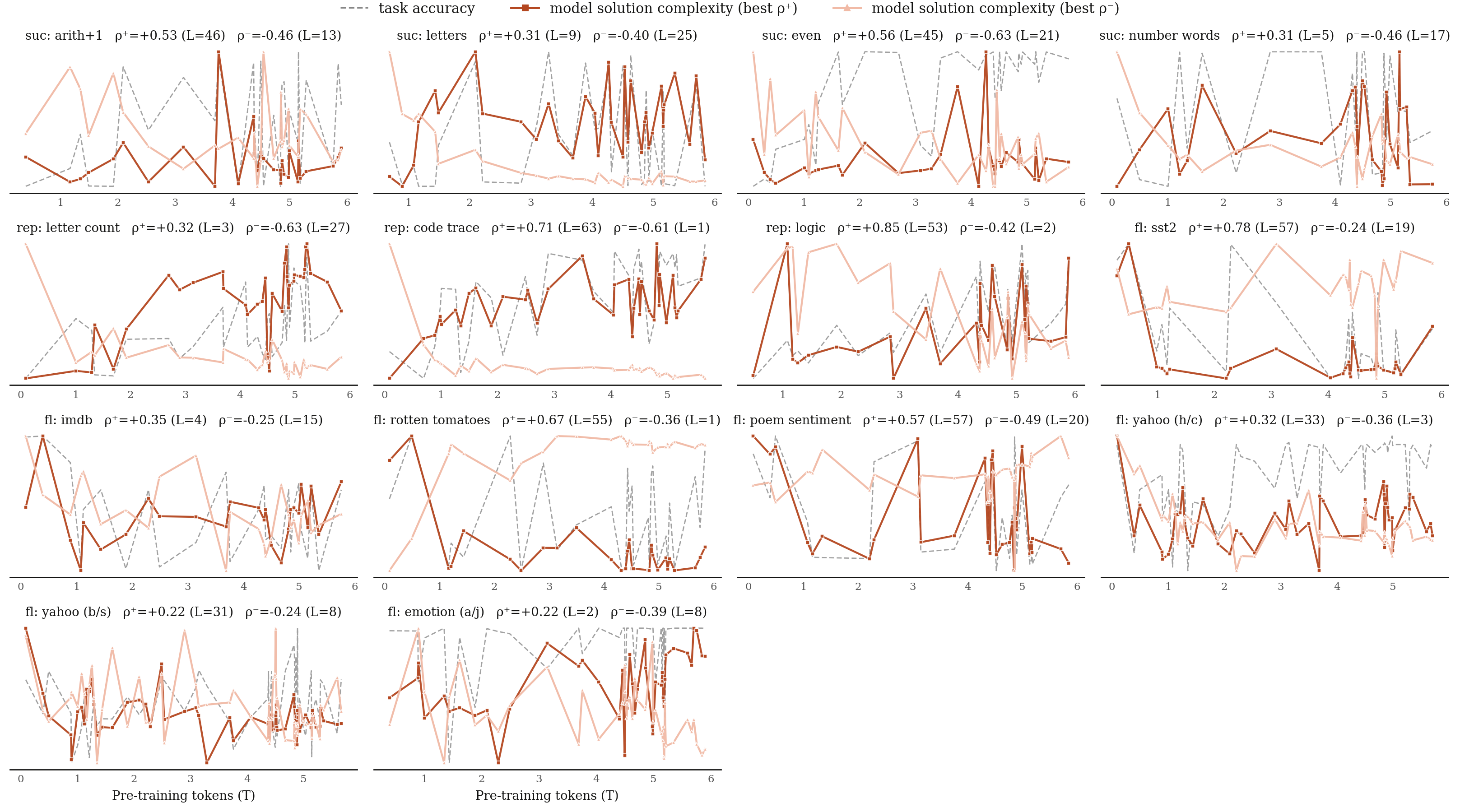}
  \caption{Pre-training dynamics of task accuracy and the best \(\rho^{+}\) and \(\rho^{-}\) layers for participation ratio computed on activations.}
  \label{fig:a-pr-act}
\end{figure}

\begin{figure}[H]
  \centering
  \includegraphics[width=\textwidth]{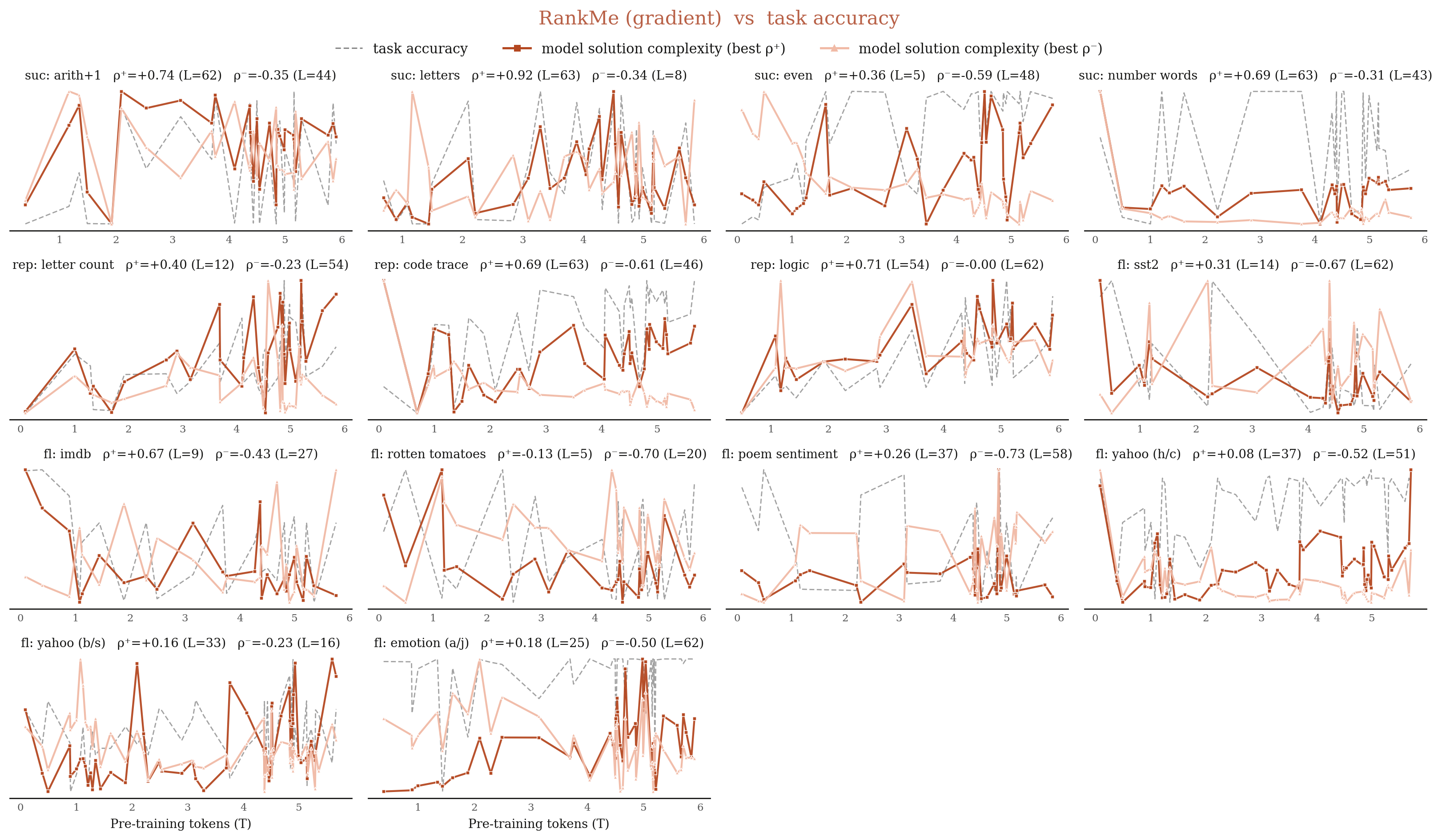}
  \caption{Pre-training dynamics of task accuracy and the best \(\rho^{+}\) and \(\rho^{-}\) layers for RankMe computed on per-example gradients.}
  \label{fig:a-rankme-grad}
\end{figure}

\begin{figure}[H]
  \centering
  \includegraphics[width=\textwidth]{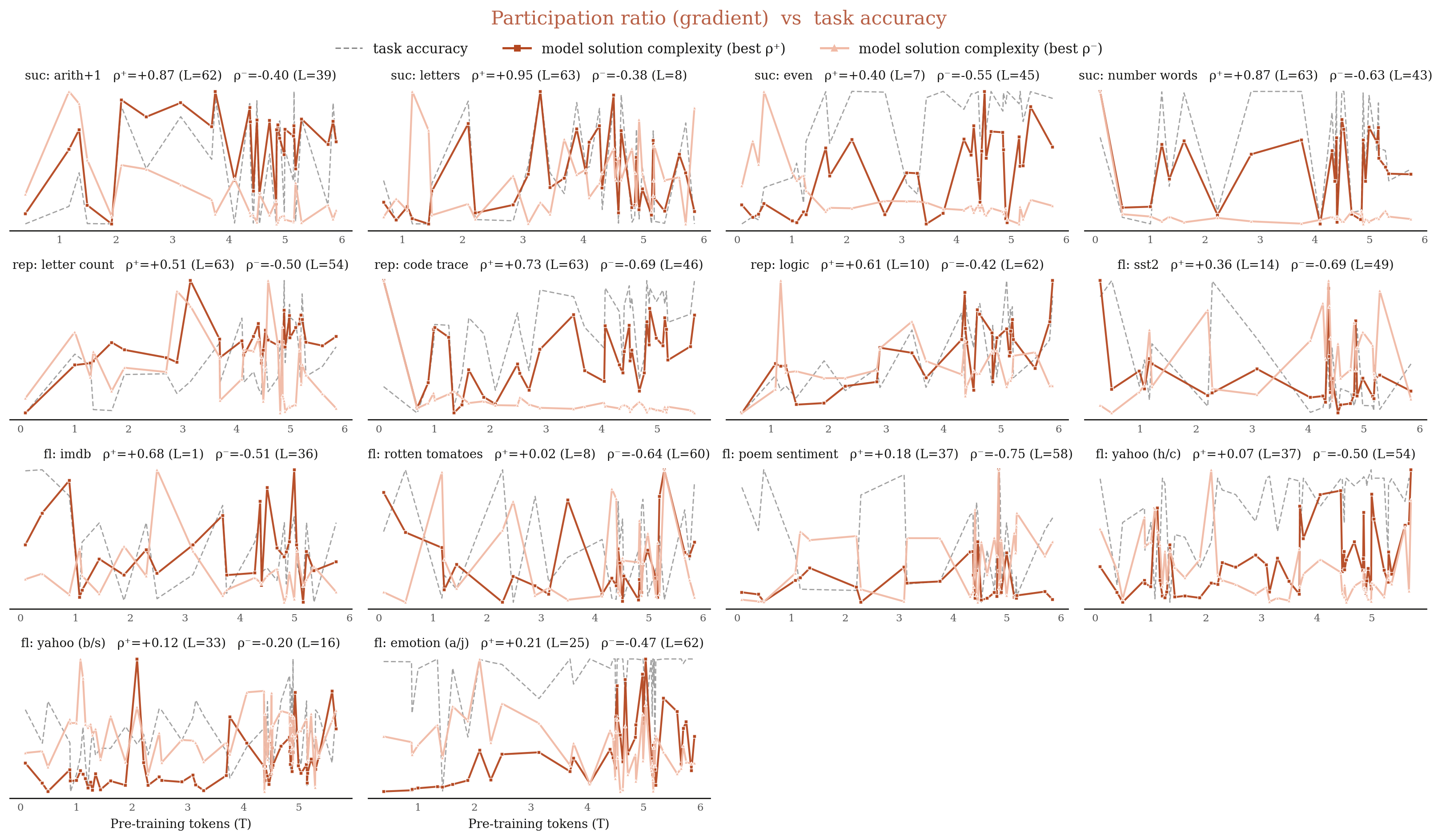}
  \caption{Pre-training dynamics of task accuracy and the best \(\rho^{+}\) and \(\rho^{-}\) layers for the participation ratio of the gradient gram matrix.}
  \label{fig:a-pr-grad}
\end{figure}

\begin{figure}[H]
  \centering
  \includegraphics[width=\textwidth]{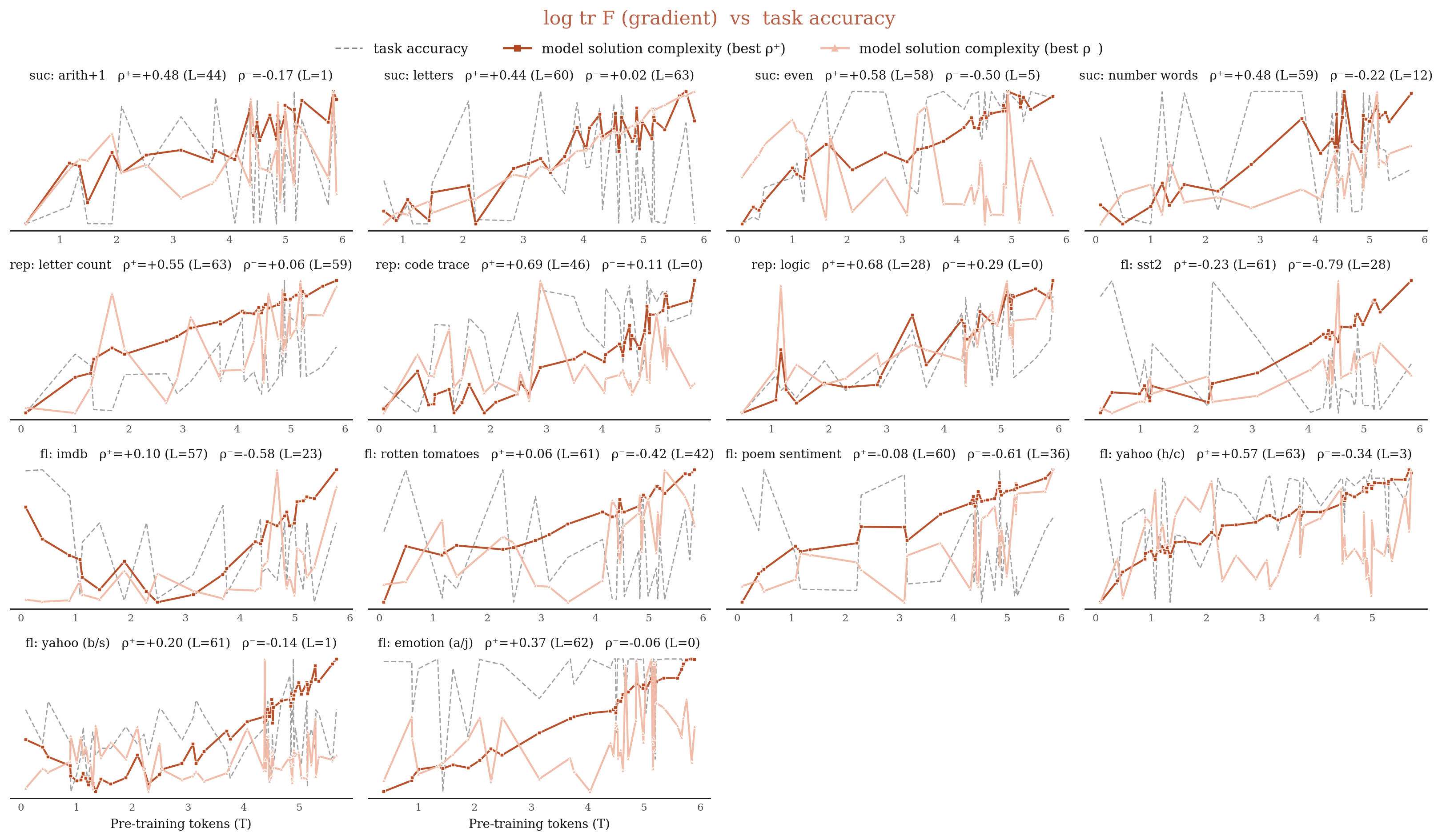}
  \caption{Pre-training dynamics of task accuracy and the best \(\rho^{+}\) and \(\rho^{-}\) layers for \(\log \operatorname{tr} F\), the total per-example gradient magnitude.}
  \label{fig:a-logtrf}
\end{figure}

\begin{figure}[H]
  \centering
  \includegraphics[width=\textwidth]{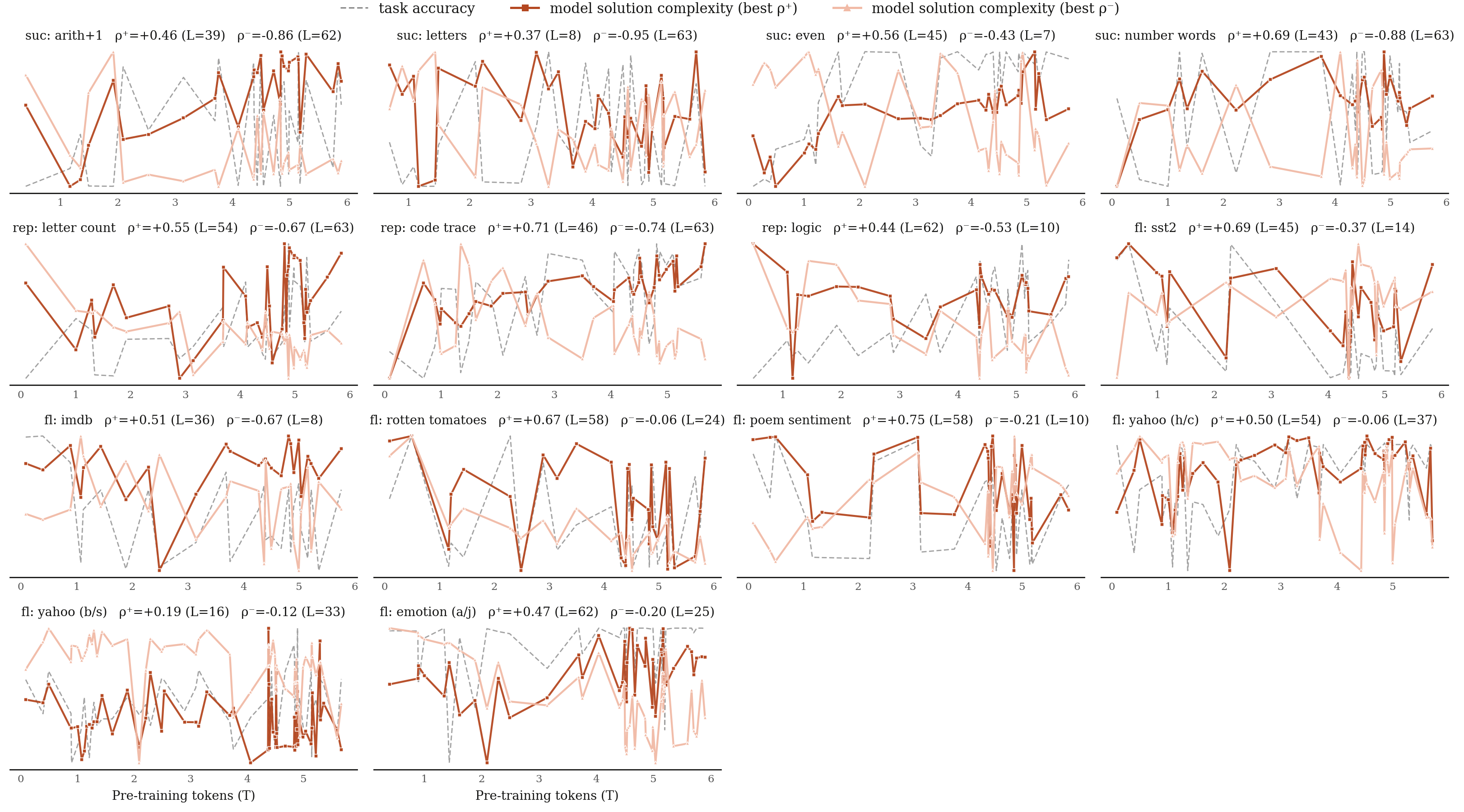}
  \caption{Pre-training dynamics of task accuracy and the best \(\rho^{+}\) and \(\rho^{-}\) layers for \(\lambda_1 / \operatorname{tr} F\), sharpness as a fraction of total curvature.}
  \label{fig:a-lambda1}
\end{figure}

\begin{figure}[H]
  \centering
  \includegraphics[width=\textwidth]{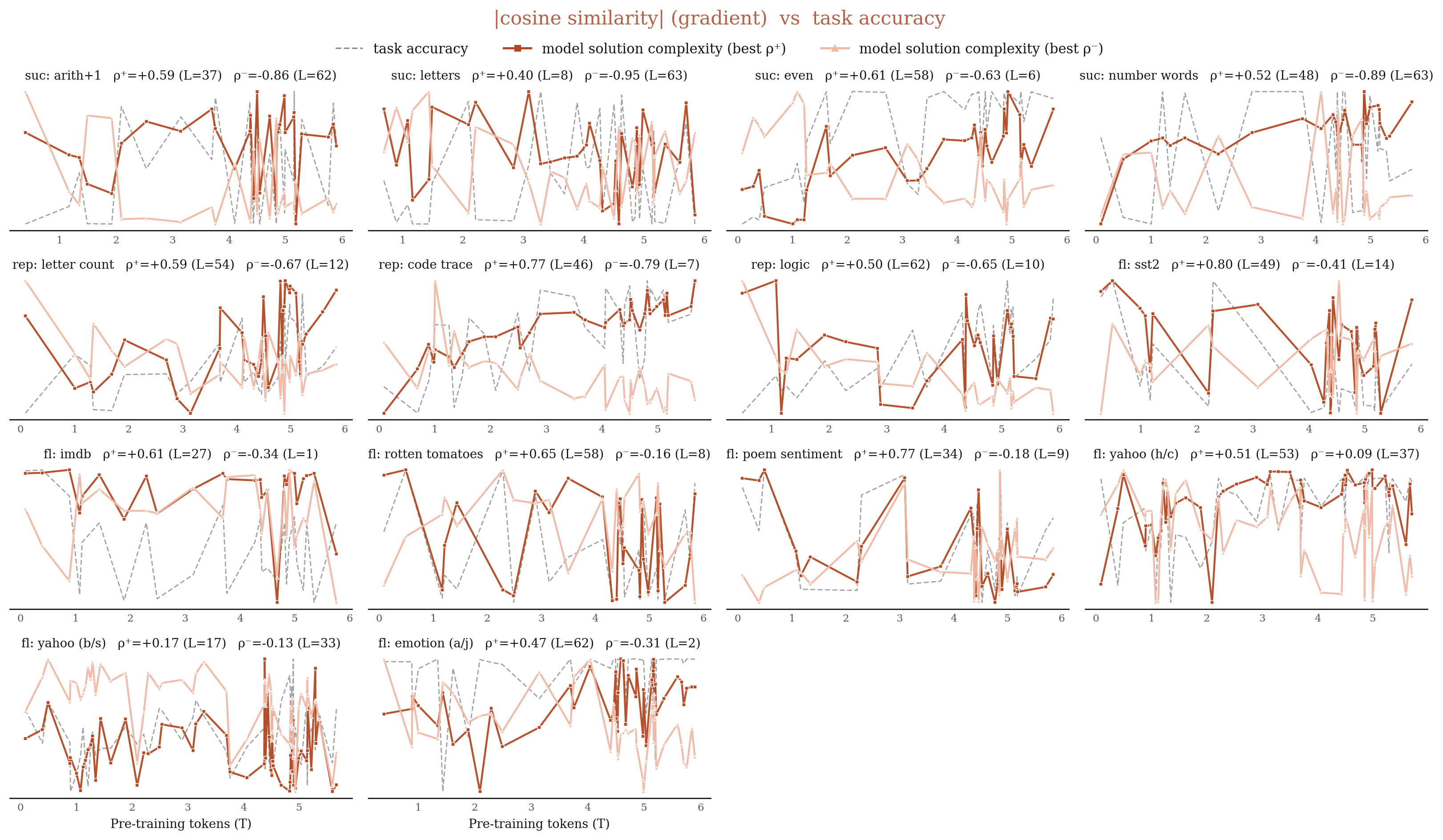}
  \caption{Pre-training dynamics of task accuracy and the best \(\rho^{+}\) and \(\rho^{-}\) layers for $|$cosine similarity$|$, the mean absolute pairwise alignment between per-example gradients.}
  \label{fig:a-cos}
\end{figure}

\FloatBarrier

\end{document}